\documentclass{article} 

\usepackage{arxiv_preamble}

\newcommand{\RR}{\mathbb{R}}

\usepackage{ifthen}
\newcommand{\compileversion}{bodyappendix} 

\def\mytitle{Unlocking Out-of-Distribution Generalization in Transformers via Recursive Latent Space Reasoning}
\title{\mytitle}

\author{%
  {\bfseries Awni Altabaa \qquad Siyu Chen \qquad John Lafferty \qquad Zhuoran Yang} \\[0.5em]
  Department of Statistics \& Data Science, Yale University \\[0.2em]
  \texttt{\{awni.altabaa, siyu.chen.sc3226, john.lafferty, zhuoran.yang\}@yale.edu}
}

\date{} 

\begin{document}

\ifthenelse{\equal{\compileversion}{bodyappendix}}{%
    \makecustomtitle
    {\mytitle}
    {
    Systematic, compositional generalization beyond the training distribution remains a core challenge in machine learning---and a critical bottleneck for the emergent reasoning abilities of modern language models.
    This work investigates out-of-distribution (OOD) generalization in Transformer networks using a GSM8K-style modular arithmetic on computational graphs task as a testbed. 
    We introduce and explore a set of four architectural mechanisms aimed at enhancing OOD generalization: \coloremph{\textit{(i)} input-adaptive recurrence}; \coloremph{\textit{(ii)} algorithmic supervision}; \coloremph{\textit{(iii)} anchored latent representations via a discrete bottleneck}; and \coloremph{\textit{(iv)} an explicit error-correction mechanism}.
    Collectively, these mechanisms yield an architectural approach for native and scalable latent space reasoning in Transformer networks with robust algorithmic generalization capabilities.
    We complement these empirical results with a detailed mechanistic interpretability analysis that reveals how these mechanisms give rise to robust OOD generalization abilities.
}
    {\today}
    {\url{https://github.com/Awni00/algorithmic-generalization-transformer-architectures}}

    \vspace{1em}
    \vskip 0.5em  
\begin{figure}[h!]
    \centering
    \begin{subfigure}[t]{0.24\linewidth}
        \centering
        \includegraphics[width=0.90\linewidth]{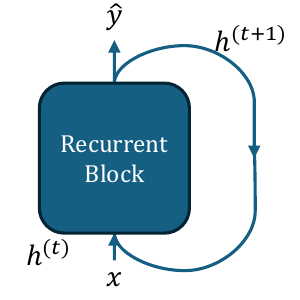}
        \captionsetup{width=.95\linewidth,font=scriptsize}
        \caption{\scriptsize Recurrence \& Adaptive Computation}
    \end{subfigure}
    \begin{subfigure}[t]{0.24\linewidth}
        \centering
        \includegraphics[width=0.90\linewidth]{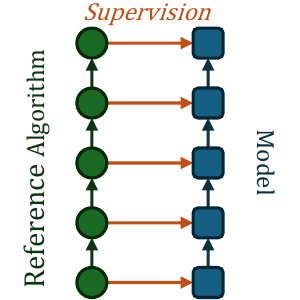}
        \captionsetup{width=.95\linewidth,font=scriptsize}
        \caption{\scriptsize Algorithmic Supervision}
    \end{subfigure}
    \begin{subfigure}[t]{0.24\linewidth}
        \centering
        \includegraphics[width=0.90\linewidth]{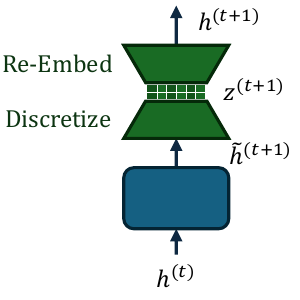}
        \captionsetup{width=.95\linewidth,font=scriptsize}
        \caption{\scriptsize Anchored Discrete Latent Space}
    \end{subfigure}
    \begin{subfigure}[t]{0.24\linewidth}
        \centering
        \includegraphics[width=0.90\linewidth]{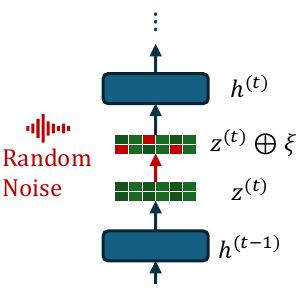}
        \captionsetup{width=.95\linewidth,font=scriptsize}
        \caption{\scriptsize Error Correction}
    \end{subfigure}
    \caption{\textbf{Four key mechanisms enabling robust out-of-distribution generalization in transformer architectures.}
    \textbf{(a)} Recurrence and input-adaptive computation allows models to dynamically allocate computational resources based on problem complexity.
    \textbf{(b)} Algorithmic supervision guides the learning process through structured intermediate representations.
    \textbf{(c)} Anchored discrete latent spaces provide stable reference points for compositional reasoning.
    \textbf{(d)} Error correction mechanisms enable iterative refinement of predictions through feedback loops.
    Together, these mechanisms enable transformers to develop recursive reasoning patterns that generalize beyond their training distribution.}\label{fig:mechanisms}
\end{figure}
    \clearpage

    \vspace{1em}
    \setcounter{tocdepth}{2}
    {\color{YaleBlue}\tableofcontents}
    \clearpage


\section{Introduction}\label{sec:intro}

Systematic algorithmic generalization stands as a critical milestone and a grand challenge in machine learning research~\citep{pollack1990recursive,socher2012semantic,lake2018generalization,velivckovic2021neural}. This ability is fundamental to human cognition, stemming from our capacity for \textit{systematic compositionality}---algebraically producing novel combinations from known components and making strong generalizations from limited data~\citep{chomsky1957syntactic,fodor1988connectionism,lake2017building}. Achieving such generalization necessitates learning universal, scalable problem-solving algorithms. Even in humans, acquiring such algorithmic understanding often requires explicit step-by-step supervision. Once an algorithm is learned, however, humans can generalize its application far beyond the domain of previously encountered stimuli or problems~\citep{anderson1982acquisition,singley1989transfer}.

The reasoning capabilities of artificial intelligence systems have advanced rapidly in recent years, built upon the foundation of large language models. In particular, chain-of-thought (CoT) techniques have been central to enhancing the reasoning capabilities of these systems~\citep{wei2022chain,kojima2022large,chung2022scalinginstructionfinetunedlanguagemodels,liu2023logicot}, especially in domains like mathematics~\citep{cobbe_training_2021,lewkowycz2022solving,lightman2023letsverifystepstep,shao2024deepseekmathpushinglimitsmathematical}.
CoT enables a model to receive supervision on learning a reference problem-solving procedure during training and allows the model to emulate this procedure at test-time. This progress presents a unique opportunity to make significant strides on foundational challenges related to reasoning in artificial intelligence.

Despite these advancements, out-of-distribution (OOD) generalization---particularly the type of \textit{length generalization} involved in algorithmic reasoning (i.e., generalizing from simpler or smaller problem instances to larger or more complex ones)---has remained a central challenge and limitation for Transformer-based~\citep{vaswani2017attention} language models~\citep{anilExploringLengthGeneralization2022,kazemnejadImpactPositionalEncoding2023,jelassi2023lengthgeneralizationarithmetictransformers,zhouWhatAlgorithmsCan2023}. While chain-of-thought techniques alleviate this to some degree by enabling the learning of more complex algorithmic procedures, the ability to generalize far outside the training distribution remains a significant obstacle~\citep{stechly2024chain,zhouTransformersCanAchieve2024}.

In this work, we investigate the architectural and methodological mechanisms that underpin algorithmic OOD generalization in Transformer networks. To facilitate a systematic investigation, we focus our study on a simple yet scalable mathematical reasoning task: performing modular arithmetic on computational graphs. 
This task allows us to study OOD and algorithmic generalization in a controlled manner—with complexity directly parameterized by graph size and depth—while also capturing the core essence of established mathematical reasoning benchmarks like GSM8K~\citep{cobbe_training_2021}, which are central to evaluating the reasoning capabilities of large language models.
Furthermore, this task possesses a compositional nature; it can be solved by learning a core set of skills (e.g., a set of modular arithmetic operations and the ability to traverse the graph one layer at a time) and scaling up their application to solve larger and more complex problem instances. We use this task to explore the following guiding question: 

\begin{highlightbox}
\begin{center}
    What are the \coloremph{architectural mechanisms} and \coloremph{inductive biases} needed for robust \coloremph{OOD algorithmic generalization} in Transformers?
\end{center}
\end{highlightbox}

We find that while standard CoT training techniques enable good in-distribution performance and a limited degree of OOD generalization, the learned solutions are not robust or universal, and their performance rapidly degrades as test inputs grow in complexity beyond the training regime. We propose and explore a set of four simple architectural and methodological mechanisms, built upon the Transformer architecture, to facilitate the learning of robust and generalizable algorithmic solutions: \coloremph{\textit{(i)} input-adaptive recurrence}; \coloremph{\textit{(ii)} algorithmic supervision}; \coloremph{\textit{(iii)} anchored latent representations via a discrete bottleneck}; and \coloremph{\textit{(iv)} an explicit error-correction mechanism}. When combined, these mechanisms yield an architectural approach for native and scalable \coloremph{\textit{latent space reasoning}} in Transformer networks, demonstrating robust algorithmic generalization capabilities. In particular, on our mathematical reasoning task, our method achieves perfect generalization on inputs that are several times larger than those seen during training. We complement our architectural proposal and empirical results with a mechanistic interpretability analysis to reveal \textit{how} these architectural proposals enable sharp OOD generalization, what circuits they learn, and why those circuits facilitate robust OOD generalization.



\section{Related Work}\label{sec:related_work}

Our work is related to several strands of fundamental machine learning research, including issues of out-of-distribution generalization, architectural mechanisms such as recurrence and discretization, chain-of-thought and intermediate supervision methods, and work on mechanistic interpretability techniques.

\textbf{Out-of-Distribution Generalization.} Out-of-distribution (OOD) generalization, along with related capabilities such as compositionality and systematicity, poses a fundamental challenge in machine learning research~\citep{pollack1990recursive,baxter2000model,socher2012semantic,barrett2018measuring,hupkes2020compositionality}. These capabilities are crucial for developing AI systems that can reliably apply learned knowledge to novel scenarios, a hallmark of robust intelligence~\citep{fodor1988connectionism,lake2017building,goyal2022inductive}. A particularly important type of OOD generalization, especially for algorithmic reasoning tasks, is \textit{length generalization}---the ability to generalize from simpler or shorter training instances to significantly longer and more structurally complex instances. This has proven to be a key limitation of Transformer-based~\citep{vaswani2017attention} language models~\citep{anilExploringLengthGeneralization2022,kazemnejadImpactPositionalEncoding2023,jelassi2023lengthgeneralizationarithmetictransformers,zhouWhatAlgorithmsCan2023}. While chain-of-thought techniques alleviate this to some degree by enabling the learning of more complex algorithmic procedures, the ability to generalize far outside the training distribution remains a significant obstacle~\citep{stechly2024chain,zhouTransformersCanAchieve2024}.

\textbf{Recurrence.} Recurrence forms a foundational architectural principle in neural networks, particularly for tasks that involve sequential data or inherently iterative processes~\citep{elman1990finding,jordan1997serial,hochreiter1997long}. These architectures are designed to emulate step-by-step computations by maintaining and updating an internal state, making them well-aligned with problems that have a recursive or layered solution structure. Sequence-to-sequence recurrent architectures for sequence transduction and neural machine translation advanced the state of the art~\citep{cho-etal-2014-learning,sutskever2014sequence}, and were instrumental to the development of attention mechanisms and the Transformer architecture~\citep{vaswani2017attention}. While standard Transformers do not possess a recurrent structure, recurrent variants of the Transformer architecture were explored soon after its introduction~\citep{dehghaniUniversalTransformers2019}. Whereas standard recurrent neural networks apply their recurrence across time or sequence length, recurrent Transformer architectures are \textit{parallel in time} due to the parallel attention mechanism, but recurrent across computational depth---that is, the same Transformer layer is applied iteratively to the sequence as a whole. The recurrent inductive biases have been demonstrated to confer certain advantages in generalization~\citep{fanLoopedTransformersLength2024,yangLoopedTransformersAre2024}. In our work, recurrence is a key architectural mechanism encoding important inductive biases that aid the discovery of scalable recursive algorithms for solving the underlying mathematical problem.

\textbf{Adaptive Computation.} A critical challenge is handling inputs with varying complexity, where a fixed amount of computation may be inefficient or insufficient. This motivates the concept of adaptive computation, wherein a model can dynamically adjust its computation time, for example by varying the number of recurrent iterations, based on the demands of the input. An important work in this domain is the \textit{Adaptive Computation Time (ACT)} mechanism proposed by~\citet{gravesAdaptiveComputationTime2017} for recurrent neural networks, which explicitly models and learns how many computational steps are needed as a function of the input. A version of the ACT mechanism is incorporated in the recurrent Transformer architecture proposed by~\citet{dehghaniUniversalTransformers2019}. However, a drawback of such mechanisms is their complexity and difficulty of training. Although efforts have been made to explore simpler adaptive computation methods~\citep{baninoPonderNetLearningPonder2021a}, an even simpler approach is explored by~\citet{schwarzschildCanYouLearn2021,bansalEndtoendAlgorithmSynthesis2022}, where the halting time is not explicitly modeled by the network, and instead the number of recurrent iterations is scaled at inference time based on the size of the input. This simpler approach can be easier to train, and has been shown to improve out-of-distribution generalization. More recently,~\citet{geipingScalingTestTimeCompute2025} explored the viability of this approach as a way to perform test-time scaling in large language models. In our work, we similarly scale computation time by proportionately scaling the number of recurrent iterations in order to solve more complex problem instances, generalizing far beyond the training distribution.

\textbf{Discreteness in Neural Networks.} Symbolic AI systems derive their power from manipulating discrete symbols according to well-defined rules, which enables robust, precise, and interpretable reasoning~\citep{newel1976computer,fodor1988connectionism}. Given this rich tradition of using discrete symbolic states in artificial intelligence, many works have subsequently explored incorporating such discrete latent representations into neural networks~\citep{garcez2008neural,salakhutdinov2009deep,courville2011spike,agustsson2017soft,oordNeuralDiscreteRepresentation2018}. Additionally, discreteness is often a central characteristic of \textit{constructions} of Transformer networks for specific tasks. For example,~\citet{weissThinkingTransformers2021} develops a programming language that represents Transformer-based computation with discrete internal mechanisms. Additionally,~\citet{smolensky2024mechanismssymbolprocessingincontext} constructs a Transformer network for a compositional in-context learning task, which features discreteness in both its latent states and attention mechanism. In our work, we explore the use of discrete latent states as a means of \textit{anchoring} the latent representation to a common, depth-invariant space to enable scaling computation far beyond the training distribution while avoiding representational shift across computational depth.

\textbf{Chain-of-Thought \& Algorithmic Supervision.} Chain-of-thought techniques have been central to enhancing the reasoning capabilities of large language models. Early usage of the term ``chain-of-thought'' referred to prompting techniques that condition a model to generate a sequence of intermediate steps before arriving at the final answer~\citep{kojima2022large,wei2022chain,nye2021show}. For example,~\citet{wei2022chain} demonstrated that prompting the LLM with a few CoT exemplars caused the model to generate an analogous step-by-step solution, which significantly improved performance on a range of arithmetic, commonsense, and symbolic reasoning tasks.~\citet{kojima2022large} showed that LLMs can be ``zero-shot'' reasoners in the sense that simply asking the model to reason step-by-step, without providing in-context learning CoT exemplars, can be sufficient to elicit chain-of-thought-style reasoning and improve performance. Modern usage of the term ``chain-of-thought'' has extended beyond prompting methods, as it now forms a key component of the \textit{training} pipeline of LLMs, wherein a model is explicitly trained on demonstrations of step-by-step solutions to problems of interest, such as mathematical reasoning~\citep{chung2024scaling,liu2023logicot,lewkowycz2022solving}. In some situations, chain-of-thought training can be interpreted as providing explicit supervision to align the model to a particular algorithm or procedure for solving a problem, as opposed to simply providing supervision via input-output examples. In our work, we explore traditional chain-of-thought training techniques as baselines, as well as incorporate algorithmic supervision to the internal states of our proposed method.

\textbf{Mechanistic Interpretability.} In our work, we carry out a mechanistic interpretability analysis to probe \textit{how} the model has learned to solve the task and \textit{why} it can do so robustly, generalizing far outside the training distribution. In recent years, there has been a resurgence in work on interpretability, with new techniques being introduced that aim to understand modern large language models~\citep{elhage2021mathematical,meng2022locating,elhage2022superposition,olsson2022context,bricken2023monosemanticity,ameisen2025circuit}.~\citet{elhage2021mathematical} is an influential work in this area of research, introducing a conceptual framework and new terminology that continues to be used in subsequent work. A key early achievement in this line of work is the discovery of ``induction head'' circuits in large language models~\citep {olsson2022context}, which perform a two-step copying operation that is crucial for in-context learning. In our work, we identify a similar mechanism in our recurrent models that is used to copy previously computed variable values. This involves first retrieving the parent variables' names in the first layer, then using these variable names to retrieve their values in the second layer, which are computed elsewhere in the sequence of latent states. Such work is often described as \textit{circuit analysis}, where the goal is to identify sub-networks that are responsible for particular functions. A key method for validating hypotheses about the functions of different model components is \textit{causal interventions} like activation patching or ablations~\citep{meng2022locating,geiger2021causal,geiger2024finding}, which involves systematically modifying parts of the model or input to observe effects on behavior or internal states. We use related causal intervention techniques in our own mechanistic interpretability analysis in this work. Finally, the work by~\citet{nandaprogress,tian2024composing} is relevant as it specifically investigated how Transformers perform arithmetic, reverse-engineering a modular addition algorithm learned by the feedforward network in a Transformer layer—a phenomenon we also observe in our models.

\section{Problem Setup}

\subsection{Task Description: Modular Arithmetic on Computational Graphs}

We formally introduce the task of \textit{modular arithmetic on computational graphs} as follows. 

\textbf{Task Description.} A \textit{computation graph} is a directed acyclic graph (DAG) representing a network of mathematical computations, where nodes correspond to variables and edges describe the dependencies between them. 
As illustrated in \Cref{fig:dag_example} with an example, the \textit{\color{ForestGreen} leaf nodes} in this DAG are directly assigned numerical values (e.g., ${\color{ForestGreen}x_7}\gets 20$). All other \textit{\color{RoyalBlue} non-leaf nodes} are defined as functions of their parent nodes in the computation graph.
In particular, the value of each non-leaf node is computed by applying one or more specified operations to the values of its parent nodes. In our experiments, we consider \textit{modular arithmetic} operations (addition, multiplication, or subtraction), with the prime number $p=23$ as the modular base. For example, in \Cref{fig:dag_example} we have 
 ${\color{RoyalBlue} x_{23}} \gets { \color{ForestGreen} x_{7}} {\color{OrangeRed} +} {\color{ForestGreen}   x_{42}} {\color{OrangeRed}(\mathtt{mod}~ p)}$
and ${\color{RoyalBlue} x_{101}} \gets { \color{RoyalBlue} x_{23}} {\color{OrangeRed} \times} {\color{RoyalBlue}   x_{91}} {\color{OrangeRed}(\mathtt{mod}~ p)}$. 
 In the following, we let $N$ and $L$ denote the total number of nodes and the number of leaf nodes, respectively. 
 We consider graphs with up to $128$ nodes, and let $\calV = \sset{x_{1}, \ldots, x_{128}}$ denote the set of variable names. 
 

 \textbf{Data Generation Process.} A problem instance in this task is specified by the \emph{values of the leaf nodes} and \emph{a computation graph} depicting the computations that determine the values of all non-leaf nodes. In particular, given parameters $N$ and $L$, an input instance is generated as follows: 
 \begin{enumerate}[leftmargin=30pt, itemsep=0pt, parsep=0pt] 
    \item [(i)] Randomly generate a DAG with $N$ nodes, $L$ of which are leaf nodes.  
    \item [(ii)] Randomly assign a variable name from $\calV$ to each node.
    \item [(iii)] Randomly assign numerical values to the leaf nodes from $\calN = \sset{0, 1, \ldots, 22}$.
    \item [(iv)] For each non-leaf node, randomly assign operations from $\calO = \sset{+, -, \times}$ to define its computation based on its parent nodes
\end{enumerate}
The instance generated by (i)--(iv) is stored as a token sequence, where each variable name, numerical value, and operation is assigned a unique token. 
A special \emph{separation token} $\sep$ is used to separate different formulas.  
For example, the instance depicted in~\Cref{fig:dag_example} is represented as the following \textit{token sequence}:

{\small
\begin{empheq} 
    [box=\fcolorbox{MorandiPink}{MorandiPink!15}]{equation}\label{eq:token_sequence}
    \begin{aligned}
    &\valtoken{20} \token{\to} \vartoken{x_7} \sep 
     \valtoken{2} \token{\to} \vartoken{x_{42}} \sep 
     \valtoken{6} \token{\to} \vartoken{x_{88}} \sep 
     \valtoken{14} \token{\to} \vartoken{x_{115}} \\
    &\vartoken{x_7} \opertoken{+} \vartoken{x_{42}} \token{\to} \vartoken{x_{23}} \sep 
     \vartoken{x_{42}} \opertoken{+} \vartoken{x_{88}} \token{\to} \vartoken{x_{91}} \sep 
     \vartoken{x_{88}} \opertoken{\times} \vartoken{x_{115}} \token{\to} \vartoken{x_{55}} \\
    &\vartoken{x_{23}} \opertoken{\times} \vartoken{x_{91}} \token{\to} \vartoken{x_{101}} \sep 
     \vartoken{x_{91}} \opertoken{-} \vartoken{x_{88}} \opertoken{+} \vartoken{x_{55}} \token{\to} \vartoken{x_{30}}
    \end{aligned}
\end{empheq}}

\textbf{Target Output \& Evaluation Metric.} 
Given a generated problem instance, the task is to compute the value of every node in the computation graph; these values are uniquely determined by steps (i)--(iv) above.
We consider the model output to be correct only if \textit{all} node values are computed correctly (i.e., the input graph is fully solved). 

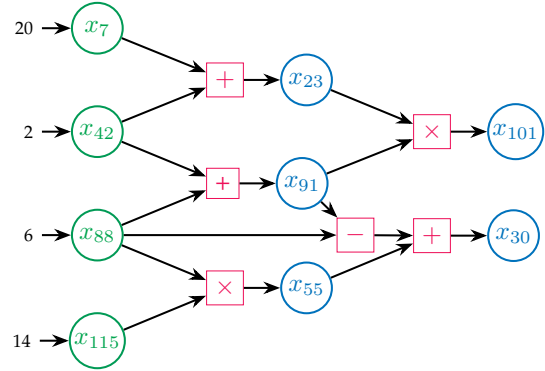
\begin{wrapfigure}{R}{0.45\linewidth}
    \centering
    \resizebox{\linewidth}{!}{
    \begin{tikzpicture}[
    node distance=0.7cm and 2.0cm,
    var/.style={draw, circle, minimum size=0.7cm, inner sep=1pt, thick},
    greenvar/.style={var, draw=ForestGreen, text=Green},
    bluevar/.style={var, draw=RoyalBlue, text=RoyalBlue},
    val/.style={font=\scriptsize},
    arr/.style={->, >=Stealth[], thick},
    block/.style={draw, rectangle, minimum height=0.7cm, minimum width=1.0cm, thick}
]

\node[greenvar] (l1n1) {$x_{7}$};
\node[val, left=0.4cm of l1n1] (v1) {20};
\draw[arr] (v1) -- (l1n1);

\node[greenvar, below=of l1n1] (l1n2) {$x_{42}$};
\node[val, left=0.4cm of l1n2] (v2) {2};
\draw[arr] (v2) -- (l1n2);

\node[greenvar, below=of l1n2] (l1n3) {$x_{88}$};
\node[val, left=0.4cm of l1n3] (v3) {6};
\draw[arr] (v3) -- (l1n3);

\node[greenvar, below=of l1n3] (l1n4) {$x_{115}$};
\node[val, left=0.4cm of l1n4] (v4) {14};
\draw[arr] (v4) -- (l1n4);

\node[draw = OrangeRed, text = OrangeRed, right=1.5 cm of $(l1n1)!0.5!(l1n2)$] (op1) {$+$};
\draw[arr] (l1n1) -- (op1);
\draw[arr] (l1n2) -- (op1);

\node[draw = OrangeRed, text = OrangeRed, minimum size=0.2cm, right=1.5cm of $(l1n2)!0.5!(l1n3)$] (op3) {+};
\draw[arr] (l1n2) -- (op3);
\draw[arr] (l1n3) -- (op3);

\node[draw = OrangeRed, text = OrangeRed, minimum size=0.2cm, right=1.5cm of $(l1n3)!0.5!(l1n4)$] (op4) {$\times$};
\draw[arr] (l1n3) -- (op4);
\draw[arr] (l1n4) -- (op4);

\node[bluevar, right=0.5cm of op1](l2n1){$x_{23}$};
\node[bluevar, right=0.5cm of op3] (l2n2) {$x_{91}$};
\node[bluevar, right=0.5cm of op4] (l2n3){$x_{55}$};
\draw[arr] (op1) -- (l2n1);
\draw[arr] (op3) -- (l2n2);
\draw[arr] (op4) -- (l2n3);

\node[draw = OrangeRed, text = OrangeRed, minimum size=0.2cm, right = 1.5cm of $(l2n1)!0.5!(l2n2)$] (op5) {$\times$};
\draw[arr] (l2n1) -- (op5);
\draw[arr] (l2n2) -- (op5);

\node[draw = OrangeRed, text = OrangeRed, minimum size=0.2cm, right=3.3cm of $(l1n3)$] (op6) {$-$};
\draw[arr] (l2n2) -- (op6);
\draw[arr] (l1n3) -- (op6);
\node[draw = OrangeRed, text = OrangeRed, minimum size=0.2cm, right=1.5cm of $(l2n2)!0.5!(l2n3)$] (op7) {$+$};
\draw[arr] (op6) -- (op7);
\draw[arr] (l2n3) -- (op7);

\node[bluevar, right= 0.5cm of op5] (l3n1) {$x_{101}$};
\node[bluevar, right= 0.5cm of op7] (l3n2) {$x_{30}$};

\draw[arr] (op5) -- (l3n1);
\draw[arr] (op7) -- (l3n2);

\end{tikzpicture}
    }
    \caption{An illustration of an instance in \textit{modular arithmetic on computational graphs} task. The goal is to compute the values of all nodes in the graph. For example, here $x_{23} = 20 + 2 = 22$ and $x_{55} = 6 \times 14 = 15$. Recall that we consider modular arithmetic with base $p = 23$.}
    \label{fig:dag_example}
\end{wrapfigure}

\textbf{Out-of-Distribution  Generalization.} 
Our primary focus in this work is to investigate the ability of Transformer networks to learn general problem-solving procedures or algorithms that enable \textit{out-of-distribution} (OOD) generalization. 
The complexity of each problem instance can be explicitly parameterized by graph size, enabling precise measurement of a model's ability to generalize to inputs more complex than those encountered during training. 
In particular, in this mathematical reasoning task, OOD generalization is evaluated by training Transformer models on problem instances with ${\color{RoyalBlue} N \leq 32}$ nodes and testing them on instances of varying sizes, up to ${\color{RoyalBlue} N = 128}$ (a fourfold increase). 
Such generalization requires the ability to process larger inputs and adaptively scale computation time during testing, beyond what was encountered in the training regime.

This synthetic task captures the core essence of mathematical reasoning benchmarks like GSM8K~\citep{cobbe_training_2021}, which are pivotal for evaluating the reasoning capabilities of large language models. Similar to GSM8K, our task involves a combinatorial structure combined with arithmetic computations. However, a key simplification is that variable names are directly tokenized, bypassing natural language representation. This focused design, while retaining the critical combinatorial structure and rule-based nature inherent in mathematical reasoning, facilitates a more straightforward and modular interpretation of the learned Transformer model's internal mechanisms, as will be shown in \Cref{sec:mechinterp}.

\subsection{Limitations of Standard Transformers with CoT Training}

To establish a baseline and motivate the need for alternative approaches, we evaluate standard Transformer architectures on our synthetic task using two primary training paradigms. 

\textbf{End-to-End Training.}  The first baseline is \methodcolorhighlight{feedforwardcolor}{End-to-End} training, where the Transformer models are trained to directly output the final values of all nodes given the problem input, without explicit intermediate steps. The input token sequences are in the form of \eqref{eq:token_sequence}, and we employ various  Transformer models with diverse architectures. See \Cref{sec:cot_details} for details. 

\textbf{Chain-of-Thought (CoT) Training.} The second baseline is based on autoregressive \methodcolorhighlight{cotsupervisioncolor}{Chain-of-Thought (CoT)}  training~\citep{wei2022chain,cobbe_training_2021,lewkowycz2022solving,chung2024scaling,yePhysicsLanguageModels2024}, a prevalent technique for enabling multi-step reasoning in LLMs.  Instead of directly outputting the final answer, CoT trains a model to generate a sequence of intermediate reasoning steps (the ``thought process'') that culminates in the solution. 
For our task, CoT intermediate steps consist of explicit demonstrations of the step-by-step computation of nodes within a given computation graph. 
In particular, in \methodcolorhighlight{cotsupervisioncolor}{CoT} training, the Transformer model receives an input prompt consisting of the token representation of the computation graph (as in \eqref{eq:token_sequence}), followed by a special $\token{\mathtt{CoT}}$ token. This special token signals the beginning of the CoT reasoning, which outlines the computation of each node in topological order. Each step in the trajectory involves: (1) recalling the equation defining the node's value, (2) recalling the values of its dependent nodes, and (3) performing the arithmetic computation. For example, computing node $\token{x_{101}}$ from \Cref{fig:dag_example} would appear in the CoT as:
\begin{equation*}
    {\color{CadetBlue}\mathtt{[...Input\, Prompt...]}} \token{\mathtt{CoT}} {\color{CadetBlue}\mathtt{[...]}} \vartoken{x_{101}} = \vartoken{x_{23}} \MUL \vartoken{x_{91}} = \valtoken{22} \MUL \valtoken{8} = \valtoken{15}
\end{equation*}
Here, the ${\color{CadetBlue}\mathtt{[...Input\, Prompt...]}}$ gives the description of the problem instance, and ${\color{CadetBlue}\mathtt{[...]}}$ denotes the preceding portion of the chain-of-thought trajectory up to node $\token{x_{91}}$, which in particular includes the computation of the values of $\token{x_{23}}$ and $\token{x_{91}}$. An example of a full CoT example from the training data is provided in~\Cref{subsec:cot}. 

\textbf{Implementation.}
We train  causal Transformer models from scratch using both \methodcolorhighlight{feedforwardcolor}{End-to-End} 
 and \methodcolorhighlight{cotsupervisioncolor}{CoT}  supervision on randomly generated problem instances  with graph sizes {\color{RoyalBlue}$N \leq 32$}. 
 At inference time, models are prompted with the input and generation is performed using \emph{greedy decoding}. 
 End-to-End models directly output all node values given the input, while 
 CoT models autoregressively generate the solution, including the full CoT trajectory. We evaluate performance based on the proportion of instances where the model computes \emph{all} node values correctly, with a particular focus on OOD generalization to new, randomly generated graphs of varying sizes up to $N = 128$. 
 For all methods, an extensive hyperparameter search was conducted (covering layers, model dimension, and positional encoding), and the best-performing configuration of each method was selected for comparison. A detailed experimental setup for these baseline experiments is provided in~\Cref{sec:cot_details}.

\textbf{Observed OOD Generalization Deficiencies.}
We find that \methodcolorhighlight{cotsupervisioncolor}{Chain-of-Thought} training enables models to solve larger graphs compared to those trained \methodcolorhighlight{feedforwardcolor}{End-to-End} without chain-of-thought supervision (\Cref{fig:method_ood_comparison}). While the best-performing CoT models exhibit \emph{a limited degree of OOD generalization} to moderately larger graphs ($N \leq 32 \leadsto N\approx 40$), this capability rapidly deteriorates as graph sizes exceed the training regime. 
In the next section, we propose a series of architectural mechanisms that address these generalization challenges.

\section{Reasoning in Latent Space with Algorithmic Supervision}\label{sec:our_method}

\subsection{Mechanisms for Effective OOD Generalization.} \label{subsec:mechanisms}


Effective OOD generalization on complex reasoning tasks hinges on a model's ability to learn and emulate an underlying scalable \textit{algorithm}. This requires the model to, implicitly or explicitly, execute an iterative procedure that adapts to input complexity. Designing \emph{inductive biases} to support the discovery of such scalable, compositional solutions is a central challenge in machine learning~\citep{baxter2000model,lake2018generalization,barrett2018measuring,goyal2022inductive}. Chain-of-thought (CoT) techniques attempt this by having the model sequentially generate a token representation of a computational process. However, this restriction to a token-based, autoregressive format often yields brittle ``algorithms" that fail to generalize robustly, especially as longer CoT sequences are needed for more complex inputs. These well-documented length generalization issues~\citep{anilExploringLengthGeneralization2022,kazemnejadImpactPositionalEncoding2023,jelassi2023lengthgeneralizationarithmetictransformers,zhouWhatAlgorithmsCan2023,stechly2024chain,zhouTransformersCanAchieve2024} underscore CoT's limitations in effectively emulating truly scalable algorithmic procedures. This work, therefore, proposes alternative mechanisms to facilitate the learning of such iterative algorithms directly within a model's latent processing. 

Our proposal features four key architectural mechanisms: \coloremph{\emph{(i)} recurrent Transformer blocks}, \coloremph{\emph{(ii)} algorithmic supervision}, \coloremph{\emph{(iii)} discretization in latent space}, and \coloremph{\emph{(iv)} a self-correction scheme}. 
Collectively, these mechanisms constitute an architecture enabling native latent-space reasoning, leading to effective OOD generalization. \Cref{fig:mechanisms} illustrates the four mechanisms as individual components, while \Cref{fig:architecture_diagram} depicts the unified architectural proposal. In the following, we present the four proposed mechanisms and the essence of their implementation, deferring certain implementation details to~\Cref{sec:latent_supervision_details}. 

\textbf{Algorithm to Emulate.} To solve this task, a natural algorithmic solution that is well-aligned with the Transformer architecture is to \coloremph{compute the values in the computation graph one layer at a time}.  This can be realized through a recursive process that iteratively applies the same computational modules. 
Specifically, each iteration of the algorithm computes values one layer deeper in the computation graph by fetching the necessary dependent values for nodes at the current layer and then performing the required modular arithmetic.
In particular, for the example in \Cref{fig:dag_example}, in the first iteration, we evaluate variables $\{ {\color{ForestGreen} x_{7}}, { \color{ForestGreen} x_{42}},{ \color{ForestGreen} x_{88}}, { \color{ForestGreen} x_{115}}\} $. In the second iteration, we evaluate $\{ {\color{RoyalBlue} x_{23}}, {\color{RoyalBlue} x_{91}}, {\color{RoyalBlue} x_{55}}\}$. In the last iteration, we evaluate $\{ {\color{RoyalBlue} x_{101}}, {\color{RoyalBlue} x_{30}} \}$. Note that each iteration involves the same type of computation, providing a succinct and scalable recursive problem-solving algorithm.

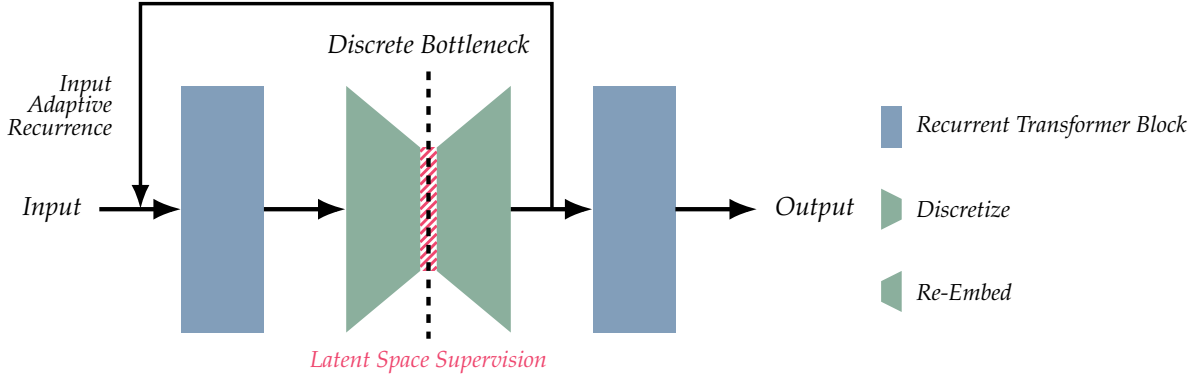
\begin{figure}
    \centering
    \resizebox{\linewidth}{!}{
        \definecolor{transformerblockcol}{HTML}{829EBA} 
\definecolor{bottleneckcolor}{HTML}{8BAF9D}     
\definecolor{supervisioncolor}{HTML}{ef476f}    

\begin{tikzpicture}[>=Latex, line width=1.5pt, every node/.style={font=\small}]

\draw[->] (-1,0) -- (0,0);
\node[left=2pt] at (-1,0) {\textit{Input}};

\draw[fill=transformerblockcol,draw=none] (0,1.5) rectangle (1,-1.5);

\draw[->] (1,0) -- (2,0);

\draw[fill=bottleneckcolor,draw=none]
      (2,1.5)--(2,-1.5)--(2.9,-0.75)--(2.9,0.75)--cycle;
\draw[fill=bottleneckcolor,draw=none]
      (4,1.5)--(4,-1.5)--(3.1,-0.75)--(3.1,0.75)--cycle;

\def\xmin{2.9}
\def\xmax{3.1}
\def\ymin{-0.75}
\def\ymax{0.75}

\begin{scope}
  \clip (\xmin, \ymin) rectangle (\xmax, \ymax);
  \foreach \i in {-1.5,-1.4,...,2.0} {
    \draw[supervisioncolor, line width=1pt]
      ({\xmin}, {\i}) -- ({\xmax}, {\i + 0.2});
  }
\end{scope}

\node[below=2pt,text=supervisioncolor] at (3,-1.5)
      {\footnotesize\textit{Latent Space Supervision}};

\draw[->] (4,0) -- (5,0);
\draw[fill=transformerblockcol,draw=none] (5,1.5) rectangle (6,-1.5);

\draw[->, very thick]
      (4.5,0)
      -- (4.5,2.5)
      -- (-0.5,2.5)
      -- (-0.5,0);

\node[anchor=east] at (-0.7,1.5) {\footnotesize\textit{Input}};
\node[anchor=east] at (-0.7,1.25) {\footnotesize\textit{Adaptive}};
\node[anchor=east] at (-0.7,1) {\footnotesize\textit{Recurrence}};

\draw[dashed] (3,1.675) -- (3,-1.575);
\node[above] at (3,1.75) {\textit{Discrete Bottleneck}};

\draw[->] (6,0) -- (7,0);
\node[right=2pt] at (7,0) {\textit{Output}};

\begin{scope}[yshift=1cm,xshift=8.5cm]
  \draw[fill=transformerblockcol,draw=none] (0,0.25) rectangle (0.25,-0.25);
  \node[right=1pt, align=left] at (0.25,0) {\footnotesize\textit{Recurrent Transformer Block}};

  \draw[fill=bottleneckcolor,draw=none]
        (0,-0.75) -- (0,-1.25) -- (0.25,-1.125) -- (0.25,-0.875) -- cycle;
  \node[right=1pt] at (0.25,-1) {\footnotesize\textit{Discretize}};

  \draw[fill=bottleneckcolor,draw=none]
        (0,-1.875) -- (0,-2.125) -- (0.25,-2.25) -- (0.25,-1.75) -- cycle;
  \node[right=1pt] at (0.25,-2) {\footnotesize\textit{Re-Embed}};
\end{scope}

\end{tikzpicture}
    }
    \caption{Overview of the proposed architecture for OOD generalization. It features a recurrent Transformer block, latent algorithmic supervision, and a discretization mechanism to anchor representations across iterations. Self-correction mechanism is not represented here. }\label{fig:architecture_diagram}
    \vskip-12pt
\end{figure}

\textbf{Mechanism 1: Recurrence \& Input-Adaptive Computation.} 
The iterative and recursive structure of the target layer-by-layer algorithm naturally motivates a \coloremph{recurrent architecture}. We employ a recurrent Transformer block \citep{dehghaniUniversalTransformers2019} with the goal that each application emulates one algorithmic iteration—that is, computing values for one additional layer of the computation graph. 
An input instance is represented as a sequence of $n$ tokens $X = (x_1, \ldots, x_n)$, as described in~\eqref{eq:token_sequence}. This is embedded to form a sequence of embedding vectors $E_1^{(0)}, \ldots, E_n^{(0)}$, and recurrently processed with the recurrent transformer block
\begin{empheq}[box=\fcolorbox{MorandiPink}{MorandiPink!20}]{equation}
\label{eq:recurrent_block}
    \begin{aligned}
        (E_1^{(t+1)}, \ldots, E_n^{(t+1)}) &\gets \mathrm{RecurrentTransformerBlock}(E_1^{(t)}, \ldots, E_n^{(t)}), \ t = 1, 2, \ldots, T.
    \end{aligned}
\end{empheq}
The output is linearly read out from the final embedding states $E_1^{(T)}, \ldots, E_n^{(T)}$.
Crucially, the number of recurrent iterations, $T$, is not fixed but \coloremph{adapts to input complexity}, scaling linearly with the depth of the computation graph. This input-adaptive recurrence allows the model to dynamically scale its computation time proportionate to the problem's requirements, a key capability for OOD generalization to larger graphs. Unlike CoT methods that scale computation by generating progressively longer linear sequences of tokens, recurrence introduces inductive biases favoring recursive solution structures, which are inherently more scalable. 
This recurrent structure also provides key \emph{computational} advantages compared to autoregressive chain-of-thought methods: in our recurrent architecture, each step can perform parallel processing across the \emph{entire context} instead of being constrained to perform computation sequentially token-by-token, yielding more efficient use of working memory since the full computational trace is not serially materialized. 
The use of recurrence to adaptively scale computation time is a well-established concept for tackling tasks with variable complexity~\citep{gravesAdaptiveComputationTime2017,dehghaniUniversalTransformers2019,baninoPonderNetLearningPonder2021a,schwarzschildCanYouLearn2021,bansalEndtoendAlgorithmSynthesis2022,fanLoopedTransformersLength2024,geipingScalingTestTimeCompute2025}. 

\textbf{Mechanism 2: Latent State Algorithmic Supervision.} While recurrence (Mechanism 1) provides the capacity for iterative computation, it does not inherently guarantee that the model will learn the desired layer-by-layer algorithmic procedure. 
To instill this structure, we introduce \coloremph{latent state algorithmic supervision}. 
Unlike CoT, which supervises intermediate computation in token space, our mechanism provides supervision directly within the model's latent representation space at each recurrent step, steering the internal states to align with the step-by-step execution of our target algorithm. 
Specifically, at each recurrent iteration $t$, a shared \emph{linear readout layer} is used to predict node values from their current latent embeddings $E_i^{(t)}$. The training loss applied to these predictions at each recurrent iteration is designed to align the model with the target layer-by-layer algorithm. In particular, for each iteration $t$, it penalizes errors in the predicted values for nodes that are algorithmically computable within $t$ processing steps (i.e., of depth $t$ or less) as follows
\begin{empheq}[box=\fcolorbox{MorandiPink}{MorandiPink!20}]{equation}
\begin{aligned}
    \text{AlgorithmAlignmentLoss} = \sum_{t=1}^{T} \sum_{\substack{i \in [n]}} \Ind{\mathrm{Depth}(x_i) \leq t} \cdot \ell\paren{W_{\mathtt{value}} \,E_i^{(t)}, \mathrm{Value}(x_i)},
\end{aligned}
\end{empheq}
where $\mathrm{Depth}(x_i)$ is the node's depth in the computation graph, $\mathrm{Value}(x_i)$ is its ground-truth value, and $\ell$ is the cross-entropy loss. 
Thus, the algorithm alignment loss supervises the model such that at iteration $t$, it computes the values of all nodes in the input at computational depth less than or equal to $t$.
For example, in \Cref{fig:dag_example}, supervision at $t=1$ applies to leaf nodes (e.g., $x_7$), while at $t=2$ it extends to include second-layer nodes (e.g., $x_{23}$), and so on.
This iterative supervision encourages the model to progressively build up the solution, computing the graph one effective layer deeper with each recurrent step.

\textbf{Mechanism 3: Anchoring Latent Representation via Discretization.}
Recurrent models can suffer from representational drift across recurrent iterations during extended out-of-distribution computation, arising from error accumulation when computation scales beyond the training regime. To mitigate this and ensure stable processing across many iterations, we introduce a \coloremph{discretization mechanism} that \emph{anchors} the model's latent representation while scaling computation through recurrence. Specifically, after each iteration, the model's continuous hidden states are projected into a structured, discrete symbolic space and then immediately re-embedded to form the input for the next recurrent step. This forces the intermediate representations at each iteration to begin and end in a shared structured space, thereby maintaining semantic stability even when computation extends beyond the training regime. 
Ultimately, this anchoring constrains the model to learn a \emph{depth-invariant} computational process, which is key to generalizing to longer computational depths than seen during training.

We implement this anchoring using a structured tokenization and embedding scheme, enabling each token’s internal state to evolve recurrently while remaining grounded in a shared discrete space.
In our task of \emph{modular arithmetic on computational graphs}, the discrete latent space is structured as a product of four factors: token syntax, variable identity, numerical value, and operation type. To illustrate the structure of the discrete space, consider the input sequence ``$\token{17} \token{=} \token{x_{42}} \sep$''. This sequence is tokenized into symbolic factors as follows:
\begin{center}
\begin{tabular}{ccrccccl}
     &  & & \texttt{syntax} & \texttt{variable} & \texttt{operation} & \texttt{value} & \\[8pt]
    $\token{17}$ & $\to$ & $[$ & $\mathtt{value}$ & $\mathtt{N/A}$ & $\mathtt{N/A}$ & $17$ & $]$\\
    $\token{=}$ & $\to$ & $[$ & $\token{=}$ & $\mathtt{N/A}$ & $\mathtt{N/A}$ & $\mathtt{N/A}$ & $]$\\
     $\token{x_{42}}$& $\to$ & $[$ & $\mathtt{variable}$ & $x_{42}$ & $\mathtt{N/A}$ & $\mathtt{empty}$ & $]$\\
    $\sep$ & $\to$ & $[$ & $\sep$ & $\mathtt{N/A}$ & $\mathtt{N/A}$ & $\mathtt{N/A}$ & $]$\\
\end{tabular}
\end{center}
Note that the \texttt{value} factor of \textit{variable} tokens (e.g., $\token{x_{42}}$ above) is \texttt{empty} at the input layer. As the model processes the input recurrently, it iteratively computes the values of different variables, updating the \texttt{value} factor of the discrete latent state. This yields a latent representation that is discrete, shared across steps, and scalable to extended computation. To map the discrete states to distributed embeddings, we train a separate embedding layer for each factor and combine the factor embeddings by summation.

At each iteration, we first apply the $\mathrm{RecurrentTransformerBlock}$, as in~\Cref{eq:recurrent_block}, forming the core computation of the recurrent step. The processed distributed representations are then discretized via argmax decoding across each symbolic factor, projecting the latent representation to a common structured space. We then re-embed the discrete state to form the vectorized input for the next iteration.
\begin{empheq}[box=\fcolorbox{MorandiPink}{MorandiPink!20}]{equation}
    \begin{aligned}
        (\tilde{E}_1^{(t+1)}, \ldots, \tilde{E}_n^{(t+1)}) &\gets \mathrm{RecurrentTransformerBlock}(E_1^{(t)}, \ldots, E_n^{(t)})\\
        z_{i, \mathtt{factor}}^{(t+1)} &\gets \argmax \sset{W_{\mathtt{factor}} \, \tilde{E}_i^{(t+1)} }\hphantom{~} \quad \mathtt{factor} \in \sset{ \mathtt{syntax}, \mathtt{variable}, \mathtt{operation}, \mathtt{value}} \\
        E_{i, \mathtt{factor}}^{(t+1)} &\gets \mathrm{FactorEmbed}(z_{i, \mathtt{factor}}^{(t+1)}) \quad \mathtt{factor} \in \sset{ \mathtt{syntax}, \mathtt{variable}, \mathtt{operation}, \mathtt{value}} \\
        E_i^{(t+1)} &\gets E_{i, \mathtt{syntax}}^{(t+1)} + E_{i, \mathtt{variable}}^{(t+1)} + E_{i, \mathtt{operation}}^{(t+1)} + E_{i, \mathtt{value}}^{(t+1)}.
    \end{aligned}
\end{empheq}

\textbf{Mechanism 4: Learning to Self-Correct.} 
Finally, to enhance the robustness of the learned algorithm, especially as the number of computational steps increases and makes the process more susceptible to error propagation, we introduce a  \coloremph{self-correction scheme}.  This mechanism aims to equip the model with the ability to recover from such intermediate mistakes. To facilitate this robustness, we train the model by intentionally introducing errors into its reasoning process. Specifically, at each recurrent iteration, with a small probability, we randomly corrupt a selection of the value components within the model's discrete latent states. This training regimen forces the model to learn to detect when a previously computed value is incorrect (due to our induced corruption or its own misstep) and then to correct this error in a subsequent computational step before proceeding with the task.

\subsection{Experimental Results \& Discussion}

Combining these mechanisms yields an architecture capable of effectively generalizing far beyond the training distribution to much larger and more complex inputs. To evaluate the effects of the different mechanisms we propose, we study a collection of methods, each implementing a different subset of these mechanisms.

These methods are listed in~\Cref{tab:method_guide}. The \methodcolorhighlight{feedforwardcolor}{Feedforward End-to-End} method does not implement any of the proposed mechanisms. The \methodcolorhighlight{recurrentcolor}{Recurrent End-to-End} method partially implements \emph{Mechanism 1} as it uses recurrence but lacks input-adaptive computation. The \methodcolorhighlight{cotsupervisioncolor}{Chain-of-Thought} method partially implements \emph{Mechanism 1} since the length of the chain-of-thought trajectory scales with the complexity of the problem. It also partially implements \emph{Mechanism 2} because the next-token prediction objective on the chain-of-thought sequences provides supervision on the intermediate steps, although this supervision is not directly applied to the latent states. The \methodcolorhighlight{continuoussupervisioncolor}{Continuous Latent Space Supervision} method fully implements \emph{Mechanisms 1 \& 2}. It is a recurrent model featuring input-adaptive computation and latent state algorithmic supervision. However, we omit the discretization mechanism (\emph{Mechanism 3}), thereby maintaining continuous distributed latent states. The \methodcolorhighlight{discretesupervisioncolor}{Discrete Latent Space Supervision} method incorporates the discretization mechanism, implementing \emph{Mechanisms 1, 2, \& 3}. Finally, the \methodcolorhighlight{discretesupervisionrecorrectioncolor}{Discrete Latent Space Supervision $\circlearrowleft$} method further incorporates the error correction mechanisms, thus implementing all four mechanisms.

\begin{table}[ht]
    \caption{\textit{Guide to Implementation of Proposed Mechanisms in Baselines.} The leftmost column shows the method names of the different baselines and ablations we consider, matching the figure legends. $\implemented$ indicates that a method implements the given mechanism, $\notimplemented$ indicate that the mechanism is not implemented, and $\partiallyimplemented$ indicate that it is partially implemented.}\label{tab:method_guide}
    \centering
    \resizebox{\textwidth}{!}{
    \begin{tabular}{ccccc}
        \toprule
        \textit{Method} / \textit{Mechanism} & \textit{Mechanism 1} & \textit{Mechanism 2} &  \textit{Mechanism 3} & \textit{Mechanism 4}\\\midrule
        \methodcolorhighlight{feedforwardcolor}{Feedforward End-to-End} & \notimplemented & \notimplemented & \notimplemented & \notimplemented\\
        \methodcolorhighlight{recurrentcolor}{Recurrent End-to-End} & \partiallyimplemented & \notimplemented & \notimplemented & \notimplemented\\
        \methodcolorhighlight{cotsupervisioncolor}{Chain-of-Thought} & \partiallyimplemented & \partiallyimplemented & \notimplemented & \notimplemented \\
        \methodcolorhighlight{continuoussupervisioncolor}{Continuous Latent Space Supervision} & \implemented & \implemented & \notimplemented & \notimplemented\\
        \methodcolorhighlight{discretesupervisioncolor}{Discrete Latent Space Supervision} & \implemented & \implemented & \implemented & \notimplemented\\
         \methodcolorhighlight{discretesupervisionrecorrectioncolor}{Discrete Latent Space Supervision $\circlearrowleft$} & \implemented & \implemented & \implemented & \implemented\\\bottomrule
    \end{tabular}
 }
\end{table}

\begin{figure}[ht]
    \begin{minipage}{0.52\textwidth}
        \centering
        \includegraphics[width=\linewidth]{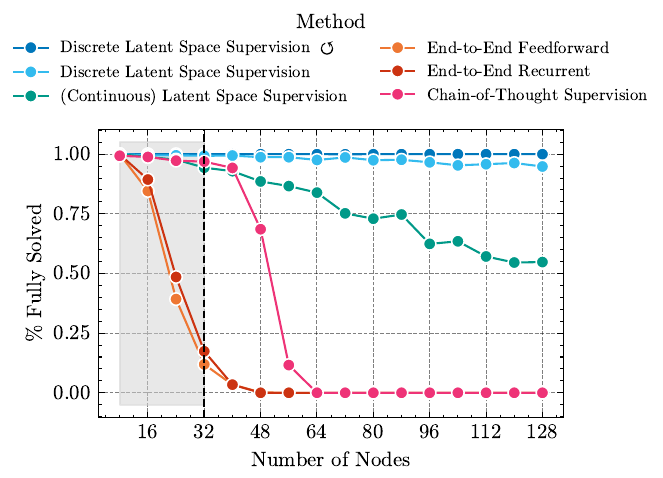}
        \captionof{figure}{Out-of-Distribution generalization performance of different methods on the mathematical reasoning task.}
        \label{fig:method_ood_comparison}
    \end{minipage}%
    \hfill
    \begin{minipage}{0.47\textwidth}
            \centering
            \vskip 40pt
            \includegraphics[width=\linewidth]{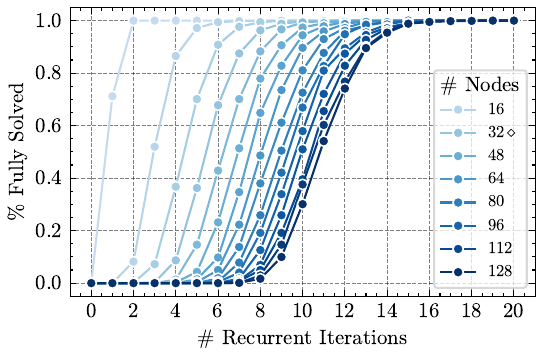}
            \captionof{figure}{Effective out-of-distribution generalization via input-adaptive scaling of computation time. This depicts \methodcolorhighlight{discretesupervisionrecorrectioncolor}{Discrete Latent Space Supervision $\circlearrowleft$}}
            \label{fig:ourmethod_stepwise_results}
    \end{minipage}
\end{figure}

\textbf{\textit{Enabling Robust Algorithmic OOD Generalization.}} \Cref{fig:method_ood_comparison} depicts the OOD generalization performance of our methods, ablating across the ingredients described above, as well as the aforementioned \methodcolorhighlight{cotsupervisioncolor}{Chain-of-Thought} and \methodcolorhighlight{feedforwardcolor}{End-to-End} baselines. As previously mentioned, we find that the \methodcolorhighlight{feedforwardcolor}{End-to-End}  models (both recurrent and feedforward) fail to effectively learn the task (with respect to our stringent ``fully solved'' metric) beyond small graph sizes, even in-distribution. The recurrent models slightly outperform the feedforward models. \methodcolorhighlight{cotsupervisioncolor}{Chain-of-Thought} supervision enables a significant improvement, yielding near-perfect performance in-distribution ($N \leq 32$), and a limited degree of out-of-distribution generalization. To assess our proposed mechanisms for robust OOD generalization in Transformers, we evaluate three classes of models incorporating different subsets of those ingredients. We find that this enables a dramatic improvement in OOD generalization, with performance improving further as more ingredients are incorporated. When all proposed ingredients are incorporated, i.e., \methodcolorhighlight{discretesupervisionrecorrectioncolor}{Discrete Latent Space Supervision $\circlearrowleft$}\footnote{Here, \methodcolorhighlight{discretesupervisionrecorrectioncolor}{$\circlearrowleft$} denotes self-correction.},  the model robustly achieves \methodcolorhighlight{discretesupervisionrecorrectioncolor}{near-perfect performance} across all OOD splits we examined.

\textbf{\textit{Depth-Invariance for Scalable Reasoning.}} Generalizing to problem instances more complex than those seen during training requires some mechanism of scaling computation proportionately. The chain-of-thought solution to this challenge is to scale the length of the autoregressively generated CoT trace, carrying out computation through the sequential generation of tokens. While this can yield some success, it is inherently limited: computation is forced into a token-by-token format rather than the model's native latent representation space, constraining efficiency and robustness. In this work, we explore a different approach based on recurrence with input-adaptive recurrent depth, introducing inductive biases that enforce a \coloremph{depth-invariant structure} in the learned solution. That is, the model learns a solution such that the computational description at every step of the solution process is the same, making it possible to scale it to depths far larger than those seen during training. This notion parallels other architectural invariances studied in geometric deep learning --— such as translation, rotation, or permutation equivariance --- where networks preserve behavior under transformations aligned with the task structure~\citep{bronstein2017geometric,bronstein2021geometricdeeplearninggrids,gerken2023geometric}. Here, the recurrence imposes invariance under the network’s own iterative action, yielding a scalable, recursive algorithm capable of solving much larger and more complex instances.

\textbf{\textit{The Importance of Anchored Discrete Representations.}} In~\Cref{fig:method_ood_comparison}, \methodcolorhighlight{continuoussupervisioncolor}{Continuous Latent Space Supervision} denotes a recurrent model where the continuous latent states receive step-by-step algorithmic supervision, but the latent states are not discretized in between recurrent block iterations as they are in \methodcolorhighlight{discretesupervisioncolor}{Discrete Latent Space Supervision}. 
We see that, while this outperforms the \methodcolorhighlight{cotsupervisioncolor}{Chain-of-Thought} baseline, which is limited to linear reasoning paths, its out-of-distribution performance slowly degrades as we test on progressively larger inputs, which require increasing recurrent depth and computation time. 
We attribute this to accumulating noise in the continuous vector representations --- a phenomenon exacerbated when scaling test-time compute for larger problem instances --- which eventually causes representations to drift from the semantically meaningful manifold learned during training. In \methodcolorhighlight{discretesupervisioncolor}{Discrete Latent Space Supervision}, the model receives step-by-step algorithmic supervision as with its continuous counterpart, but now we additionally discretize the latent representation, then re-embed using a common embedder that is shared across recurrent iterations. This has the effect of ``anchoring'' the latent states to a common, semantically-consistent representation space, allowing the model to scale up computational depth without accumulating noise. We observe that this yields significantly improved OOD generalization.
\aanote{Can this be related, intuitively, to ideas from coding theory? Channel coding, etc.}

\textbf{\textit{Error-Correction Leads to Greater Robustness in Scaling.}} 
In \methodcolorhighlight{discretesupervisionrecorrectioncolor}{Discrete Latent Space Supervision $\circlearrowleft$}, we introduce explicit supervision for error correction by randomly corrupting the model's latent space with some small probability during training. 
While the model may make occasional errors, it is able to correct them in the next recurrent iteration, thereby yielding near-perfect OOD generalization.
Interestingly, we find that \coloremph{error correction requires more layers} in the recurrent block in order to succeed. An intuitive explanation is that effective error correction requires greater computational depth \textit{per step}: the model must first identify and correct errors from prior steps before executing the current step's computation.

\textbf{\textit{Robust Test-time Scaling.}} On many tasks, the computation time required to solve a problem instance is proportional to its size or complexity. Consequently, solving problems larger than those encountered during training necessitates scaling computation time beyond the training regime. In our setting, where the model's reasoning process is latent, we achieve this by increasing the number of recurrent iterations. \Cref{fig:ourmethod_stepwise_results} depicts the proportion of input instances solved as a function of the number of recurrent iterations. Increasing the number of iterations enables solving incrementally larger and harder problem instances. Our architectural mechanisms enable this robust scaling beyond the training regime.

\textbf{\textit{Details, Extensions \& Further Ablations.}} In the appendices, we provide further discussion and present additional experimental results. Here, we briefly highlight a few aspects of these extensions. Across all methods, we find that hyperparameter choice can be critical. In particular, we find that the choice of positional encoding and model depth is especially important. In the above results, we always report the best model within each method after a hyperparameter search, the details of which are provided in the appendix. Additionally, for the chain-of-thought baselines, we explore multiple schemes for the design of the reasoning chains and present the best results here. 


Now that we have demonstrated the effectiveness of the proposed architectural mechanisms for robust OOD generalization, we next conduct a mechanistic interpretability analysis to probe the precise computational circuits learned by each component of our model.

\section{Mechanistic Interpretability}\label{sec:mechinterp}

In this section, we aim to answer the following questions via a detailed study of the model's inner workings:  
\begin{highlightbox}
\begin{center}
    \textit{(\romannumeral1)} What algorithm does the trained model implement? \\\textit{(\romannumeral2)} Why is the trained model able to generalize to OOD data?
\end{center}
\end{highlightbox}




To answer these questions, we first propose hypotheses on the functionality of each model block: first-layer attention, second-layer attention, and the final MLP. For each of these hypotheses, we 
conduct controlled experiments where we apply causal interventions to specific parts of the input and isolate the effect on model activations to identify the function of each component. 
Our methodology builds on prior work on causal interpretability in neural networks~\citep{geiger2021causal,meng2022locating,geiger2024finding}, but is tailored specifically to interpreting \emph{recurrent} transformer models. We provide complete details of our experimental methodology in the appendix.

\subsection*{Induction Head \& Modular Addition Mechanism}
To understand the algorithm implemented by the trained model, we analyze in detail the recurrent Transformer model trained with our proposed \methodcolorhighlight{discretesupervisioncolor}{Discrete Latent Space Supervision} method on the mathematical reasoning task.
The recurrent Transformer model is configured with two layers, 16 attention heads, and a hidden state dimension of 256. For more details on the model configuration, please refer to \Cref{sec:mechinterp_details}.
We summarize our mechanism analysis results in \Cref{fig:2_layer_model}, where we reveal an \coloremph{induction head} mechanism operating within the two-layer attention block and a  \coloremph{modular addition} mechanism in the final feedforward layer.
To better understand the model's behavior, let us take an example equation in the following format:
\begin{equation*}
    \sep \: \VAR{0} \: \token{+} \:\VAR{1} \token{+} \: \VAR{2} \: \token{=} \: \RHS.
\end{equation*}
We can break down the model's computation into three main components at the Right-Hand Side (RHS) position:
\begin{itemize}
    \item The first layer attention heads copy the ``$\mathtt{variable}$'' factored embeddings of variables $\VAR{0}$, $\VAR{1}$, and $\VAR{2}$ to the RHS position, which let the model know the variable names at the RHS position.
    \item The second layer attention heads use the copied variable names to retrieve the computed values of variables $\VAR{0}$, $\VAR{1}$, and $\VAR{2}$ from the previous equations through an induction-head mechanism.
    \item The last feedforward layer computes the sum of the values of the variables on the LHS and outputs the result to the RHS position.
\end{itemize}
\begin{figure}[h]
    \centering
    \includegraphics[width=0.8\textwidth]{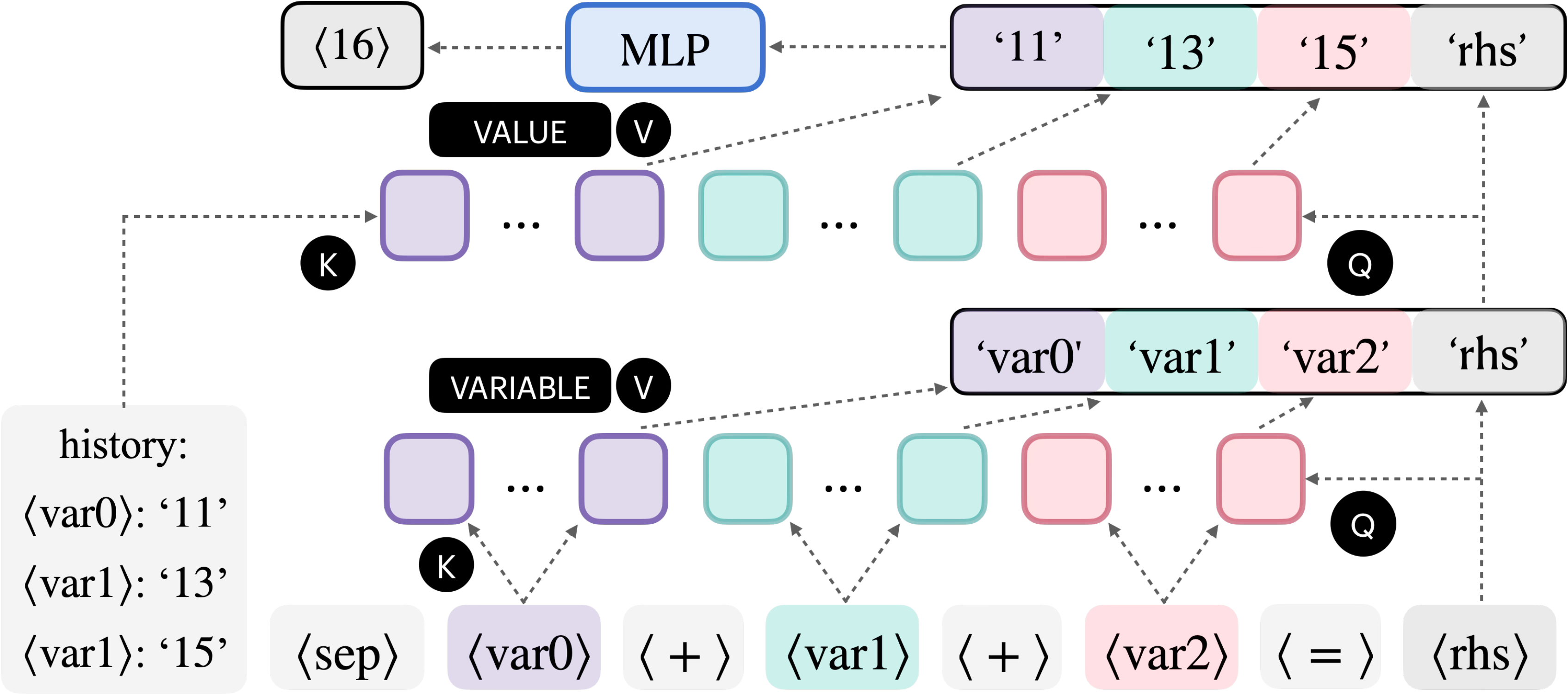}
    \caption{Illustration of the two-layer model performing the modular addition task. The colored squares represent attention heads, grouped by the variable positions they attend to. Black rectangles indicate the embedding components chosen by the value projection matrix. $\langle\cdot\rangle$ denotes tokens, and `$\, \cdot \,$' denotes embedding components.}
    \label{fig:2_layer_model}

\end{figure}

\textbf{First Layer Attention Performs Variable Copying.} The attention heads in the first layer are grouped by the variable position they attend to, reflecting an attention pattern that is dependent on  \coloremph{relative position}, as illustrated in \Cref{fig:l0} (left). For the token embeddings of $\VAR{0}$, $\VAR{1}$, and $\VAR{2}$, which comprise four separate factored embedding types (\texttt{syntax}, \texttt{variable}, \texttt{operation}, and \texttt{value}), the value and output projection matrices of each head group \emph{select a subspace} of these token embeddings containing only the \texttt{variable} embeddings. 
This is evident in \Cref{fig:l0} (right), which plots the norm amplification for different factored embedding types. 
More details on the norm amplification calculation can be found in \Cref{sec:mechinterp_details}.

This shows that the \coloremph{first layer attention copies the \textit{variable names} of its parents}, which will later be used to obtain their values in the second layer.

\textbf{Second Layer Attention Implements Variable-Dependent Induction Head Mechanism.} 
The second layer's attention heads then retrieve the corresponding values of variables $\VAR{0}$, $\VAR{1}$, and $\VAR{2}$ from the previous equations through an induction-head mechanism~\citep{olsson2022context}.
Specifically, all the attention heads are also grouped by which variable value they are retrieving.
For example, let us suppose that the first head group is responsible for retrieving the value of $\VAR{0}$. 
Then, the attention heads within this group will find the first occurrence of $\VAR{0}$, which will be the RHS of some previous equation.  
This particular position is the first time the value of $\VAR{0}$ is computed. 
And these attention heads will then copy the ``$\mathtt{value}$'' factored embedding of $\VAR{0}$ also to the current RHS position.
In summary, the variable names copied in the first layer are used as queries to \coloremph{retrieve these variables' values}, searching over the RHS of previous equations.


\textbf{Feedforward Layer Performs Modular Addition.} The second layer MLP implements a sophisticated \coloremph{modular addition} mechanism that computes the sum of the three variable values modulo 23. The MLP receives as input the sum of three transformed value embeddings from the attention layer --- one for each variable position. These embeddings exhibit a periodic structure that naturally lends itself to frequency domain analysis.

Through systematic experimentation where we vary all three input values from 0 to 22 and apply three-dimensional Discrete Fourier Transform (DFT) analysis, we observe a fascinating computational pattern. At the MLP's pre-activation stage, the representation is dominated by a bias term (the $(0,0,0)$ frequency component). As signals propagate through the MLP layers, this bias progressively diminishes while diagonal frequency components of the form $(a,a,a)$ are amplified, where $a \in \{1, \ldots, 22\}$. 
That is, the Fourier components where $\VAR{0}$, $\VAR{1}$, and $\VAR{2}$ have the same frequency are amplified.
These diagonal frequencies encode precisely the information needed for modular arithmetic: they represent sinusoidal functions of the sum $x+y+z$.

The MLP essentially performs the computation through combinations of terms like $\cos(2\pi a(x+y+z)/23)$, where the periodic nature of trigonometric functions naturally handles the modulo operation. This frequency-based approach aligns with recent findings on how neural networks implement modular arithmetic~\citep{nandaprogress, tian2024composing, doshi2024grok}. We provide detailed experimental evidence and visualizations of this mechanism in \Cref{sec:mechinterp_details}, including DFT analysis at multiple network positions showing the progressive amplification of sum-encoding frequencies. 

\begin{figure}[h]
\centering
\begin{subfigure}[b]{0.78\textwidth}
    \centering
    \includegraphics[width=\textwidth]{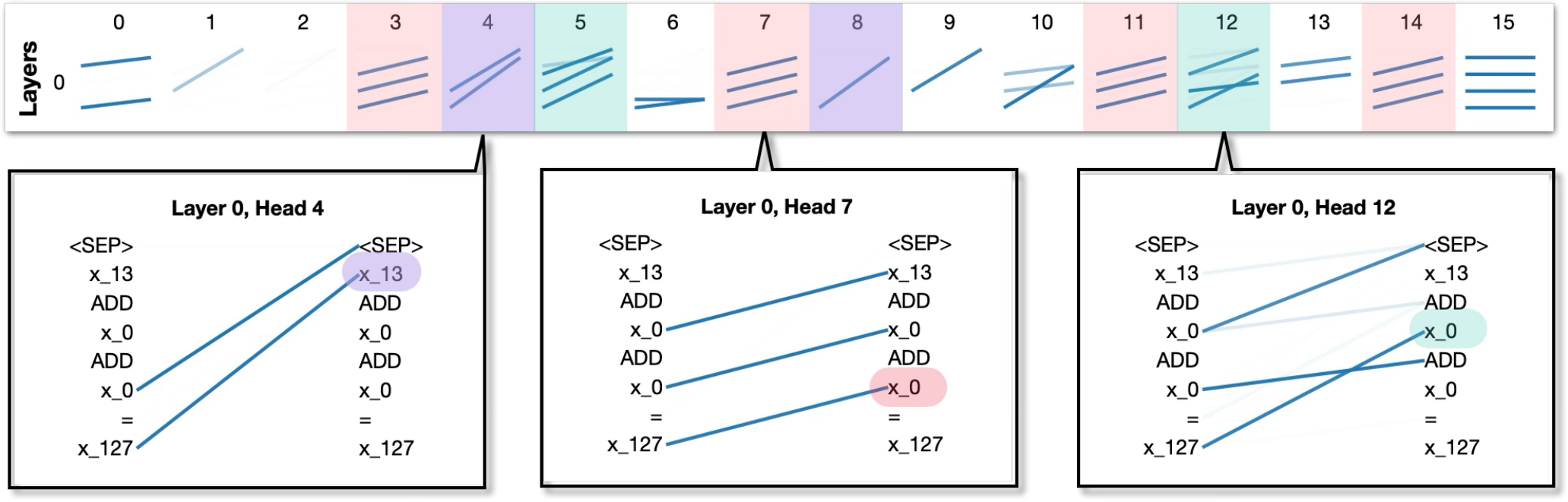}
\end{subfigure}
\hfill
\begin{subfigure}[t]{0.20\textwidth}
    \centering
    \includegraphics[width=\textwidth]{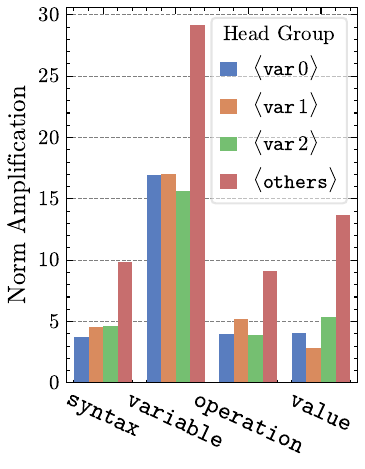}
\end{subfigure}
\caption{\textbf{Left.} An illustration of the functionality of attention heads by groups in the first attention layer. Head 4 and 8 attend to the first variable position, Head 5 and 12 attend to the second variable position, Head 3, 7, 11, 14 attend to the third variable position, and the remaining heads attend to the RHS position or do not show a clear attention pattern. \textbf{Right.} Norm amplification of each factor's embeddings passed through the combined attention OV matrix by head groups. $\token{\mathtt{others}}$ exhibits significantly higher norm amplification, primarily because head 15 performs a self-copy operation at the RHS position.}
\label{fig:l0}
\end{figure}

\textbf{\textit{OOD Generalization of the Trained Model.}}
The model's robust OOD generalization can be traced back to the architectural mechanisms of \methodcolorhighlight{discretesupervisioncolor}{Discrete Latent Space Supervision} guiding the model towards learning a universal and robust algorithm. In particular, the algorithm implements a variable-dependent induction head mechanism that is invariant to length, leveraging both relative-positional and variable-dependent attention patterns, which enables the model to operate over contexts of arbitrary lengths.
Thus, despite being trained on graphs with limited size, the input-adaptive recurrence, intermediate supervision, and discretization mechanisms enable the model to \coloremph{learn a scalable algorithm} capable of solving problems of increased complexity.

\section{Conclusion}\label{sec:discussion}
This work investigated \emph{algorithmic generalization} in Transformers for scalable mathematical reasoning, a domain where standard chain-of-thought approaches fail on out-of-distribution inputs. We introduced a novel architecture integrating input-adaptive recurrence, latent algorithmic supervision, state discretization, and self-correction mechanisms. Collectively, these mechanisms enabled our models to achieve near-perfect OOD performance by facilitating robust, scalable reasoning directly within their internal latent representations, overcoming the brittleness of sequential token-based methods. Mechanistic interpretability further illuminated how these components achieve systematic generalization. 
While our synthetic mathematical reasoning task offers analytical clarity for investigating fundamental principles---such as adaptive recurrence and discrete latent bottlenecks---future work should explore extending these principles to more diverse, less-structured, and multi-task settings.

    \newpage
    \printbibliography

    \clearpage
    \appendix

\clearpage
\section{Experimental Details on Chain-of-Thought \& End-to-End Baselines}\label{sec:cot_details}

This section provides further experimental details on the chain-of-thought and end-to-end baselines.

\subsection{End-to-End Baselines}

The end-to-end models in our experiments are causal encoder-only Transformer models with a fixed depth and/or number of iterations that are trained with end-to-end supervision only. That is, they receive supervision on the final solution, but do not receive fine-grained supervision on the intermediate steps to explicitly align the models to a universal algorithmic problem-solving procedure.

Within the end-to-end baselines, we consider feedforward models and recurrent models. Feedforward models have a fixed number of layers and independently-learned parameters at each layer. Recurrent models, on the other hand, have a recurrent block consisting of some number of Transformer layers, which is applied recurrently for a fixed number of iterations. 

Recognizing the importance of positional encoding for length generalization~\citep{kazemnejadImpactPositionalEncoding2023}, we explore several positional encoding methods for each class of methods that we evaluate. In particular, we evaluate learned absolute positional embeddings~\citep{vaswani2017attention} (AbPE), Rotary Positional Encoding~\citep{su2023roformerenhancedtransformerrotary} (RoPE), No Positional Encoding~\citep{kazemnejadImpactPositionalEncoding2023} (NoPE), and the relative positional-encoding method proposed by~\citep{he2021debertadecodingenhancedbertdisentangled} (DeBERTa).

We perform a hyperparameter search across each of these factors, varying the number of recurrent iterations $T$, the number of layers per recurrent block $L$, the hidden state dimension $D$, and the positional encoding method. As described in the main text, we train on a dataset of examples with up to $32$ nodes, and evaluate on examples varying in size from $8$ nodes to $128$ nodes. ~\Cref{fig:baselines_all_models_fullseqacc} depicts the average OOD performance as measured by the ``\% Fully Solved'' metric for each baseline model configuration. The results in the main text correspond to the best-performing end-to-end models according to this metric. In particular, the best-performing recurrent model is RoPE-T4L2H16D256, and the best-performing feedforward model is DeBERTa-T1L8H16D256. Note that the naming scheme describes the positional encoding method, the number of recurrent steps $T$, the number of layers $L$ in the Transformer block, the number of attention heads $H$, and the model dimension $D$. $T=1$ corresponds to a ``feedforward'' model with no recurrence.

\Cref{fig:baseline_figures} depicts additional experimental results for the end-to-end baseline experiments.

\begin{figure}[h]
    \centering
    \includegraphics{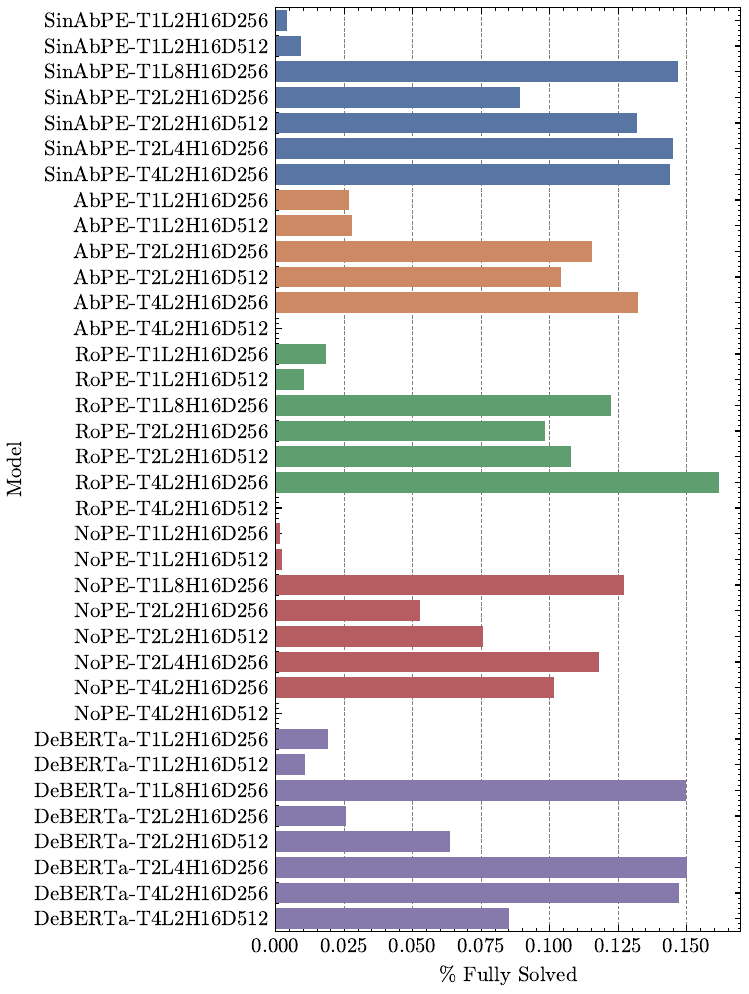}
    \caption{A comparison of average OOD generalization performance of different feedforward and recurrent baselines, varying architectural hyperparameters. This is computed as the average of the ``\% Fully Solved'' metric computed on inputs of varying size from $N=8$ to $N=128$.}
    \label{fig:baselines_all_models_fullseqacc}
\end{figure}

\begin{figure}[h]
    \centering
    \begin{subfigure}[t]{0.45\linewidth}
        \centering
        \includegraphics[width=\linewidth]{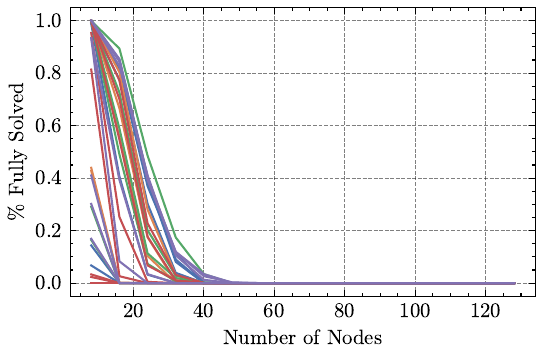}
        \caption{Each line corresponds to an experimental run. Lines are color-coded by positional encoding, but other architectural hyperparameters vary and are not represented.}
    \end{subfigure}
    \hfill
    \begin{subfigure}[t]{0.45\linewidth}
        \centering
        \includegraphics[width=\linewidth]{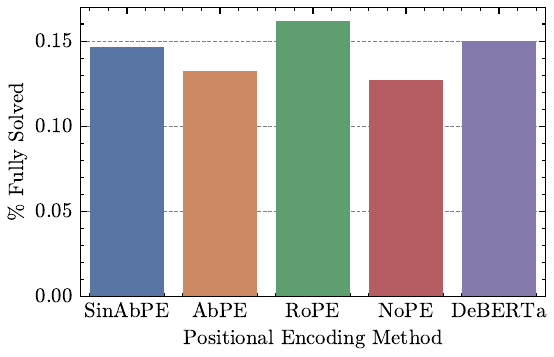}
        \caption{Average ``\% Fully Solved'' across test splits for the best model of each positional encoding method. The relative positional encoding methods, RoPE and DeBERTa perform best.}
    \end{subfigure}

    \begin{subfigure}[t]{0.45\linewidth}
        \centering
        \includegraphics[width=\linewidth]{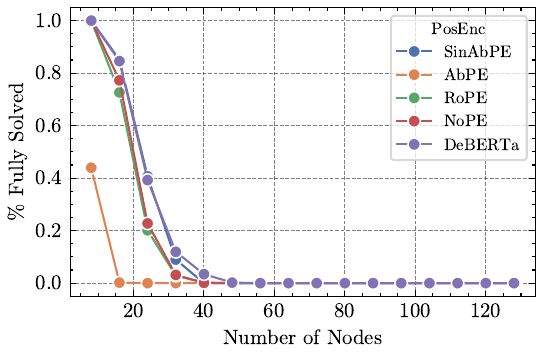}
        \caption{\% Fully solved by graph size for best model of each positional encoding method in the \textit{feedforward} baselines.}
    \end{subfigure}
    \hfill
    \begin{subfigure}[t]{0.45\linewidth}
        \centering
        \includegraphics[width=\linewidth]{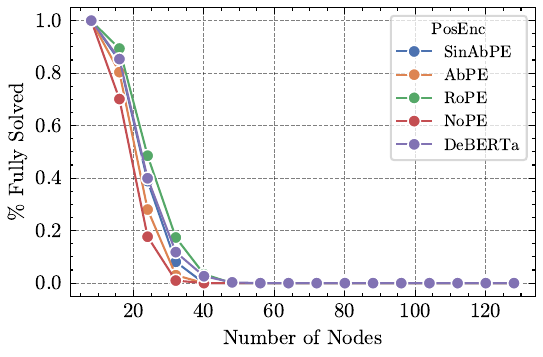}
        \caption{\% Fully solved by graph size for best model of each positional encoding method in the \textit{recurrent} baselines.}
    \end{subfigure}

    \begin{subfigure}[t]{0.45\linewidth}
        \centering
        \includegraphics[width=\linewidth]{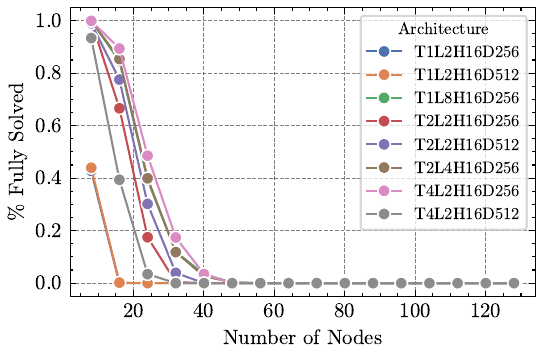}
        \caption{\% Fully solved by graph size for the best model of each architectural configuration.  Recurrent models slightly outperform feedforward models. Computational depth (i.e., $T \cdot L$) is crucial, with shallow models performing poorly even on the smallest in-distribution inputs.}
    \end{subfigure}
    \hfill
    \begin{subfigure}[t]{0.45\linewidth}
        \centering
        \includegraphics[width=\linewidth]{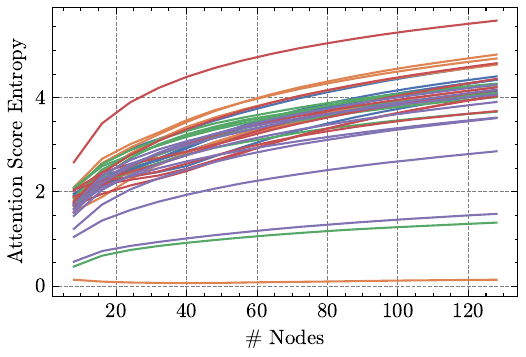}
        \caption{Average attention score entropy by input size. Attention scores disperse as the input size increases.}
    \end{subfigure}
    \caption{Further experimental results for end-to-end baselines. All end-to-end models struggle to generalize beyond the training distribution, regardless of architectural hyperparameters.}
    \label{fig:baseline_figures}
\end{figure}

\clearpage

\subsection{Chain-of-Thought Baselines} \label{subsec:cot}

The chain-of-thought baselines in our experiments are causal Transformer language models that are trained with a next-token prediction objective on sequence data that includes a step-by-step solution of the problem instance. The models are evaluated by prompting them with the problem instance and autoregressively generating the entire chain-of-thought via a greedy decoding procedure. 

We begin by providing more details on the construction of the chain-of-thought trajectories for these baselines, then provide further details on the experimental setup and present additional results.

\subsubsection{Chain-of-Thought Trajectories}

We experiment with a few different types of chain-of-thought trajectories, providing different levels and styles of supervision on the intermediate computation.

As described in the main text, the first part of the sequence is always the description of the input problem, which matches the format of the other methods we consider: a sequence of equations that define a computational graph to be solved. This is then followed by a special $\token{\mathtt{CoT}}$ token which indicates the end of the input and the beginning of the chain-of-thought. The chain-of-thought involves solving each variable in the input in linear order, one-by-one.

We experiment with two types of CoT trajectories that vary the level of detail. The first provides supervision on the values only. The CoT simply recalls that variables one-by-one and computes their values, without recalling the equation that defined them.
\begin{equation*}
    {\color{CadetBlue}\mathtt{[...Input\, Prompt...]}} \token{\mathtt{CoT}} {\color{CadetBlue}\mathtt{[...]}} \vartoken{x_{101}} \token{=} \valtoken{4}
\end{equation*}

The second type of CoT trajectory involves first recalling the equation that defined the variable, then recalling the values of the variables in the equation, and then computing the value of the desired variable. This requires a longer chain-of-thought but provides richer supervision.
\begin{equation*}
    {\color{CadetBlue}\mathtt{[...Input\, Prompt...]}} \token{\mathtt{CoT}} {\color{CadetBlue}\mathtt{[...]}} \vartoken{x_{101}} \token{=} \vartoken{x_{23}} \opertoken{+} \vartoken{x_{91}} \token{=} \valtoken{22} \opertoken{+} \valtoken{5} \token{=} \valtoken{4}
\end{equation*}

Below, we provide an example of a \textit{full} CoT trajectory on an input with $N = 32$ nodes.
\aawarning{Add corresponding graph for this CoT Trajectory example.}
{\small
\begin{empheq} 
    [box=\fcolorbox{MorandiPink}{MorandiPink!15}]{equation*}\label{eq:full_cot_example}
    \begin{aligned}
        &\valtoken{2}\token{=}\vartoken{x_{3}}\sep
\valtoken{2}\token{=}\vartoken{x_{30}}\sep
\valtoken{18}\token{=}\vartoken{x_{12}}\sep
\valtoken{14}\token{=}\vartoken{x_{11}}\sep
\\ 
&\valtoken{15}\token{=}\vartoken{x_{20}}\sep
\valtoken{8}\token{=}\vartoken{x_{23}}\sep
\vartoken{x_{30}}\token{=}\vartoken{x_{9}}\sep
\vartoken{x_{23}}\opertoken{+}\vartoken{x_{3}}\token{=}\vartoken{x_{22}}\sep
\\ 
&\vartoken{x_{20}}\opertoken{\times}\vartoken{x_{23}}\token{=}\vartoken{x_{27}}\sep
\vartoken{x_{3}}\opertoken{+}\vartoken{x_{22}}\token{=}\vartoken{x_{0}}\sep
\vartoken{x_{3}}\opertoken{+}\vartoken{x_{22}}\opertoken{\times}\vartoken{x_{11}}\token{=}\vartoken{x_{26}}\sep
\\ 
&\vartoken{x_{20}}\opertoken{-}\vartoken{x_{22}}\opertoken{+}\vartoken{x_{23}}\token{=}\vartoken{x_{13}}\sep
\vartoken{x_{22}}\token{=}\vartoken{x_{24}}\sep
\vartoken{x_{12}}\opertoken{\times}\vartoken{x_{23}}\opertoken{-}\vartoken{x_{0}}\token{=}\vartoken{x_{17}}\sep
\\ 
&\vartoken{x_{11}}\opertoken{\times}\vartoken{x_{26}}\token{=}\vartoken{x_{28}}\sep
\vartoken{x_{13}}\opertoken{-}\vartoken{x_{11}}\opertoken{+}\vartoken{x_{23}}\token{=}\vartoken{x_{21}}\sep
\vartoken{x_{17}}\opertoken{-}\vartoken{x_{3}}\token{=}\vartoken{x_{25}}\sep
\\ 
&\vartoken{x_{30}}\opertoken{\times}\vartoken{x_{17}}\opertoken{-}\vartoken{x_{23}}\token{=}\vartoken{x_{6}}\sep
\vartoken{x_{17}}\token{=}\vartoken{x_{16}}\sep
\vartoken{x_{11}}\opertoken{+}\vartoken{x_{21}}\token{=}\vartoken{x_{7}}\sep
\\ 
&\vartoken{x_{28}}\opertoken{+}\vartoken{x_{17}}\opertoken{-}\vartoken{x_{21}}\token{=}\vartoken{x_{14}}\sep
\vartoken{x_{7}}\token{=}\vartoken{x_{15}}\sep
\vartoken{x_{7}}\token{=}\vartoken{x_{31}}\sep
\\ 
&\vartoken{x_{12}}\opertoken{+}\vartoken{x_{3}}\opertoken{+}\vartoken{x_{14}}\token{=}\vartoken{x_{5}}\sep
\vartoken{x_{14}}\token{=}\vartoken{x_{19}}\sep
\vartoken{x_{23}}\opertoken{-}\vartoken{x_{5}}\opertoken{\times}\vartoken{x_{7}}\token{=}\vartoken{x_{29}}\sep
\\ 
&\vartoken{x_{5}}\token{=}\vartoken{x_{18}}\sep
\vartoken{x_{25}}\opertoken{+}\vartoken{x_{23}}\opertoken{-}\vartoken{x_{19}}\token{=}\vartoken{x_{4}}\sep
\vartoken{x_{14}}\opertoken{\times}\vartoken{x_{29}}\opertoken{-}\vartoken{x_{5}}\token{=}\vartoken{x_{2}}\sep
\\ 
&\vartoken{x_{29}}\opertoken{\times}\vartoken{x_{28}}\opertoken{-}\vartoken{x_{7}}\token{=}\vartoken{x_{1}}\sep
\vartoken{x_{3}}\opertoken{\times}\vartoken{x_{23}}\opertoken{\times}\vartoken{x_{18}}\token{=}\vartoken{x_{8}}\sep
\\ 
&\vartoken{x_{8}}\opertoken{-}\vartoken{x_{28}}\opertoken{-}\vartoken{x_{0}}\token{=}\vartoken{x_{10}} {\color{Purple} \,\token{\bm{\mathtt{CoT}}}}\, \vartoken{x_{3}}\token{=}\valtoken{2}\sep
\vartoken{x_{30}}\token{=}\valtoken{2}\sep
\\ 
&\vartoken{x_{12}}\token{=}\valtoken{18}\sep
\vartoken{x_{11}}\token{=}\valtoken{14}\sep
\vartoken{x_{20}}\token{=}\valtoken{15}\sep
\vartoken{x_{23}}\token{=}\valtoken{8}\sep
\\ 
&\vartoken{x_{9}}\token{=}\vartoken{x_{30}}\token{=}\valtoken{2}\sep
\vartoken{x_{22}}\token{=}\vartoken{x_{23}}\opertoken{+}\vartoken{x_{3}}\token{=}\valtoken{10}\sep
\vartoken{x_{27}}\token{=}\vartoken{x_{20}}\opertoken{\times}\vartoken{x_{23}}\token{=}\valtoken{5}\sep
\\ 
&\vartoken{x_{0}}\token{=}\vartoken{x_{3}}\opertoken{+}\vartoken{x_{22}}\token{=}\valtoken{12}\sep
\vartoken{x_{26}}\token{=}\vartoken{x_{3}}\opertoken{+}\vartoken{x_{22}}\opertoken{\times}\vartoken{x_{11}}\token{=}\valtoken{7}\sep
\\ 
&\vartoken{x_{13}}\token{=}\vartoken{x_{20}}\opertoken{-}\vartoken{x_{22}}\opertoken{+}\vartoken{x_{23}}\token{=}\valtoken{13}\sep
\vartoken{x_{24}}\token{=}\vartoken{x_{22}}\token{=}\valtoken{10}\sep
\\ 
&\vartoken{x_{17}}\token{=}\vartoken{x_{12}}\opertoken{\times}\vartoken{x_{23}}\opertoken{-}\vartoken{x_{0}}\token{=}\valtoken{17}\sep
\vartoken{x_{28}}\token{=}\vartoken{x_{11}}\opertoken{\times}\vartoken{x_{26}}\token{=}\valtoken{6}\sep
\\ 
&\vartoken{x_{21}}\token{=}\vartoken{x_{13}}\opertoken{-}\vartoken{x_{11}}\opertoken{+}\vartoken{x_{23}}\token{=}\valtoken{7}\sep
\vartoken{x_{25}}\token{=}\vartoken{x_{17}}\opertoken{-}\vartoken{x_{3}}\token{=}\valtoken{15}\sep
\\ 
&\vartoken{x_{6}}\token{=}\vartoken{x_{30}}\opertoken{\times}\vartoken{x_{17}}\opertoken{-}\vartoken{x_{23}}\token{=}\valtoken{3}\sep
\vartoken{x_{16}}\token{=}\vartoken{x_{17}}\token{=}\valtoken{17}\sep
\\ 
&\vartoken{x_{7}}\token{=}\vartoken{x_{11}}\opertoken{+}\vartoken{x_{21}}\token{=}\valtoken{21}\sep
\vartoken{x_{14}}\token{=}\vartoken{x_{28}}\opertoken{+}\vartoken{x_{17}}\opertoken{-}\vartoken{x_{21}}\token{=}\valtoken{16}\sep
\\ 
&\vartoken{x_{15}}\token{=}\vartoken{x_{7}}\token{=}\valtoken{21}\sep
\vartoken{x_{31}}\token{=}\vartoken{x_{7}}\token{=}\valtoken{21}\sep
\vartoken{x_{5}}\token{=}\vartoken{x_{12}}\opertoken{+}\vartoken{x_{3}}\opertoken{+}\vartoken{x_{14}}\token{=}\valtoken{13}\sep
\\ 
&\vartoken{x_{19}}\token{=}\vartoken{x_{14}}\token{=}\valtoken{16}\sep
\vartoken{x_{29}}\token{=}\vartoken{x_{23}}\opertoken{-}\vartoken{x_{5}}\opertoken{\times}\vartoken{x_{7}}\token{=}\valtoken{10}\sep
\\ 
&\vartoken{x_{18}}\token{=}\vartoken{x_{5}}\token{=}\valtoken{13}\sep
\vartoken{x_{4}}\token{=}\vartoken{x_{25}}\opertoken{+}\vartoken{x_{23}}\opertoken{-}\vartoken{x_{19}}\token{=}\valtoken{7}\sep
\\ 
&\vartoken{x_{2}}\token{=}\vartoken{x_{14}}\opertoken{\times}\vartoken{x_{29}}\opertoken{-}\vartoken{x_{5}}\token{=}\valtoken{9}\sep
\vartoken{x_{1}}\token{=}\vartoken{x_{29}}\opertoken{\times}\vartoken{x_{28}}\opertoken{-}\vartoken{x_{7}}\token{=}\valtoken{16}\sep
\\ 
&\vartoken{x_{8}}\token{=}\vartoken{x_{3}}\opertoken{\times}\vartoken{x_{23}}\opertoken{\times}\vartoken{x_{18}}\token{=}\valtoken{1}\sep
\vartoken{x_{10}}\token{=}\vartoken{x_{8}}\opertoken{-}\vartoken{x_{28}}\opertoken{-}\vartoken{x_{0}}\token{=}\valtoken{6}
    \end{aligned}
\end{empheq}}


\subsubsection{Experimental Details \& Additional Results}

We perform a hyperparameter search varying: the number of recurrent iterations $T$, the number of layers per recurrent block $L$, the hidden state dimension $D$, and the positional encoding method. As described in the main text, we train on a dataset of examples with up to $32$ nodes, and evaluate on examples varying in size from $8$ nodes to $128$ nodes. ~\Cref{fig:cot_baselines_all_models_fullseqacc} depicts the average OOD performance as measured by the ``\% Fully Solved'' metric for each baseline model configuration. The results in the main text correspond to the best-performing CoT-supervised model according to this metric, which is the RoPE-T4L2H16D256 model.

\Cref{fig:cot_baseline_figures} depicts additional experimental results for the end-to-end baseline experiments. We highlight a few observations here:

\begin{highlightbox}
\begin{itemize}
    \item ~\Cref{fig:cot_baselines_all_models_varfullseqacc} shows that some models are able to recall the equation structure correctly in their CoT, but are unable to robustly compute the values correctly. This suggests that a common source of error in the CoT baselines is the arithmetic computation, rather than copying equations from the input.
    \item As with the end-to-end baselines, the positional encoding method was critical for performance and length generalization. Among the methods we evaluated, we found NoPE to perform best, generalizing well to 40 nodes when trained on $N \leq 32$ nodes. The other positional encoding methods fail to generalize beyond the training regime. No method generalized robustly beyond $40$ nodes.
    \item As with the end-to-end baselines, the computational depth of the model had a significant effect on performance. In particular, 4 layer models failed to learn the task well, but 8-layer models achieved good in-distribution performance and a limited degree of out-of-distribution generalization.
\end{itemize}
\end{highlightbox}

\begin{figure}[h]
    \centering
    \includegraphics{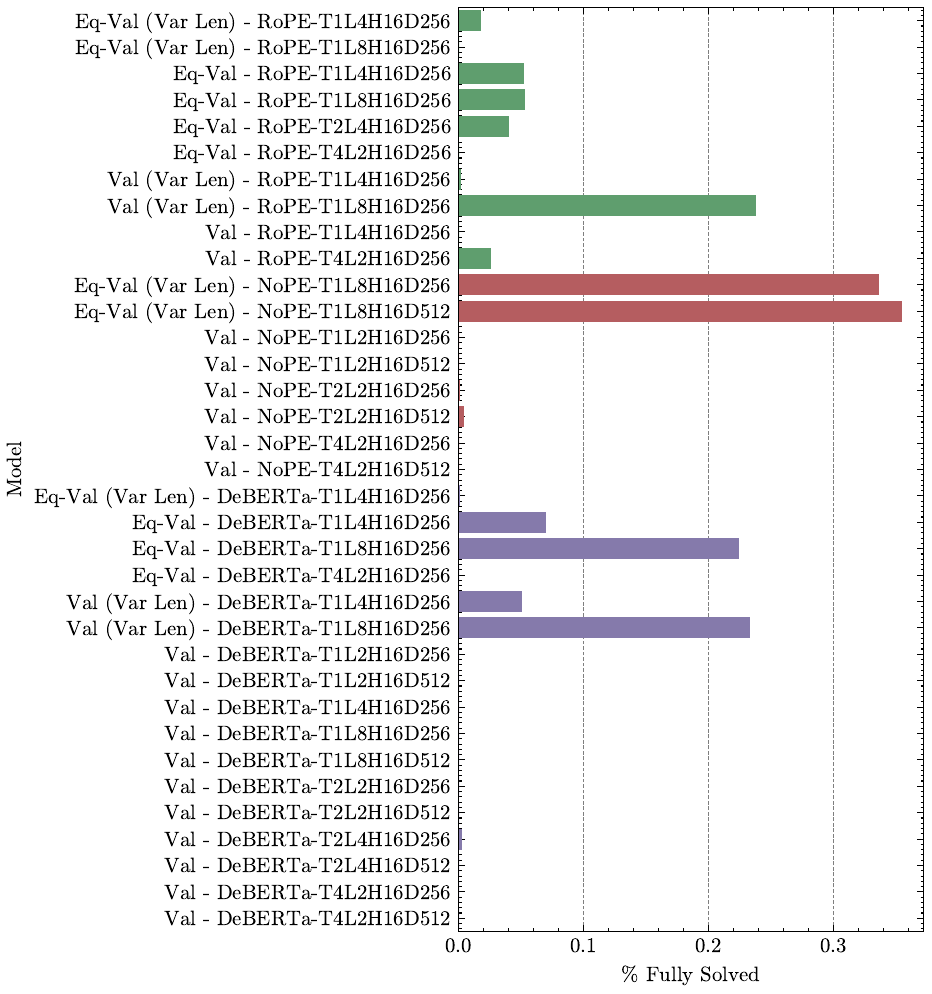}
    \caption{A comparison of average OOD generalization performance of different CoT-supervised baselines, varying architectural hyperparameters. The metric is full sequence accuracy, which measures the proportion of inputs where every node's value is computed correctly. The naming scheme matches the previous section, but adds a prefix describing the format of the CoT trajectories. ``Val'' means that the CoT trajectory directly computes the values of each variable, whereas ``Eq-Val'' first recalls the equations and then computes the values. Here, ``(Var Len)'' indicates runs where the input problem size is variable and randomly sampled in $N \leq 32$, rather than being only $N=32$.}
    \label{fig:cot_baselines_all_models_fullseqacc}
\end{figure}

\begin{figure}[h]
    \centering
    \includegraphics{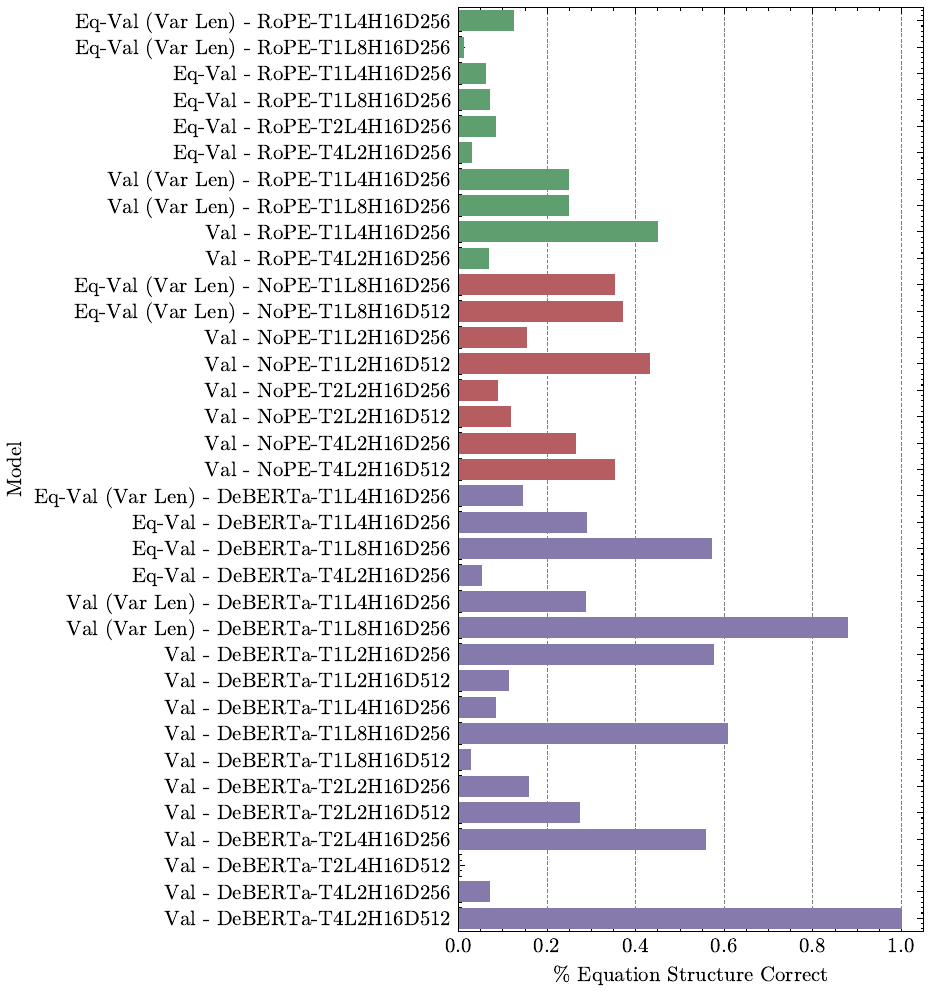}
    \caption{A comparison of average OOD generalization performance of different CoT-supervised baselines, varying architectural hyperparameters. The metric is ``\% Equation Structure Correct'', which measures the proportion of inputs where the autoregressively generated CoT has the correct equation structure (without checking whether the values computed are correct).} \label{fig:cot_baselines_all_models_varfullseqacc}
\end{figure}

\begin{figure}[h]
    \centering
    \begin{subfigure}[t]{0.45\linewidth}
        \centering
        \includegraphics[width=\linewidth]{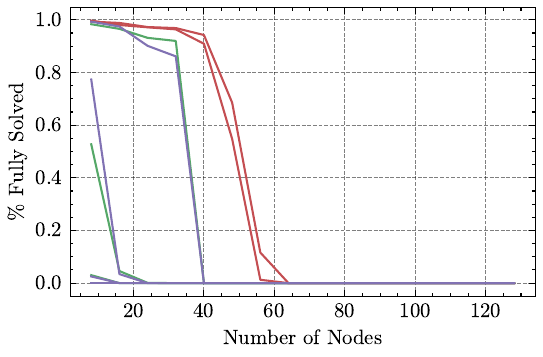}
        \caption{Each line corresponds to an experimental run. Lines are color-coded by positional encoding, but other architectural hyperparameters vary and are not represented.}
    \end{subfigure}
    \hfill
    \begin{subfigure}[t]{0.45\linewidth}
        \centering
        \includegraphics[width=\linewidth]{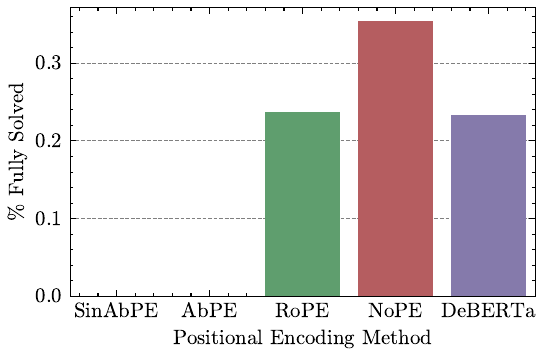}
        \caption{Average OOD performance across test splits for the best model of each positional encoding method.}
    \end{subfigure}

    \begin{subfigure}[t]{0.45\linewidth}
        \centering
        \includegraphics[width=\linewidth]{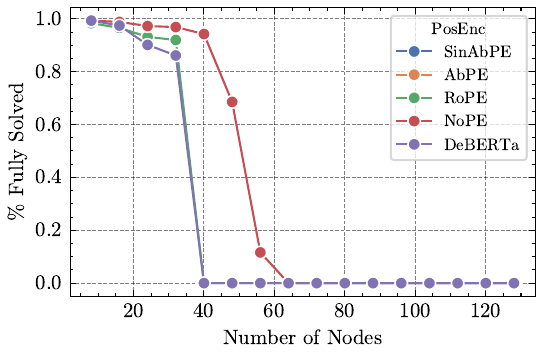}
        \caption{\% Fully solved by graph size for the best model of each positional encoding method. We find NoPE to achieve the best out-of-distribution generalization performance, generalizing well to 40 nodes when trained on $N \leq 32$ nodes. The other positional encoding methods fail to generalize beyond the training regime.}
    \end{subfigure}
    \hfill
    \begin{subfigure}[t]{0.45\linewidth}
        \centering
        \includegraphics[width=\linewidth]{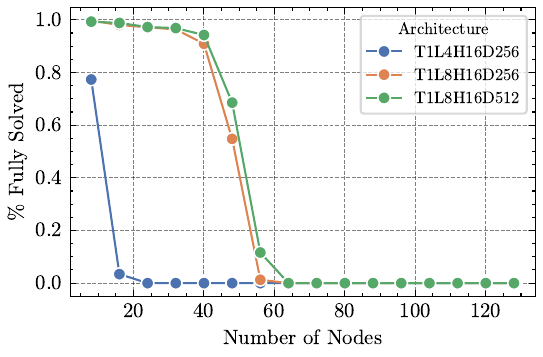}
        \caption{\% Fully solved by graph size for the best model of each architectural configuration. Computational depth (i.e., $T \cdot L$) is crucial for good performance, with shallow models performing poorly even in-distribution on larger inputs.}
    \end{subfigure}
    \caption{Further experimental results for CoT baselines. While chain-of-thought supervision yields improved performance over end-to-end models, out-of-distirbution generalization capabilities are limited.}
    \label{fig:cot_baseline_figures}
\end{figure}

\clearpage
\section{Details on Latent State Supervision}\label{sec:latent_supervision_details}

\subsection{Latent State Embedding Structure}

The input to the model is presented as a sequence of equations defining the value of each node in the computation graph. The vocabulary of the input includes variable names (e.g., $x_{42}$), numerical values (e.g., $17$), operations (e.g., $+$), and special symbols like equality $\token{=}$ or equation separation $\sep$.

To provide the model with supervision on each part of the input, we employ a special tokenization and embedding scheme. We use a factored structure to tokenize each symbol in the input into 4-component tokens: syntax, variable, operation, and value. For example, the input $\token{17} \token{=} \token{x_{42}} \sep ...$, is tokenized as follows before the first iteration:
\begin{center}
\begin{tabular}{ccrccccl}
     &  & & \texttt{syntax} & \texttt{variable} & \texttt{operation} & \texttt{value} & \\[8pt]
    $\token{17}$ & $\to$ & $[$ & $\mathtt{value}$ & $\mathtt{N/A}$ & $\mathtt{N/A}$ & $17$ & $]$\\
    $\token{=}$ & $\to$ & $[$ & $\token{=}$ & $\mathtt{N/A}$ & $\mathtt{N/A}$ & $\mathtt{N/A}$ & $]$\\
     $\token{x_{42}}$& $\to$ & $[$ & $\mathtt{variable}$ & $x_{42}$ & $\mathtt{N/A}$ & $\mathtt{empty}$ & $]$\\
    $\sep$ & $\to$ & $[$ & $\sep$ & $\mathtt{N/A}$ & $\mathtt{N/A}$ & $\mathtt{N/A}$ & $]$\\
\end{tabular}
\end{center}

The \textit{syntax} factor can be $\mathtt{value}$, $\mathtt{variable}$, $\mathtt{operation}$, or the special symbols $\token{=}$ or $\sep$. The \textit{variable} factor is the variable names $\sset{x_0, \ldots, x_{127}}$. The \textit{operation} factor is the set of arithmetic operations (e.g, $+,-, \times$). The \textit{value} factor is the set of numerical values (i.e., $\sset{0, \ldots, 22}$). We also include an \texttt{N/A} symbol for the \textit{variable}, \textit{operation}, and \textit{value} factors. For example, symbols with value syntax do not have a variable factor, etc. We also include a special \texttt{empty} symbol for the value factor of variable tokens. In the input to the model, the variable tokens have empty value factors because their values have not been computed yet. As the model processes the input, it iteratively computes the values of different variables and fills in their value factor.

We train a separate embedder for each factor, and map the input to vector embeddings by embedding each factor and adding the embeddings.

\subsection{Latent State Supervision}

The \methodcolorhighlight{continuoussupervisioncolor}{Continuous Latent Space Supervision}, \methodcolorhighlight{discretesupervisioncolor}{Discrete Latent Space Supervision}, and \methodcolorhighlight{discretesupervisionrecorrectioncolor}{Discrete Latent Space Supervision $\circlearrowleft$} methods share the same latent state supervision scheme. We train these recurrent models to learn to solve the input problem be computing node values one layer deeper in the computation graph with each recurrent iteration. We do this by defining a loss function at each iteration that penalizes predictions only for variables with depth less than or equal to the current iteration.

For each $\mathtt{factor} \in \sset{ \mathtt{syntax}, \mathtt{variable}, \mathtt{operation}, \mathtt{value}}$, we learn a linear read-out layer $W_{\mathtt{factor}} \in \reals^{d_{\mathrm{model}} \times \aabs{\calV_{\mathtt{factor}}}}$ that maps the vector state at the end of the recurrent iteration to a prediction of each factor. Here, $\calV_{\mathtt{factor}}$ denotes the vocabulary for the given factor (e.g., for the \textit{value} factor, this is $\sset{0, \ldots, 22, \mathtt{N/A}, \mathtt{empty}}$). 

We provide the model with supervision on its latent states by defining a loss for each factor and at each recurrent iteration. In particular, the loss function for the value factor is defined such that the model is trained to predict the values of all variables that occur at depth $\leq t$ in the computation graph. In particular, for an input sequence $X = (x_1, \ldots, x_n)$, the value factor loss at iteration $t$ is defined as
\begin{equation}
    \mathrm{Loss}(\mathtt{factor} = \mathtt{value}, \mathrm{iteration} = t) = \sum_{\substack{i \in [n] \\ \mathrm{Depth}(x_i) \leq t}} \ell\paren{W_{\mathtt{value}} \,E_i^{(t)}, \mathrm{Value}(x_i)}.
\end{equation}
where $\mathrm{Depth}(x_i)$ is the depth of the variable $x_i$ in the input computation graph, $\mathrm{Value}(x_i)$ is its computed value, and $E_i^{(t)} \in \reals^{d_\mathrm{model}}$ is the vector embedding of $x_i$ at recurrent iteration $t$. Here, $\ell$ is the cross-entropy loss.

The overall loss used to train the models is the sum of the individual factor losses at each iteration.
\begin{equation}
    \mathrm{Loss} = \sum_{\mathtt{factor}} \sum_{t} \mathrm{Loss}(\mathtt{factor} = \mathtt{value}, \mathrm{iteration} = t).
\end{equation}

\subsection{Discretization of Intermediate States}

The training procedure described above applies to the \methodcolorhighlight{continuoussupervisioncolor}{Continuous Latent Space Supervision}, \methodcolorhighlight{discretesupervisioncolor}{Discrete Latent Space Supervision}, and \methodcolorhighlight{discretesupervisionrecorrectioncolor}{Discrete Latent Space Supervision $\circlearrowleft$} methods in the same way. In the methods with a discrete latent bottleneck, we introduce an additional architectural mechanism where the read-out layers are used not only for computing the loss on the intermediate iterations, but also for mapping the latent representation to a discrete space.

In particular, letting $E_i^{(t)}$ be the embedding of the $i$-th token after $t$ recurrent iterations, we use argmax decoding of the linear read-outs to map the embedding to a discrete prediction for each factor. This discrete state is then re-embedded using the same learned embedding module to form the vectorized input $E_i^{(t+1)}$ at the next iteration. In particular, at iteration $t$, the model's forward pass is defined as follows
\begin{equation}
    \begin{aligned}
        (\tilde{E}_1^{(t+1)}, \ldots, \tilde{E}_n^{(t+1)}) &\gets \mathrm{RecurrentTransformerBlock}(E_1^{(t)}, \ldots, E_n^{(t)})\\
        z_{i, \mathtt{factor}}^{(t+1)} &\gets \argmax \sset{W_{\mathtt{factor}} \, \tilde{E}_i^{(t+1)} }\hphantom{~} \quad \mathtt{factor} \in \sset{ \mathtt{syntax}, \mathtt{variable}, \mathtt{operation}, \mathtt{value}} \\
        E_{i, \mathtt{factor}}^{(t+1)} &\gets \mathrm{FactorEmbed}(z_{i, \mathtt{factor}}^{(t+1)}) \quad \mathtt{factor} \in \sset{ \mathtt{syntax}, \mathtt{variable}, \mathtt{operation}, \mathtt{value}} \\
        E_i^{(t+1)} &\gets E_{i, \mathtt{syntax}}^{(t+1)} + E_{i, \mathtt{variable}}^{(t+1)} + E_{i, \mathtt{operation}}^{(t+1)} + E_{i, \mathtt{value}}^{(t+1)}. \label{eq:embeddings}
    \end{aligned}
\end{equation}

This discretization enables us to train the model with a type of \textit{teacher-forcing} across recurrent iterations. That is, we can teacher-force the inputs $z_i^{(t)}$ at each iteration $t$. This enables more efficient training.

\subsection{Self-Correction Mechanism}

In a reasoning task, each reasoning step depends crucially on the prior steps in the reasoning path. If a mistake is made at any stage, all subsequent computation is affected, and the error is often fatal. As the size of the problem and the number of computational steps scale, the likelihood of an error occurring at \textit{some point} in the reasoning process becomes large, limiting the ability to generalize indefinitely to more complex problems. To address this challenge, a reasoning model must be able to detect and correct errors as they occur in order to recover when a mistake is made in its previous computation.

We train the model to detect and correct errors by randomly corrupting the model's latent state. That is, at each iteration, with some small probability, we corrupt a random selection of the value components of the models' discrete states. To achieve good loss, the model must learn to detect when a previously-computed value is incorrect and correct it before proceeding.

\subsection{Experiment Details \& Additional Results}

As with the baselines, we explore the effect of different architectural hyperparameters, such as positional encoding and the depth of the recurrent block, on model performance.~\Cref{fig:ourmethod_all_models_fullseqacc} depicts the average OOD perfromance as measured by the ``\% Fully Solved'' metric for each model configuration in the \methodcolorhighlight{discretesupervisioncolor}{Discrete Latent Space Supervision}, and \methodcolorhighlight{discretesupervisionrecorrectioncolor}{Discrete Latent Space Supervision $\circlearrowleft$} methods. The results in the main text correspond to the best-performing models according to this metric. In particular, the best-performing \methodcolorhighlight{discretesupervisioncolor}{Discrete Latent Space Supervision} model is DeBERTa-L2H16D256, and the best-performing \methodcolorhighlight{discretesupervisionrecorrectioncolor}{Discrete Latent Space Supervision $\circlearrowleft$} model is DeBERTa-L4H16D384.

\Cref{fig:our_method_figures} depicts additional experimental results for the \methodcolorhighlight{discretesupervisioncolor}{Discrete Latent Space Supervision}, and \methodcolorhighlight{discretesupervisionrecorrectioncolor}{Discrete Latent Space Supervision $\circlearrowleft$} methods. We highlight a few observations here:
\begin{highlightbox}
\begin{itemize}
    \item The positional encoding method is critical for length generalization. The DeBERTa positional encoding method (a relative positional encoding method) performed the best by far.
    \item 2 layers for the recurrent block were sufficient for the \methodcolorhighlight{discretesupervisioncolor}{Discrete Latent Space Supervision} method. However, the recorrection mechanism of \methodcolorhighlight{discretesupervisionrecorrectioncolor}{Discrete Latent Space Supervision $\circlearrowleft$} required a deeper recurrent block. We saw no significant improvement for the re-correction mechanism with 2 layers, but with 4 layers, the re-correction mechanism kicked in and enabled near-perfect OOD generalization.
\end{itemize}
\end{highlightbox}

\begin{figure}[h]
    \centering
    \includegraphics{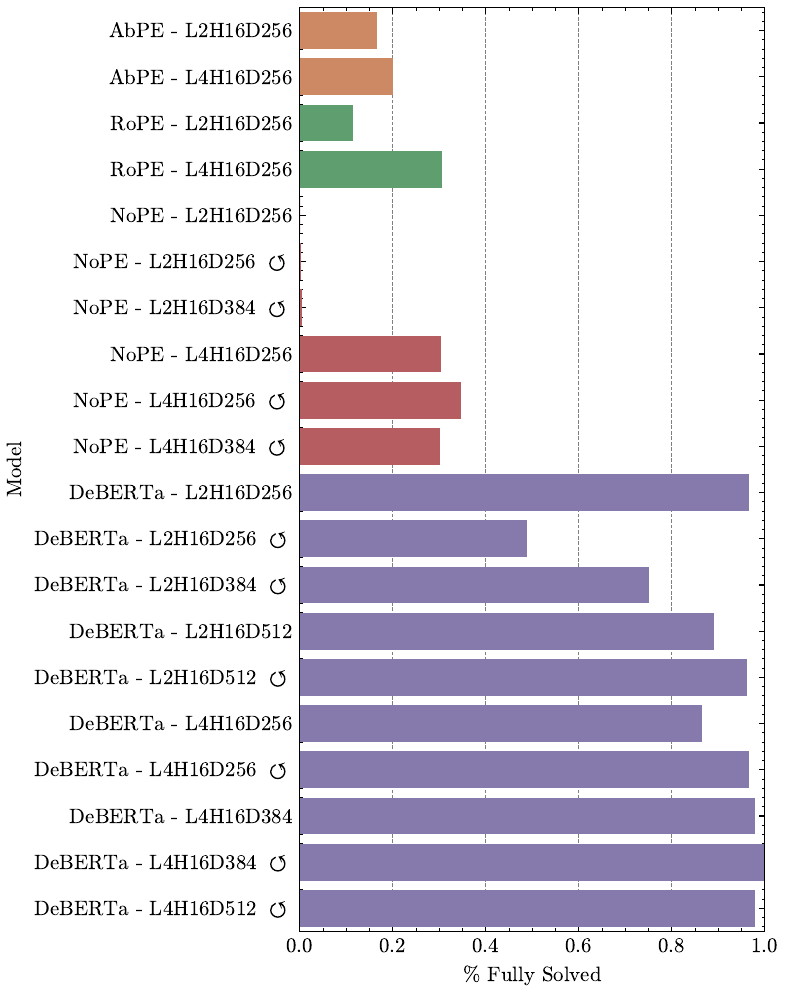}
    \caption{Average ``\% Fully Solved'', across \# nodes between 8 and 128, with training on $\leq 32$ nodes, }
    \label{fig:ourmethod_all_models_fullseqacc}
\end{figure}

\begin{figure}[h]
    \centering
    \begin{subfigure}[t]{0.45\linewidth}
        \centering
        \includegraphics[width=\linewidth]{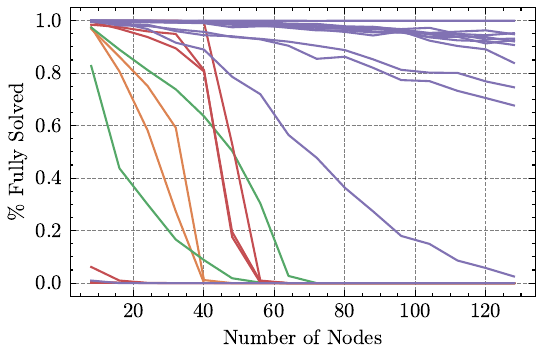}
        \caption{Each line corresponds to an experimental run. Lines are color-coded by positional encoding, but other architectural hyperparameters vary and are not represented.}
    \end{subfigure}\quad
    \begin{subfigure}[t]{0.45\linewidth}
        \centering
        \includegraphics[width=\linewidth]{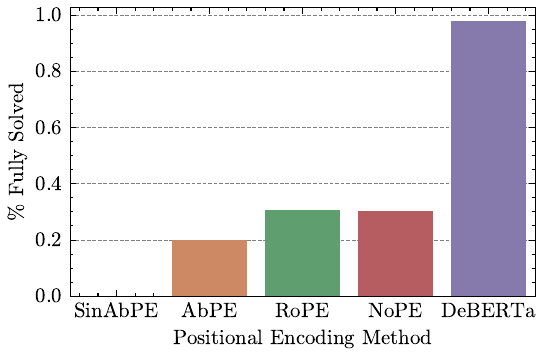}
        \caption{Average OOD performance across test splits for the best model of each positional encoding method.}
    \end{subfigure}
    \begin{subfigure}[t]{0.45\linewidth}
        \centering
        \includegraphics[width=\linewidth]{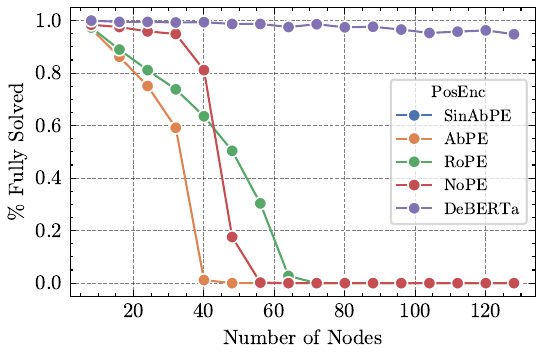}
        \caption{\% Fully solved by graph size for best model of each positional encoding method.}
    \end{subfigure}\quad
    \begin{subfigure}[t]{0.45\linewidth}
        \centering
        \includegraphics[width=\linewidth]{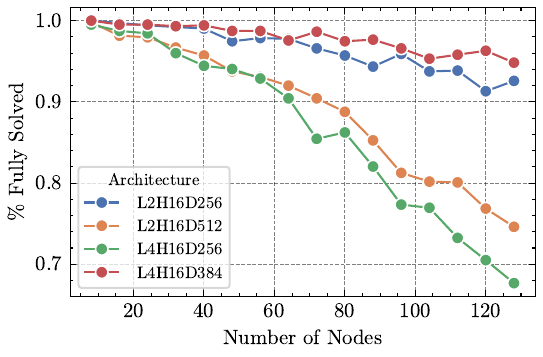}
        \caption{\% Fully solved by graph size for best model of each architectural configuration.}
    \end{subfigure}
    \caption{Further experimental results for methods exploring our proposed architectural mechanisms.}
    \label{fig:our_method_figures}
\end{figure}

\clearpage
\section{Details of Mechanistic Interpretability Analysis}\label{sec:mechinterp_details}

In this section, we provide additional experimental evidence to support our claim on the mechanism learned by the model together with the error analysis of the model's predictions.

\paragraph{Notice:} The following analysis is conducted only for showing the computation happening at the Right-Hand Side (RHS) position  in each equation, as it is the place where the model is expected to compute the final result. 

\paragraph{Model Configuration.}
We use DeBERTa-L2H16D256 trained with our proposed \methodcolorhighlight{discretesupervisioncolor}{Discrete Latent Space Supervision} method (without the re-correction mechanism) on the mathematical reasoning task. 
Specifically, the recurrent Transformer model is configured with two transformer blocks, 16 attention heads, a hidden dimension of 256. 
We use DeBERTa's relative positional encoding method.
The training data is the same as the one used in the main text. 
We choose this model setup because it is the best-performing configuration according to the ``\% Fully Solved'' metric displayed in \Cref{fig:ourmethod_all_models_fullseqacc} for a two-layer model.
In particular, we cherry-pick the best-performing model trained with the same configuration with different random seeds, which has a ``\% Fully Solved'' score of 99.98\% on the OOD test set.
We use this model to conduct the mechanism analysis for better interpretability.
We train on modular-23 addition task with maximum graph size 32. 
The total number of variables in the training data is 128.
The testing data used for mechanism analysis has the maximum graph size 128.

\paragraph{Testing Data for Interpretation Analysis.} 
To rigorously understand the inner workings of the model, we conduct controlled experiments by systematically varying the input data fed to the trained model.
In particular, each testing example is a sequence of arithmetic equations with number of nodes 128, appended with a new \textbf{probe equation} to the end of the sequence with the following format:
\begin{equation}
    \sep \: \VAR{0} \: \token{+} \:\VAR{1} \token{+} \: \VAR{2} \: \token{=} \: \RHS . 
    \label{eq:pseudo_equation}
\end{equation}
where $\VAR{0}$, $\VAR{1}$, and $\VAR{2}$ are the three variables in the probe equation, and $\RHS$ is the right-hand side of the probe equation. 
Thus, $\VAR{0}, \VAR{1}, \VAR{2}, \RHS$ are  chosen from $\mathcal{V} = \sset{x_1, \ldots, x_{128}}$, and the true values of these variables are in $\mathcal{N} = \sset{0, 1, \ldots, 22}$.

\paragraph{Additional Definitions and Notations.}
In the following, we frequently use the following definitions and notations:
\begin{itemize}
    \item \textbf{Head Output}: For a given attention head $h$, we define the head output for a query vector $q_h\in\RR^{d_h}$ for head dimension $d_h$ as 
    \begin{align*}
        \mathrm{Head~Output}(h) = \mathrm{softmax}(q_h K_h^\top / \sqrt{d_k}) V_h W_O^{(h)}, 
    \end{align*}
    where $K_h$ and $V_h$ are the key and value matrices of the head $h$, respectively, and $W_O^{(h)}$ is the output projection matrix of the head $h$.
    In standard attention mechanism, each head'squery, key and value vector is obtained by applying a linear transformation to the attention input specified by $W_Q^{(h)}$, $W_K^{(h)}$, and $W_V^{(h)}$, respectively.
    The above definition can be applied to define the head output for \emph{any} query position.
    However, since our mechanism analysis focuses exclusively on the RHS position, we consistently define the head output as the output of the attention head at the RHS position.
    Here, we don't include the bias of the head output projection in the definition of the head output, as the bias applied to the final attention output is not specified to individual heads.
    \item \textbf{OV Combined Matrix}: For a group of attention heads $\mathcal{H}\subseteq[16]$, we define the OV combined matrix as
    \begin{align*}
        W_{OV}^{(\calH)} = \sum_{h\in \mathcal{H}} W_V^{(h)} W_O^{(h)},
    \end{align*}
    where $W_O^{(h)}\in \mathbb{R}^{d_h\times d}$ and $W_V^{(h)}\in \mathbb{R}^{d\times d_h}$ are the output projection matrix and the value projection matrix of the attention head $h\in\mathcal{H}$, respectively.
\end{itemize}

\subsection{Technique Overview}

To simplify the discussion, let us consider a concrete example of the probe equation:
\begin{equation}
    \sep  x_{91}  \token{+}  x_{88}  \token{+}  x_{55 }  \token{=}  x_{30}  , 
    \label{eq:pseudo_equation_example}
\end{equation}
i.e., $\VAR{0} =  x_{91}$, $\VAR{1} =  x_{88}$, $\VAR{2} =  x_{55}$, and $\RHS =  x_{30}$.
To solve this equation, at the token $\RHS$,the model needs perform the following computations:
\begin{itemize}
    \item [(i)] Identify that the \coloremph{variable names} $x_{91}$, $x_{88}$, and $x_{55}$ appear in the left-hand side of the equation;
    \item [(ii)] Retrieve the \coloremph{values} of the variables $x_{91}$, $x_{88}$, and $x_{55}$ from the previous equations, denoted by $\mathtt{value}(x_{91})$, $\mathtt{value}(x_{88})$, and $\mathtt{value}(x_{55})$, respectively;
    \item [(iii)] Compute the \coloremph{modular sum}   $\mathtt{value}(x_{91}) + \mathtt{value}(x_{88}) + \mathtt{value}(x_{55}) \mod{23}$.
\end{itemize}

In the following, we show that the first layer attention is responsible for identifying the variable names, the second layer attention is responsible for retrieving the values of the variables, and the last MLP layer computes the modular sum. Before we dive into the details, we first introduce the three interpretation techniques that we will use in the sequel.

\subsubsection{Interpretation Technique for Attention Layers}

\paragraph{Overview.} Attention mechanisms in transformers fundamentally perform information routing—they decide what information from previous token positions should be aggregated at the current position. In our arithmetic reasoning task, we hypothesize that attention heads act as specialized circuits that \coloremph{copy specific types of information} from source tokens to the RHS position where computation occurs. To understand these circuits, we need to answer two fundamental questions: 
\begin{highlightbox}
\begin{itemize}
    \item [(i)] \emph{Which} tokens does each attention head attend to? 
    \item [(ii)] \emph{What} information is being copied from those tokens?
\end{itemize}
\end{highlightbox}

Our approach combines controlled experimentation with mathematical analysis. We use probe equations with systematic variations to identify attention patterns, then employ linear algebra techniques to decode the information flow through the model's weight matrices. This methodology reveals that attention heads self-organize into functional groups, with each group specialized for a specific subtask in the arithmetic computation pipeline.

\paragraph{Application to First Layer Attention.}
For the first layer, we test whether attention heads identify which variables appear in the equation. We construct probe equations of the form in \eqref{eq:pseudo_equation} and systematically vary the \emph{variable names} (e.g., changing $\VAR{0}$ from $x_{91}$ to $x_{42}$) while keeping other variables fixed. For each configuration, we measure the relative variance of each head's output, which captures how much a head's output changes when we vary a specific input, normalized by the head's typical output magnitude. High relative variance indicates the head is sensitive to changes in that variable position (see Section~\ref{sec:mech_interp_group_structure} for the rigorous definition). Through this analysis, we discover that heads form distinct groups: heads \{4, 8\} attend to $\VAR{0}$, heads \{5, 12\} to $\VAR{1}$, and heads \{3, 7, 11, 14\} to $\VAR{2}$.

To understand what these heads copy, we analyze their combined OV matrices using norm amplification—measuring how much each factored embedding type (among $\mathtt{syntax}, \mathtt{variable}, \mathtt{operation}, \mathtt{value}$) is amplified when passed through the matrix. We find that these heads specifically amplify the \texttt{variable} factor (with amplification $\approx 15$) over other factors ($\approx 5$), confirming they copy variable \emph{identities} rather than values. 
Our analysis suggests that the first layer attention implements \coloremph{variable identification}, with specialized head groups that extract and route variable names from equation positions to the RHS for further processing.

\paragraph{Application to Second Layer Attention.}
For the second layer, we test whether attention heads retrieve variable values needed for computation. Using the same probe equation structure, we now vary the \emph{values} of variables of the probe equation (by modifying earlier equations in the sequence) while keeping variable names fixed. We again measure relative variance to identify head groups and find: heads \{0, 8, 15\} retrieve values for $\VAR{0}$, heads \{5, 10\} for $\VAR{1}$, and heads \{2, 3, 4, 7, 9\} for $\VAR{2}$.

The norm amplification analysis reveals these heads strongly amplify the \texttt{value} factor over others, confirming they copy numerical values rather than identities. Interestingly, the copied value embeddings maintain near-orthogonality between different variables while showing circulant patterns within each variable—suggesting a Fourier-like encoding that facilitates downstream arithmetic operations. In conclusion, the second layer attention implements \coloremph{value retrieval}, with specialized head groups that fetch the numerical values of variables identified by the first layer, preparing them for arithmetic computation.

\subsubsection{Interpretation Technique for the Second MLP Layer}

After the attention layers have assembled the necessary information, the second MLP must perform the actual arithmetic computation. At this point, the residual stream at the RHS position contains structured information from both attention layers: variable identities from the first layer and their corresponding values from the second layer. To understand how the MLP transforms this information into the final answer, we analyze the structure of its input representations and apply frequency domain analysis to decode the computation.

\paragraph{Understanding the MLP Input Structure.}
The input to the second MLP at the RHS position consists of value embeddings that have been copied and transformed by the second layer attention. To understand this precisely, we need to consider three key components:

First, recall that in our factored representation, the \texttt{value} factor can take 25 possible values: the numbers 0-22 for actual computations, plus special tokens \texttt{empty} and \texttt{N/A}. Each of these 25 values has a learned embedding vector $\mathtt{embed}_{\mathtt{value}}(v)$ where $v \in \{0, 1, \ldots, 22, \texttt{empty}, \texttt{N/A}\}$.

Second, the second layer attention has three distinct head groups --- $\mathcal{H}_0 = \{0, 8, 15\}$, $\mathcal{H}_1 = \{5, 10\}$, and $\mathcal{H}_2 = \{2, 3, 4, 7, 9\}$ --- each responsible for copying the value of one variable position. Each group has its own combined OV matrix that transforms the value embeddings it copies. This transformation creates what we call "new value" embeddings:
\begin{align}\label{eq:new_value_def}
\mathtt{new\_value}_i(v) = W_{OV}^{(\mathcal{H}_i)} \cdot \mathtt{embed}_{\mathtt{value}}(v)
\end{align}
where $i \in \{0, 1, 2\}$ indicates which variable position the head group attends to, and $v$ is the actual value being transformed. Intuitively, $\mathtt{new\_value}_i(v)$ measures the contribution of of $\VAR{~i}$ to the residual stream at the RHS position, after second layer attention, and before the second MLP, when $\VAR{~i}$ has value $v$.

Third, since each of the three head groups can potentially transform any of the 25 value embeddings, there exist $3 \times 25 = 75$ distinct transformed value embedding vectors in total. These 75 vectors essentially form a lookup table: for variable position $i$ with value $v$, the corresponding transformed embedding is $\mathtt{new\_value}_i(v)$.

In a concrete example: when the probe equation has $\VAR{0} = x_{91}$ with value 7, $\VAR{1} = x_{88}$ with value 15, and $\VAR{2} = x_{55}$ with value 3, the second layer attention specifically copies and transforms three of these 75 vectors --- $\mathtt{new\_value}_0(7)$, $\mathtt{new\_value}_1(15)$, and $\mathtt{new\_value}_2(3)$ --- to the RHS position. The residual stream at RHS thus contains the sum of these three transformed value embeddings, which serves as the input to the second MLP. Crucially, as we will see in \Cref{sec:interpret_mlp_detail}, these transformed embeddings exhibit a \coloremph{circulant structure} that makes them amenable to frequency analysis. 

\paragraph{Frequency Domain Analysis.}
Motivated by the circulant structure, we propose to adopt frequency domain analysis to understand how the second layer MLP performs modular addition.
Our approach involves two key steps. First, we systematically vary the inputs by creating probe equations where $\VAR{0}$, $\VAR{1}$, and $\VAR{2}$ each take all values from 0 to 22. For each configuration $(x, y, z) \in \{0, 1, \ldots, 22\}^3$, the residual stream at the RHS position contains the sum $\mathtt{new\_value}_0(x) + \mathtt{new\_value}_1(y) + \mathtt{new\_value}_2(z)$.

Second, we extract these representation vectors at four key network positions --- MLP pre-activation, post-activation, output, and final decoder --- and apply a three-dimensional Discrete Fourier Transform (DFT). This DFT, computed independently for each coordinate of the representation vector, transforms our data into a 4D tensor: three dimensions for the frequencies $(a, b, c) \in \{0, 1, \ldots, 22\}^3$ corresponding to the three input variables, and one dimension for the vector coordinates. By analyzing the norm of different frequency patterns in this tensor, we discover that the MLP \coloremph{performs modular addition through sinusoidal basis functions}, with the diagonal frequencies $(a, a, a)$, representing the sum $x + y + z$, becomes dominant as signals propagate through the network.
Therefore, the second MLP implements modular arithmetic in the frequency domain, where the periodic nature of trigonometric functions naturally handles the modulo-23 computation. This result is consistent with the findings in the literature. See,  e.g., \citet{nandaprogress, tian2024composing, doshi2024grok} for more details.




\subsection{First Layer Attention: Variable Copying}
In the followiing, we will give a detailed analysis of the first layer attention mechanism.

\subsubsection{Group Structure in the First Layer Attention}\label{sec:mech_interp_group_structure}
\begin{figure}[ht]
    \centering
    \begin{subfigure}[b]{0.48\textwidth}
        \centering
        \includegraphics[width=\linewidth]{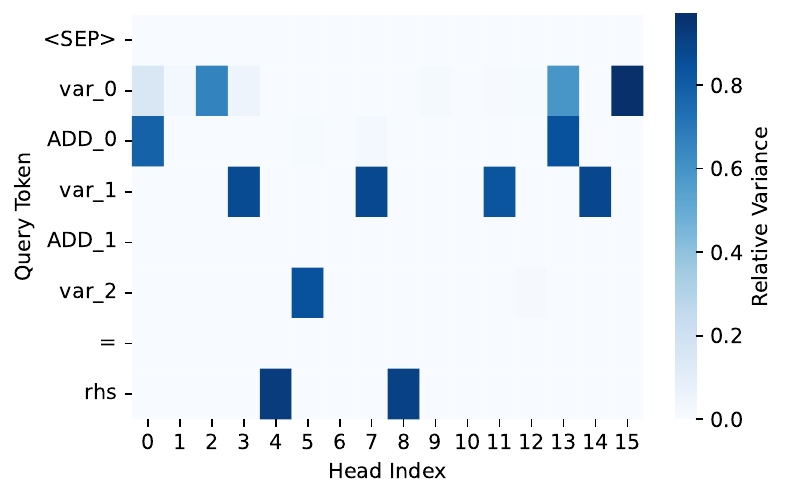}
        \caption{$\mathrm{Relative~Variance}$ for $\VAR{0}$}
        \label{fig:L0_var0_rvar}
    \end{subfigure}
    \hfill
    \begin{subfigure}[b]{0.48\textwidth}
        \centering
        \includegraphics[width=\linewidth]{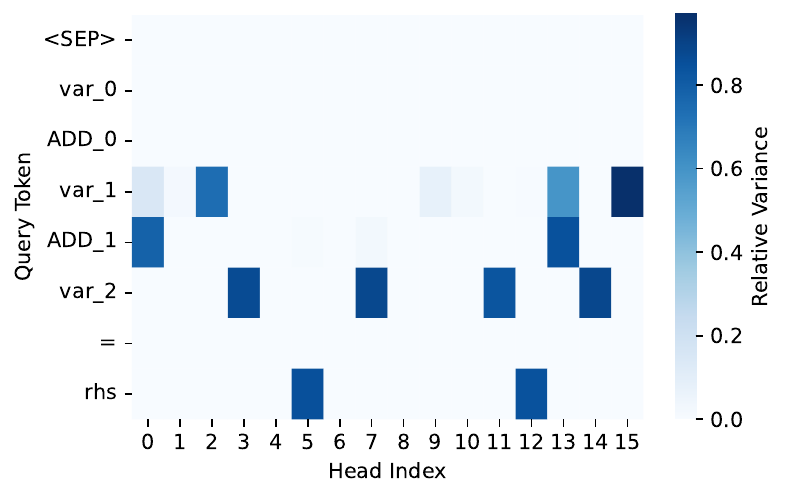}
        \caption{$\mathrm{Relative~Variance}$ for $\VAR{1}$}
        \label{fig:L0_var1_rvar}
    \end{subfigure}
    \vspace{0.5em}
    \begin{subfigure}[b]{0.48\textwidth}
        \centering
        \includegraphics[width=\linewidth]{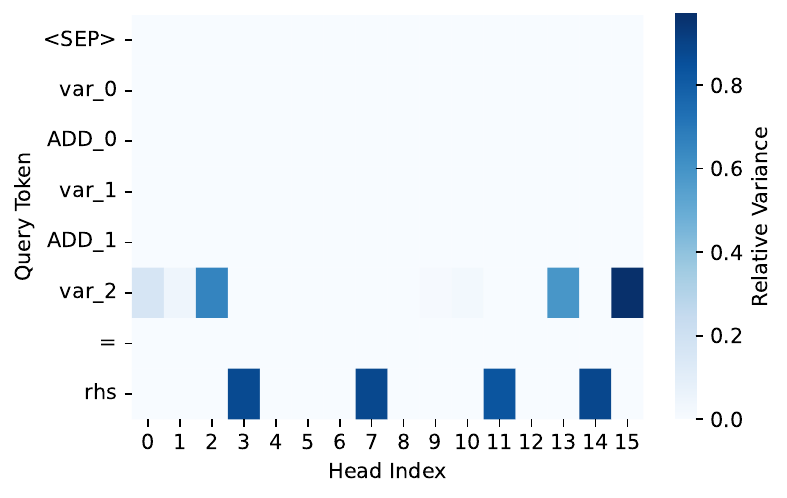}
        \caption{$\mathrm{Relative~Variance}$ for $\VAR{2}$}
        \label{fig:L0_var2_rvar}
    \end{subfigure}
    \hfill
    \begin{subfigure}[b]{0.48\textwidth}
        \centering
        \includegraphics[width=\linewidth]{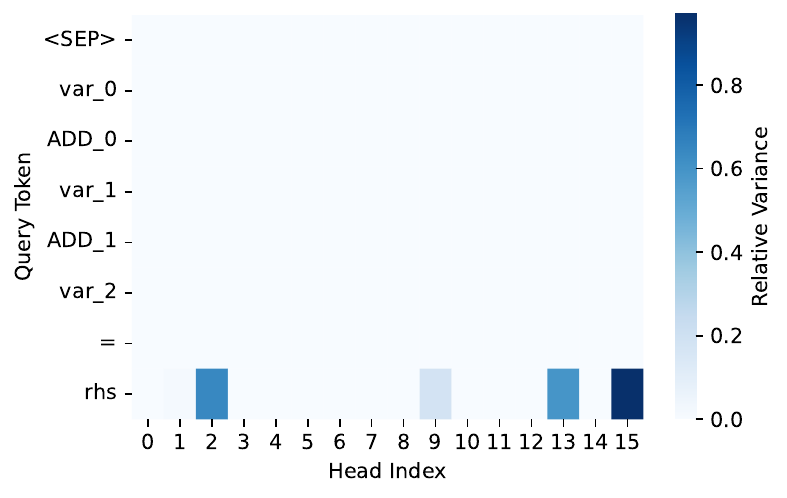}
        \caption{$\mathrm{Relative~Variance}$ for $\RHS$}
        \label{fig:rhs_rvar}
    \end{subfigure}
    \caption{Relative variance heatmaps when we vary the value of $\VAR{0}$, $\VAR{1}$, $\VAR{2}$, and $\RHS$. Each row corresponds to a query position and each column corresponds to an attention head.}
    \label{fig:L0_rvar}
\end{figure}

To rigorously demonstrate this grouping structure,

\paragraph{Experiment Design for Group Structure Detection.}
The attention heads in the first layer exhibit a clear grouping pattern based on which variable position they attend to. To identify the group structure in the first layer attention and detect which heads belong to which groups, we measure each head's \textbf{relative variance} when we vary the variable name  of each of the four variables $\VAR{0}$, $\VAR{1}$, $\VAR{2}$, and $\RHS$ in the probe equation in \eqref{eq:pseudo_equation}.

As an example, to detect which heads attend to $\VAR{0}$, we fix $\VAR{1}$, $\VAR{2}$, and $\RHS$ while randomly sampling different variables $\token{x_i}$ with $i=1,\ldots,128$ for $\VAR{0}$ in \eqref{eq:pseudo_equation}. 
Note that the variable $\token{x_i}$ must be computed in the preceding equations. 
Otherwise, the model cannot compute the value of $\token{x_i}$ and the probe equation is invalid.
As our testing data has all variables computed in the preceding equations, we collect 128 samples that only differ in the value of $\VAR{0}$.
We then compute the relative variance of each attention head's output at each position within the probe equation across the 128 samples.
Note that relative variance is a measure of how much the head's output varies in response to changes in $\VAR{0}$, and we give the rigorous definition in the next paragraph.
The analysis can also be conducted for the other variable positions, i.e., $\VAR{1}$, $\VAR{2}$, and $\RHS$, and the results are reported in \Cref{fig:L0_rvar}.

\paragraph{Relative Variance Calculation.}
Let us take $n$ different sequences, e.g., the 128 sequences in the above experiment design.  
We only consider one RHS position for the probe equation in each sequence.
For a given attention head $h$, we define the relative variance over the $n$ sequences as
\begin{align}
    \mathrm{Relative~Variance}(h) = \frac{\mathrm{tr}(\mathrm{Cov}(\mathrm{Head~Output}(h)))}{\mathbb{E}[\|\mathrm{Head~Output}(h)\|_2^2]}.
    \label{eq:relative_variance}
\end{align}
Here, the covariance matrix for a sequence of vectors $v_1, \ldots, v_n$ is defined as
\begin{align*}
    \mathrm{Cov}(v_1, \ldots, v_n) = \frac{1}{n} \sum_{i=1}^n (v_i - \bar{v})(v_i - \bar{v})^\top, 
\end{align*}
where $\bar{v}$ is the mean of the sequence over the $n$ sequences, and $\mathbb{E}[\cdot]$ is the empirical expectation over the $n$ sequences. Intuitively, the relative variance measures how much the head's output varies relative to its overall magnitude. 
\emph{A higher relative variance indicates that the attention head's output has a larger variance relative to its overall magnitude.}
Since we only change $\VAR{0}$ in the above example, a larger relative variance for a head means that the head's output is primarily influenced by $\VAR{0}$.

\paragraph{Illustration of \Cref{fig:L0_rvar}.}
In \Cref{fig:L0_rvar}, we plot the relative variance heatmaps for all 16 attention heads when we vary the variable names of $\VAR{0}$, $\VAR{1}$, $\VAR{2}$, and $\RHS$.
Each column corresponds to a different attention head, and each row corresponds to a different query position.
As our goal is to understand the mechanism at the $\RHS$ position, we focus on the \emph{last row}, which corresponds to the $\RHS$ query position in the figures. 
Each subfigure plots the relative variance heatmap for altering one particular variable.
A higher relative variance in one subfigure indicates that the attention head's output is more sensitive to changes in the corresponding variable.
Based on these results, we observe a clear group structure in the first layer's attention heads: \coloremph{Heads 4 and 8's relative variance is high only when we change the value of $\VAR{0}$, while the relative variance of the other heads is low.} 
This facts suggests that \coloremph{heads 4 and 8 attend primarily to $\VAR{0}$}. 
Similarly, \coloremph{heads 5 and 12 attend primarily to $\VAR{1}$}, and \coloremph{heads 3, 7, 11, and 14 attend primarily to $\VAR{2}$}. 
The last subfigure plots the relative variance heatmap for the $\RHS$ position.
We observe that \coloremph{heads 2, 9, 13, and 15 attend to the RHS position $\RHS$}, and the remaining heads do not exhibit a distinct pattern according to the relative variance heatmap and are unimportant.
Notice that the \coloremph{above head groups are all disjoint}. 
This further indicates that each head is specialized for a specific variable position.
This result is also backed up by the trace of the attention logits of the first layer attention heads as shown in \Cref{fig:l0}.

\paragraph{Summary of the Group Structure.}
We observe that the attention heads in the first layer exhibit a clear grouping pattern based on which variable position they attend to.
Therefore, we know that the first layer attention must be \coloremph{copying something} from the LHS variables to the RHS position.
In the following, we will conduct further analysis to identify \coloremph{what information is being copied}.

\subsubsection{First Layer Attention Copying the Variable Identity}
Here, by saying ``copying the variable identity'', we mean that the attention head is copying the factored embedding of $\mathtt{variable}$ among the four factored embeddings $\{\mathtt{syntax}, \mathtt{variable}, \mathtt{operation}, \mathtt{value}\}$.
In the previous experiment, we have identified that the first layer attention heads are grouped into four groups, each of which attends to a specific variable position.
Now, we aim to identify which of the four factored embeddings is being copied by these groups.

\paragraph{Norm Amplification Analysis.}
As shown in the last equation of \eqref{eq:embeddings}, the embedding vector $E_i^{(t)}$ is a sum of the  four factored embeddings. We have shown that each attention head group predominantly attends to a particular variable name. In this case, at $\RHS$, vectors of the form $W_{OV}^{(h)} E_i^{(t)}$ is added to the residual stream in the $t$-th Recurrent Transformer block, where $i$ is a previous token (more specifically, the token that head $h$ attends to).  Intuitively, if the OV matrix is responsible for copying a particular factor, then we would expect that it is more aligned to the subspace spanned by the embeddings of this particular factor.

To rigorously characterize this intuition, we analyze the norm amplification for each type of factored embeddings when passed through the combined OV matrix of different head groups.
Specifically, we define the norm amplification for a matrix $W_{OV}$ on input $x$ as:
\begin{align}
    \mathrm{Norm~Amplification}(W_{OV}, x) = \frac{\| W_{OV} x\|_2}{\|x\|_2}.
    \label{eq:norm_amplification}
\end{align}
Note that the above definition can be applied to any matrix $W_{OV}$ and input $x$ with agreeing dimensions.
For our analysis, we will consider $W_{OV}$ as the combined OV matrix of the attention heads in a group.
Specifically, let $\mathcal{H}\subseteq[16]$ be a group of attention heads, and let $W_O^{(h)}$ and $W_V^{(h)}$ be the output projection matrix and the value projection matrix of the attention head $h\in\mathcal{H}$, respectively.
The combined attention OV matrix for a group $\mathcal{H}\subseteq[16]$ is then defined as 
\begin{align*}
    W_{OV}^{(\calH)} = \sum_{h\in \mathcal{H}} W_V^{(h)} W_O^{(h)}.
\end{align*}
Here, the group structure is discovered by \Cref{sec:mech_interp_group_structure}. That is, the $16$ heads are split into   
\begin{align}\label{eq:group_structure_heads}
\underbrace{\{ 4, 8\}}_{\displaystyle \VAR{0}}, \qquad \underbrace{\{ 5, 12\}}_{\displaystyle \VAR{1}}, \qquad \underbrace{\{ 3, 7, 11, 14\}}_{\displaystyle \VAR{2}}, \qquad \underbrace{\{ 2, 9, 13, 15\}}_{\displaystyle \RHS}, \qquad \underbrace{\{ 0,1, 6,10\}}_{\displaystyle \textrm{unimportant}},
\end{align}
depending on which variable they attend to. 

By the definition of norm amplification, if the OV matrix is responsible for copying the identity of the variable, we expect to see a large amplification for the ``$\mathtt{variable}$'' factored embedding, and a small amplification for the other types of embeddings $\{\mathtt{syntax}, \mathtt{operation}, \mathtt{value}\}$. 
In that case, 
With  slight abuse of notation, for each $\mathtt{factored~embedding~type} \in\{\mathtt{syntax}, \mathtt{variable}, \mathtt{operation}, \mathtt{value}\}$, we can define the norm amplification for each \emph{factor type} as
\begin{align*}
    \mathrm{Norm~Amplification}(W_{OV}^{(\calH)}, \mathtt{factor~type}) = \mathbb{E}_{x\in\mathtt{factor~type}}\Bigl[\frac{\| W_{OV}^{(\calH)} x\|_2}{\|x\|_2}\Bigr].
\end{align*}
Here, $\mathbb{E}_{x\in\mathtt{factor~type}}$ is the average over the set of all factored embeddings of the same type and $\mathcal{H}$ is one of the five groups in \eqref{eq:group_structure_heads}.
For example, if we consider the ``$\mathtt{variable}$'' factored embedding type and $\mathcal{H} = \{4,8\}$, we have
\begin{align*}
    \mathrm{Norm~Amplification}(W_{OV}^{(\calH)}, \mathtt{variable}) = \mathbb{E}_{x\in\mathtt{variable}}\Bigl[\frac{\| W_{OV}^{(\calH)} x\|_2}{\|x\|_2}\Bigr], 
\end{align*}
where $x$ iterates over all the 128 factored embeddings of the type $\mathtt{variable}$ and the OV matrix is based on the head group $\{ 4, 8\}$.
In this case, this quantity measures the average norm amplification of the factored embeddings of the type $\mathtt{variable}$ when passed through the combined OV matrix of the head group $\{ 4, 8\}$.

\paragraph{Comparing the Norm Amplification Across Different Groups.}
In \Cref{fig:l0} (right), we compute the norm amplification by averaging the norm amplification over all the factored embeddings within each ``$\mathtt{factor~type}$''.
In \Cref{fig:factored_embeddings}, we further provide a histogram for each factored embedding's norm amplification for different ``$\mathtt{factor~type}$'' while different groups are highlighted in different colors, which provides a more detailed view of the norm amplification across different groups.

\Cref{fig:factored_embeddings} shows that, for each head group, the \coloremph{norm amplification for the ``$\mathtt{variable}$'' factored embedding is significantly larger} than that of the other types of embeddings. This can be seen by comparing the histograms in each subfigure with the same color. 
 This confirms our hypothesis that the OV matrix is responsible for copying the $\mathtt{variable}$ factored embeddings of the variable to the RHS position.

\begin{figure}[ht]
    \centering
    \begin{subfigure}[b]{0.48\textwidth}
        \centering
        \includegraphics[width=\linewidth]{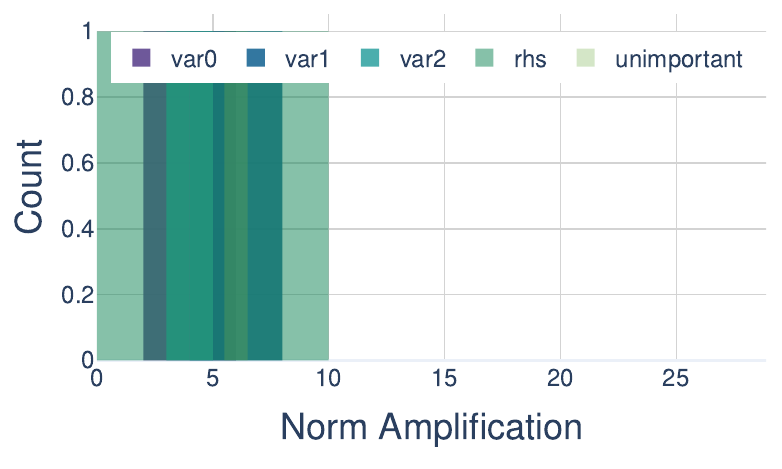}
        \caption{Norm Amplification for \texttt{OPERATION}}
        \label{fig:operation_norm_amp}
    \end{subfigure}
    \hfill
    \begin{subfigure}[b]{0.48\textwidth}
        \centering
        \includegraphics[width=\linewidth]{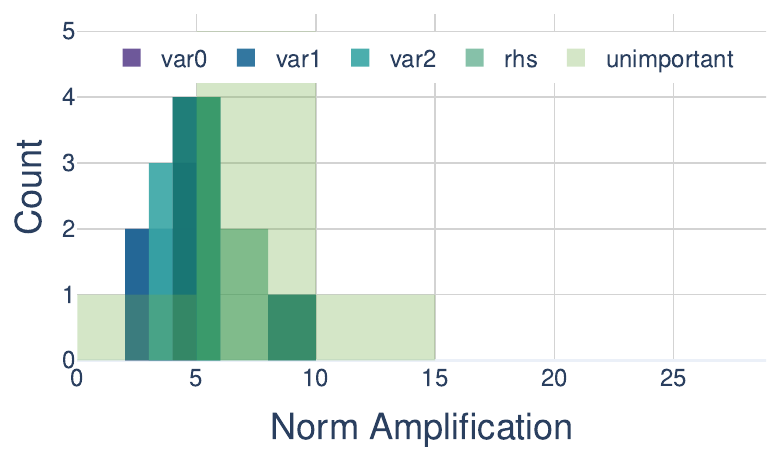}
        \caption{Norm Amplification for \texttt{SYNTAX}}
        \label{fig:syntax_norm_amp}
    \end{subfigure}
    \vspace{0.5em}
    \begin{subfigure}[b]{0.48\textwidth}
        \centering
        \includegraphics[width=\linewidth]{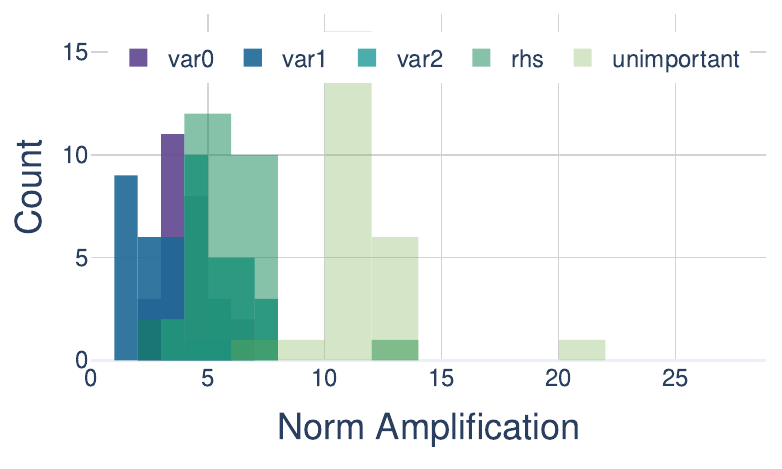}
        \caption{Norm Amplification for \texttt{VALUE}}
        \label{fig:value_norm_amp}
    \end{subfigure}
    \hfill
    \begin{subfigure}[b]{0.48\textwidth}
        \centering
        \includegraphics[width=\linewidth]{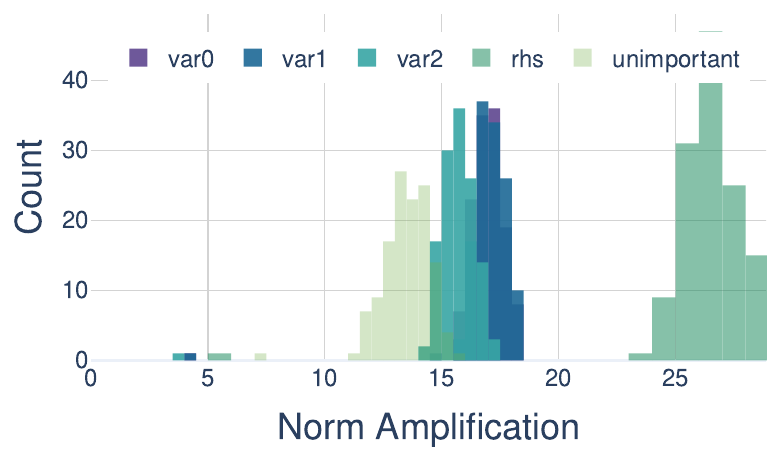}
        \caption{Norm Amplification for \texttt{VARIABLE}}
        \label{fig:variable_norm_amp}
    \end{subfigure}
    \caption{Histogram of norm amplification (defined in \eqref{eq:norm_amplification}) for the embeddings in the four factored embedding types $\{\mathtt{syntax}, \mathtt{variable}, \mathtt{operation}, \mathtt{value}\}$ when passed through the first attention layer's combined OV matrix. Each subfigure contains five histograms in different colors, while each histogram corresponds to a different group of attention heads' combined OV matrix.
    Here, the 16 attention heads are grouped by the different variable they attend to, which are $\token{\mathrm{var0}}$, $\token{\mathrm{var1}}$, $\token{\mathrm{var2}}$, and $\token{\mathrm{rhs}}$, and an additional group for the heads that do not demonstrate a clear pattern.
    }
    \label{fig:factored_embeddings}
\end{figure}

\paragraph{Additional Evidence on change of number of variables.}
We provide one interesting side-observation on how the model handles different numbers of variables in the input equations in \Cref{fig:variable_change_deg2}. Head 4 and head 8 are the two attention heads that attend to the first variable position in the first layer attention when the number of variables is 3.
When the number of variables is changed to 2, we observe that head 4 now attends to the $\sep$ token, while head 8 attends to the equal sign token of the previous equation.
This indicates that the equal sign token and the $\sep$ token act as \emph{attention sink} for head 4 and head 8, respectively.

\begin{figure}[ht]
    \centering
    \begin{subfigure}[b]{0.8\textwidth}
        \centering
        \includegraphics[width=\linewidth]{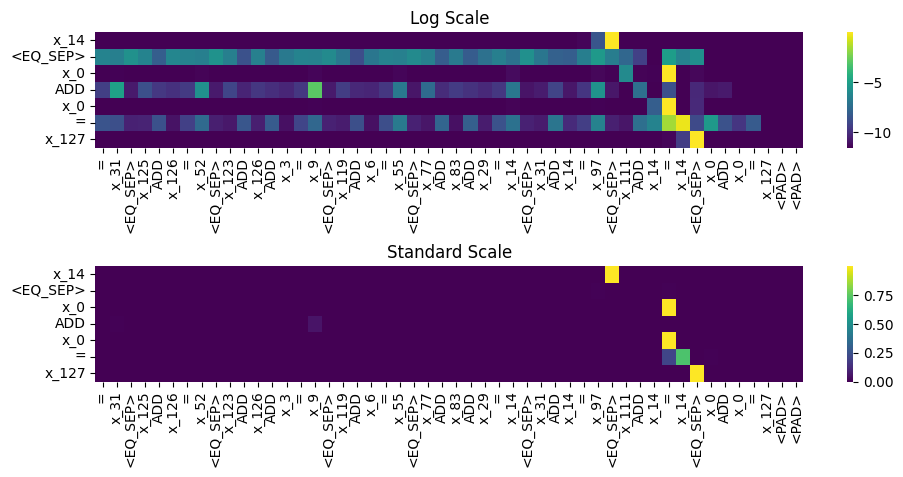}
        \caption{Head 4 for equation with 2 variables}
        \label{fig:head4_deg2}
    \end{subfigure}
    \begin{subfigure}[b]{0.8\textwidth}
        \centering
        \includegraphics[width=\linewidth]{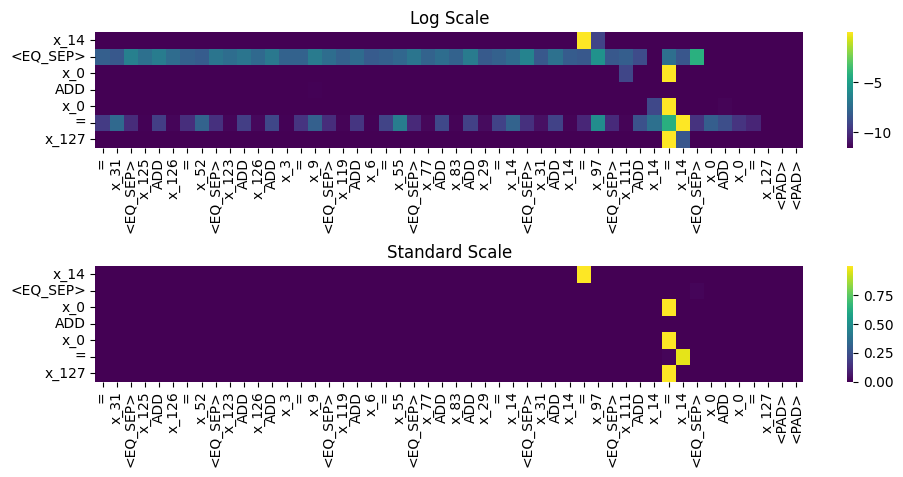}
        \caption{Head 8 for equation with 2 variables}
        \label{fig:head8_deg2}
    \end{subfigure}
    \caption{Visualization of the attention maps for the head group that attends to the first variable position in the first layer attention, which includes head 4 and head 8. Each row corresponds to a different query position, and each column corresponds to a different key position. We only show the rows within the last probe equation, and the columns within the last 50 positions in the sequence. 
    Here, we notice that at the RHS query position (token $\vartoken{x_{127}}$ in the last row), head 4 attends to the $\sep$ token and 
    attention head 8 attends to the equal sign token}
    \label{fig:variable_change_deg2}
\end{figure}

\subsubsection{First Layer MLP Residual Stream Does Not Change the Residual Stream Significantly}

\begin{figure}[h]
    \centering
    \includegraphics[width=0.5\linewidth]{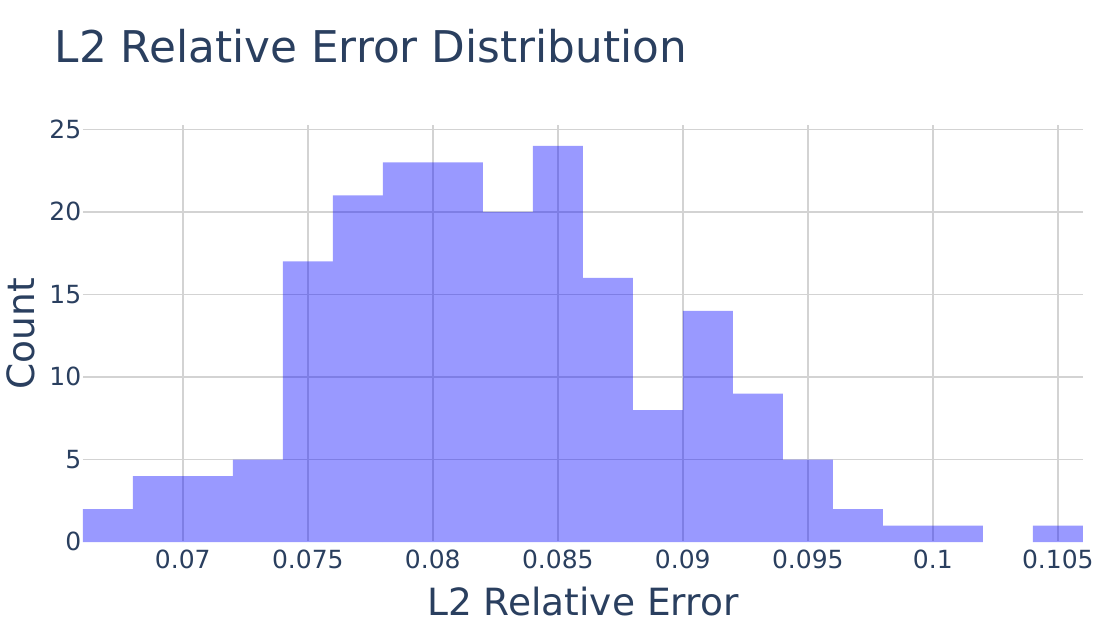}
    \caption{Histogram of the L2 relative error between the residual stream before and after the first layer MLP.}
    \label{fig:l2_relative_error}
\end{figure}
We measure the changes brought by the first MLP layer to the residual stream by computing the Relative L2 Error as:
\begin{align*}
    \text{L2 Relative Error} = \frac{\| \text{Residual Before MLP} - \text{Residual After MLP} \|_2}{\|\text{Residual Before MLP}\|_2}.
\end{align*}
This metric quantifies how much the MLP alters the original residual signal.
\Cref{fig:l2_relative_error} illustrates the heatmap of the relative L2 error computed at the $\RHS$ position across a set of 256 samples.
We observe that the relative L2 error is relatively small, which indicates that the \coloremph{MLP layer does not change the residual stream significantly}.

\subsubsection{Conclusion}

By combining the results from the first layer attention and the first layer MLP, we conclude that the first transformer block plays the following role:
\begin{highlightbox}
\coloremph{\textbf{Mechanistic Interpretation of the First Transformer Block.}}
 At the RHS position, the first transformer block copies the variable identity from the LHS of the equation to the RHS. That is, the variable names of the LHS in the probe equation \eqref{eq:pseudo_equation} are copied to the RHS position, but not their values.
\end{highlightbox}

\subsection{Second Layer Attention: Value Copying}
\begin{figure}[ht]
    \centering
    \begin{subfigure}[b]{0.8\textwidth}
        \centering
        \includegraphics[width=\linewidth]{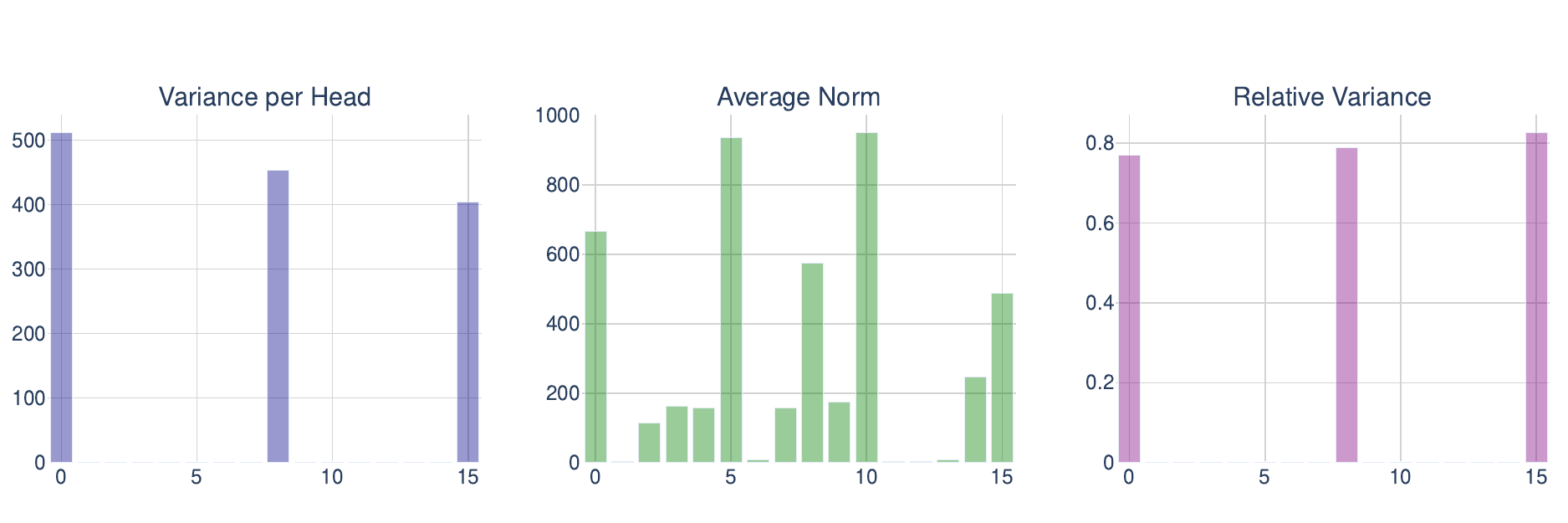}
        \caption{$\VAR{0}$ statistics}
        \label{fig:L1_attention_var0}
    \end{subfigure}
    \begin{subfigure}[b]{0.8\textwidth}
        \centering
        \includegraphics[width=\linewidth]{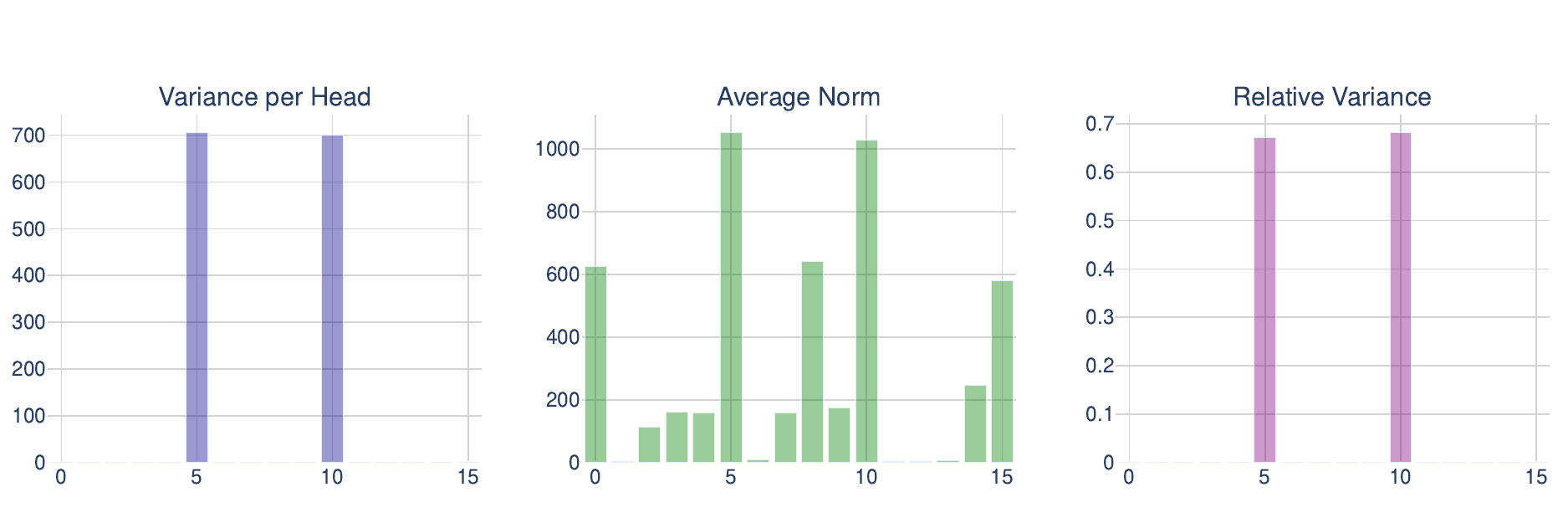}
        \caption{$\VAR{1}$ statistics}
        \label{fig:L1_attention_var1}
    \end{subfigure}
    \begin{subfigure}[b]{0.8\textwidth}
        \centering
        \includegraphics[width=\linewidth]{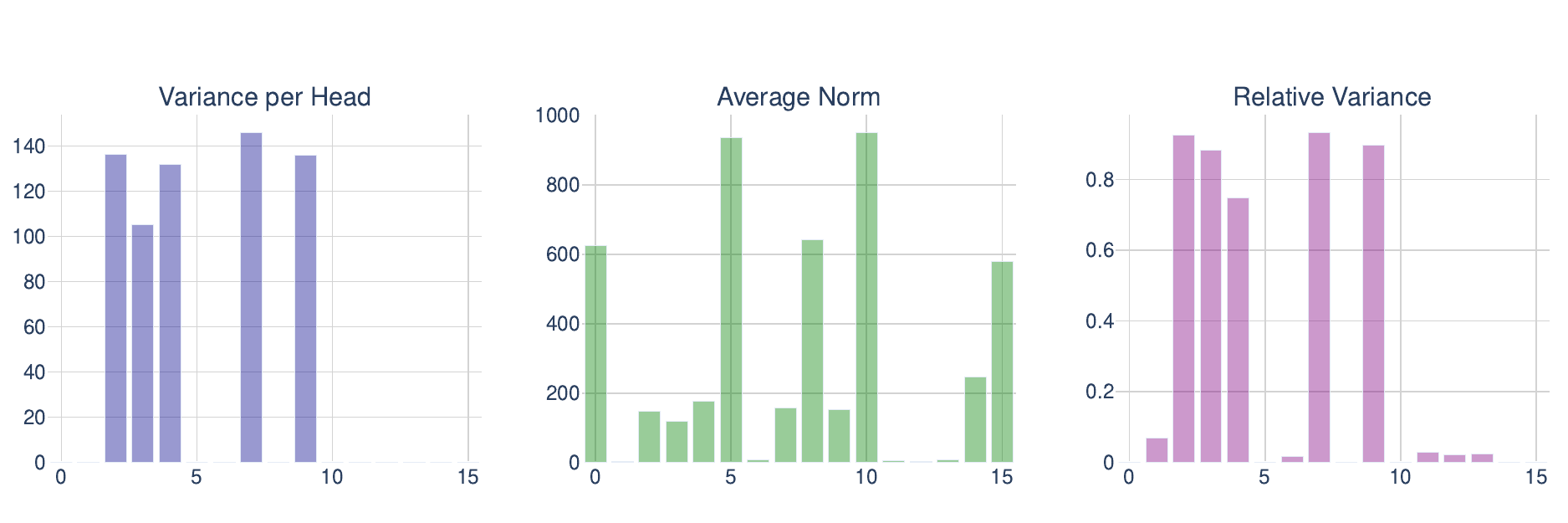}
        \caption{$\VAR{2}$ statistics}
        \label{fig:L1_attention_var2}
    \end{subfigure}
    \caption{Attention head statistics for the second layer attention. Each subfigure shows three histograms corresponding to the variance (numerator to \eqref{eq:relative_variance}), average norm (denominator to \eqref{eq:relative_variance}), and relative variance for each attention head's outputs.}
    \label{fig:L1_attention_statistics}
\end{figure}

\paragraph{A Hypothesis on the Second Layer Attention Heads.}
As we have shown previously, the first layer attention heads copy the $\mathtt{variable}$ factored embeddings of the variable to the RHS position, which tells the model the identity of all the variables on the LHS of the probe equation \eqref{eq:pseudo_equation}. 
To compute the final answer for the RHS position, the model still needs to copy the values of the variables $\VAR{0}$, $\VAR{1}$, and $\VAR{2}$ to the RHS position.
We hypothesize that the second layer attention heads will copy the values of the variables $\VAR{0}$, $\VAR{1}$, and $\VAR{2}$ to the RHS position.

\paragraph{The Second Layer Attention Heads also Have a Group Structure}
To test this hypothesis, we prepare data that contains probe equations of the same form as in \eqref{eq:pseudo_equation} and conduct a controlled experiment designed to analyze how attention heads respond to changes in the values of individual variables. 
Different from the previous experiment where we change the variable identity, this time we fix the variable identity and only change the value of each variable $\VAR{0}$, $\VAR{1}$, and $\VAR{2}$ one at a time while keeping the other two variable values fixed. 
This is achieved by altering the equations before the probe equation, which compute the value of the variable,  to be modified. 
Specifically, for each of the three variables $\VAR{0}$, $\VAR{1}$, and $\VAR{2}$, we conduct a separate experiment where we collect $N$ samples by varying only that variable's value while keeping the other two variables fixed. Then, for each variable $\VAR{~i}$ with $i=0,1,2$, we collect the second layer attention head outputs across the $N$ samples at the RHS position of the probe equation, and compute the following metrics:
\begin{itemize}
    \item The variance of the outputs (numerator in \eqref{eq:relative_variance})
    \item The average squared norm (denominator in \eqref{eq:relative_variance}) 
    \item The relative variance (ratio of the above quantities)
\end{itemize}
These metrics help us identify which heads are sensitive to changes in each variable's value.
The results are shown in \Cref{fig:L1_attention_statistics}.
We deduce from the results (especially the relative variance plots  in the right column) that the heads \{0, 8, 15\} form the first group, which copy the value for $\VAR{0}$; heads \{5, 10\} form the second group, which copy the value for $\VAR{1}$; and heads \{2, 3, 4, 7, 9\} form the third group, which copy the value for $\VAR{2}$. We denote these three groups as $\mathcal{H}_0$, $\mathcal{H}_1$, and $\mathcal{H}_2$, respectively to simplify the notation.

\paragraph{Second Layer Attention Heads Copy the Values of the Variables to the RHS Position.}
Similar to the experiment in the first layer, we compute the norm amplification coefficients for the OV matrices of the second layer attention heads, combined by groups, as shown in \Cref{fig:factored_amp_l1}.
We observe that the norm amplification coefficients for the $\mathtt{value}$ factored embedding are significantly larger than that of the other types of embeddings. 
This can be seen by noticing that the bottom left subfigure has a larger magnitude for all colors. This confirms our hypothesis that the OV matrix is responsible for copying the $\mathtt{value}$ factored embeddings of the variable to the RHS position.


\begin{figure}[ht]
    \centering
    \begin{subfigure}[b]{0.48\textwidth}
        \centering
        \includegraphics[width=\linewidth]{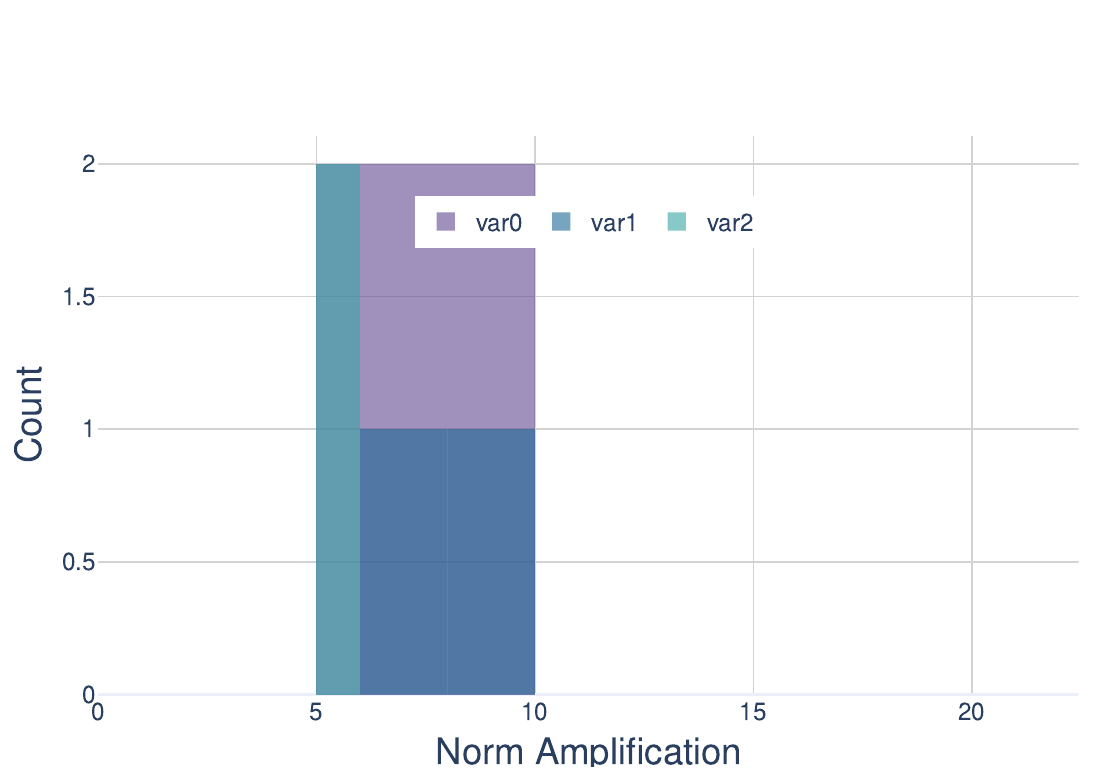}
        \caption{Norm Amplification for \texttt{OPERATION}}
        \label{fig:operation_norm_amp_l1}
    \end{subfigure}
    \hfill
    \begin{subfigure}[b]{0.48\textwidth}
        \centering
        \includegraphics[width=\linewidth]{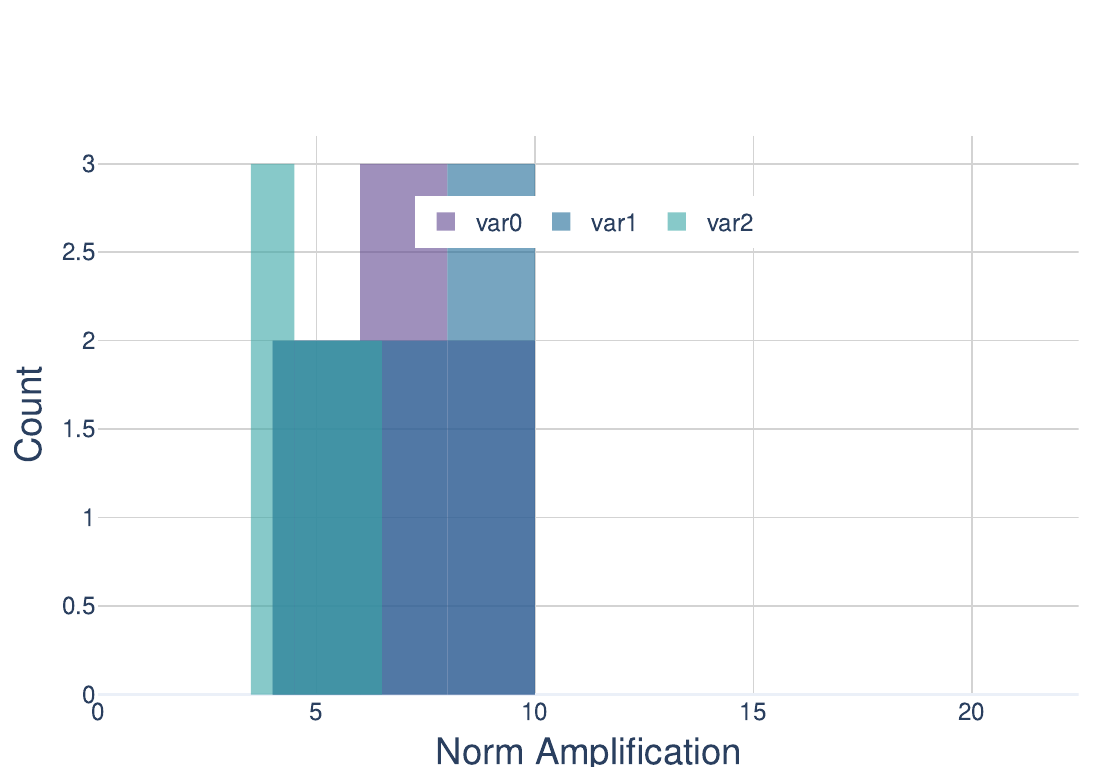}
        \caption{Norm Amplification for \texttt{SYNTAX}}
        \label{fig:syntax_norm_amp_l1}
    \end{subfigure}
    \vspace{0.5em}
    \begin{subfigure}[b]{0.48\textwidth}
        \centering
        \includegraphics[width=\linewidth]{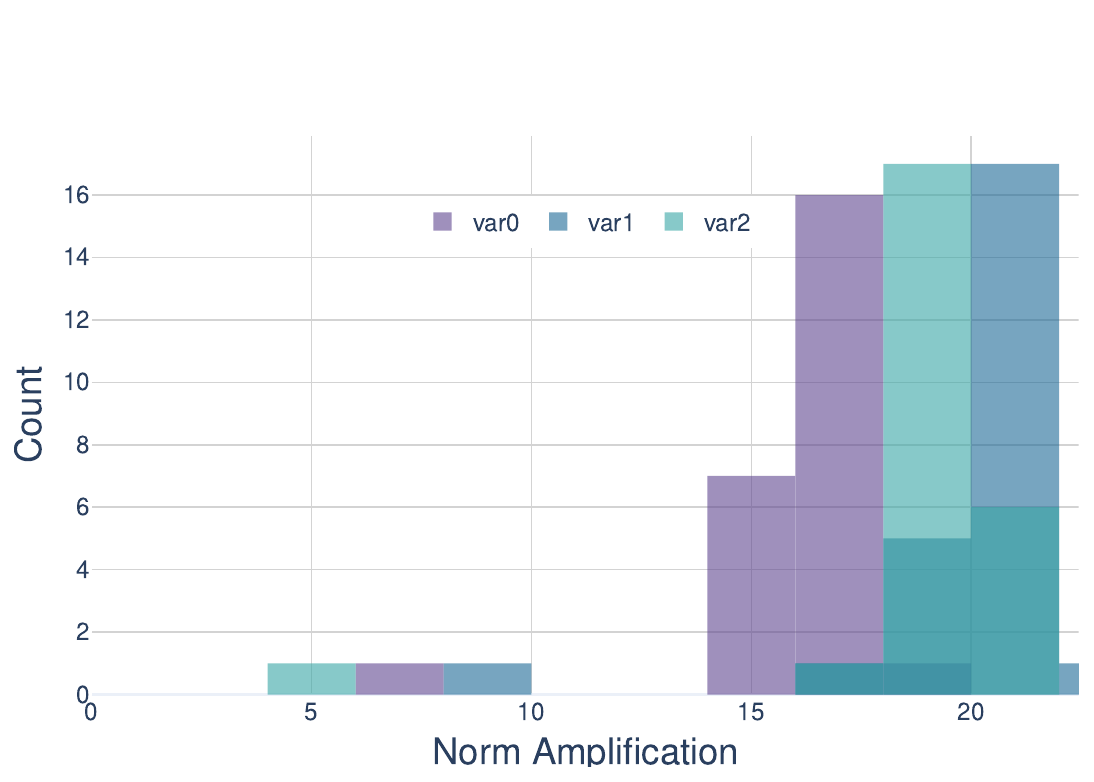}
        \caption{Norm Amplification for \texttt{VALUE}}
        \label{fig:value_norm_amp_l1}
    \end{subfigure}
    \hfill
    \begin{subfigure}[b]{0.48\textwidth}
        \centering
        \includegraphics[width=\linewidth]{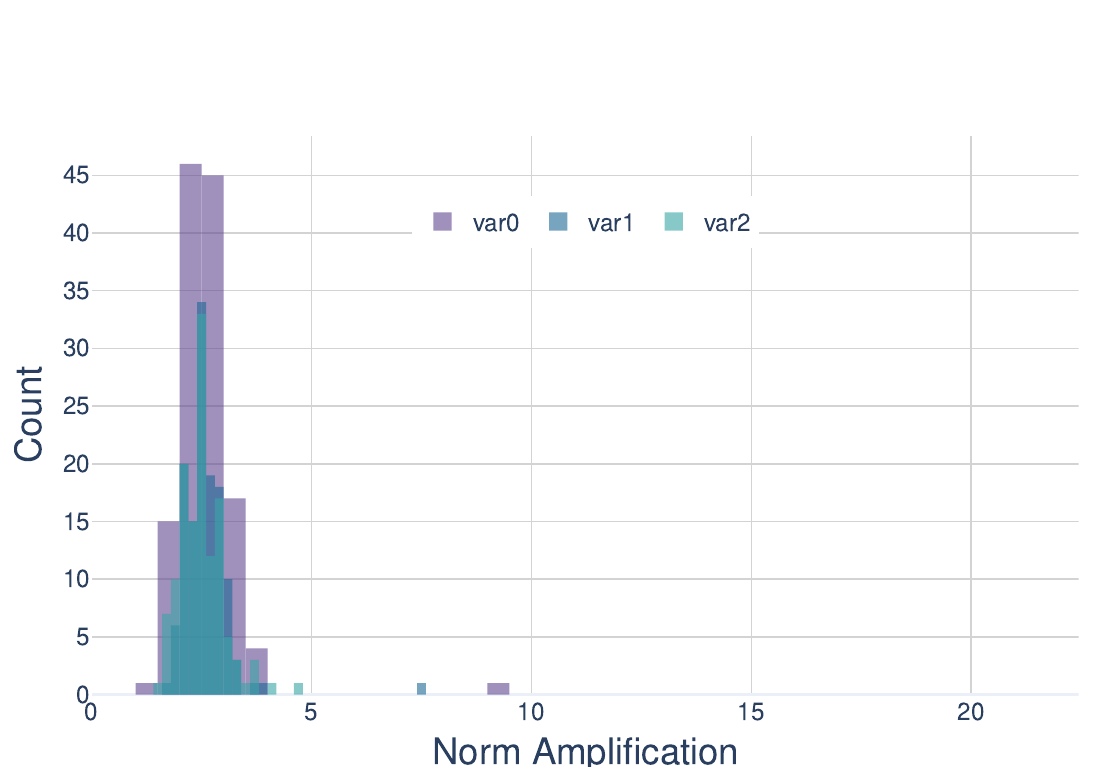}
        \caption{Norm Amplification for \texttt{VARIABLE}}
        \label{fig:variable_norm_amp_l1}
    \end{subfigure}
    \caption{Histograms of norm amplification for the four factored embedding types in the second layer attention's OV matrix. The 16 attention heads are grouped by the variable they attend to, which are $\token{\mathrm{var0}}$, $\token{\mathrm{var1}}$, and $\token{\mathrm{var2}}$.
    In each subfigure, we make three histograms each corresponding to the combined OV matrix for each group of attention heads. The three histograms in each subfigure are shown in different colors, and each histogram is for all the embeddings of the corresponding factored embedding type. 
    It can be observed that the amplification factor for the ``\texttt{value}'' factored embedding is significantly larger than that of the other types of embeddings, confirming our hypothesis that the OV matrix is responsible for copying the ``\texttt{value}'' factored embeddings of the variable to the RHS position.
    }
    \label{fig:factored_amp_l1}
\end{figure}

\subsubsection{Conclusion}
For the second layer attention, we have the following mechanistic interpretation:
\begin{highlightbox}
\coloremph{\textbf{Mechanistic Interpretation of the Second Layer Attention.}}
 The second layer attention heads copy the values of the variables $\VAR{0}$, $\VAR{1}$, and $\VAR{2}$ to the RHS position. This is achieved by using the OV matrices of the second layer attention heads to write the \texttt{value} factored embeddings of the three variables on the LHS of \eqref{eq:pseudo_equation} to the residual stream at the RHS position.
\end{highlightbox} 

\subsection{Second Layer MLP: Module Addition in the Frequency Domain} \label{sec:interpret_mlp_detail}

To study how the second layer MLP performs the modular addition operation, we first understand the structure of the input to the second layer MLP.

\subsubsection{Understanding the Input Structure to the Second Layer MLP}

Having confirmed that the second layer attention successfully copies the values of variables to the RHS position, we now examine in detail how these value embeddings are structured and prepared for the MLP's arithmetic computation.  
The key insight is that the second layer attention   transforms the value embeddings through head-specific OV matrices, creating a structured representation that facilitates downstream computation. To understand this transformation, we need to consider three aspects: the original embedding space, the transformation process, and the resulting geometric structure.

\textbf{Original Value Embedding Space.} In our factored tokenization scheme, each token is decomposed into four factors: \texttt{syntax}, \texttt{variable}, \texttt{operation}, and \texttt{value}. The \texttt{value} factor encompasses 25 distinct elements: the integers 0 through 22 (used for actual arithmetic computations in modulo-23 arithmetic), plus two special tokens --- \texttt{empty} (indicating a variable whose value has not been computed yet) and \texttt{N/A} (for non-value tokens like operators). Each of these 25 possible values has a learned embedding vector $\mathtt{embed}_{\mathtt{value}}(v) \in \mathbb{R}^{d_{\text{model}}}$ where $v \in \{0, 1, \ldots, 22, \texttt{empty}, \texttt{N/A}\}$. These embeddings are learned during training and form the basis for how numerical values are represented throughout the network.

\textbf{Head Group Transformation Process.} As we established through our variance analysis, the second layer attention heads organize into three distinct groups based on which variable position they attend to: $\mathcal{H}_0 = \{0, 8, 15\}$ attends to $\VAR{0}$, $\mathcal{H}_1 = \{5, 10\}$ attends to $\VAR{1}$, and $\mathcal{H}_2 = \{2, 3, 4, 7, 9\}$ attends to $\VAR{2}$. 
As a result, when a value embedding passes through this transformation, it becomes what we call a ``new value'' embedding defined in \eqref{eq:new_value_def}, where $W_{OV}^{(\mathcal{H}_i)}$ is the OV combined matrix for the head group $\mathcal{H}_i$, $i \in \{0, 1, 2\}$. 
Intuitively, $\mathtt{new\_value}_0(v)$ is the contribution of $\VAR{0}$ with value $v$ to the residual stream at the RHS position after the second layer attention.

By enumerating all the possible values for each variable, we have in total 75 distinct new value embeddings:   
\begin{itemize}
    \item Vectors 1-25: $\{\mathtt{new\_value}_0(v) : v \in \{0, \ldots, 22, \texttt{empty}, \texttt{N/A}\}\}$ for $\VAR{0}$
    \item Vectors 26-50: $\{\mathtt{new\_value}_1(v) : v \in \{0, \ldots, 22, \texttt{empty}, \texttt{N/A}\}\}$ for $\VAR{1}$
    \item Vectors 51-75: $\{\mathtt{new\_value}_2(v) : v \in \{0, \ldots, 22, \texttt{empty}, \texttt{N/A}\}\}$ for $\VAR{2}$
\end{itemize}

During actual computation, when processing a specific probe equation, the second layer attention selects and combines exactly three of these 75 vectors based on the actual values of the variables. For instance, consider the probe equation $x_{91} + x_{88} + x_{55} = x_{30}$ where $x_{91}$ has value 7, $x_{88}$ has value 15, and $x_{55}$ has value 3. The second layer attention performs the following:
\begin{enumerate}
    \item [(i)]Head group $\mathcal{H}_0$ attends to the position containing $x_{91}$ and copies its value embedding, transforming it to produce $\mathtt{new\_value}_0(7)$
    \item [(ii)]Head group $\mathcal{H}_1$ attends to the position containing $x_{88}$ and produces $\mathtt{new\_value}_1(15)$
    \item [(iii)]Head group $\mathcal{H}_2$ attends to the position containing $x_{55}$ and produces $\mathtt{new\_value}_2(3)$
     
\end{enumerate}
The residual stream at the RHS position then contains the vector sum:
\begin{align}
    \text{MLP Input} = \mathtt{new\_value}_0(7) + \mathtt{new\_value}_1(15) + \mathtt{new\_value}_2(3) + \text{other residual components}
\end{align}
This sum of three transformed value embeddings forms the primary input signal that the second MLP must decode to compute $(7 + 15 + 3) \bmod 23 = 2$.

\textbf{Geometric Structure Analysis.} To understand how these 75 vectors are organized in the embedding space and how this organization facilitates modular arithmetic, we analyze their pairwise relationships through cosine similarity,  shown in \Cref{fig:cosine_sim_heatmap}.

\begin{wrapfigure}{r}{0.35\textwidth}
    \centering
    \vspace{-15pt}
    \includegraphics[width=\linewidth]{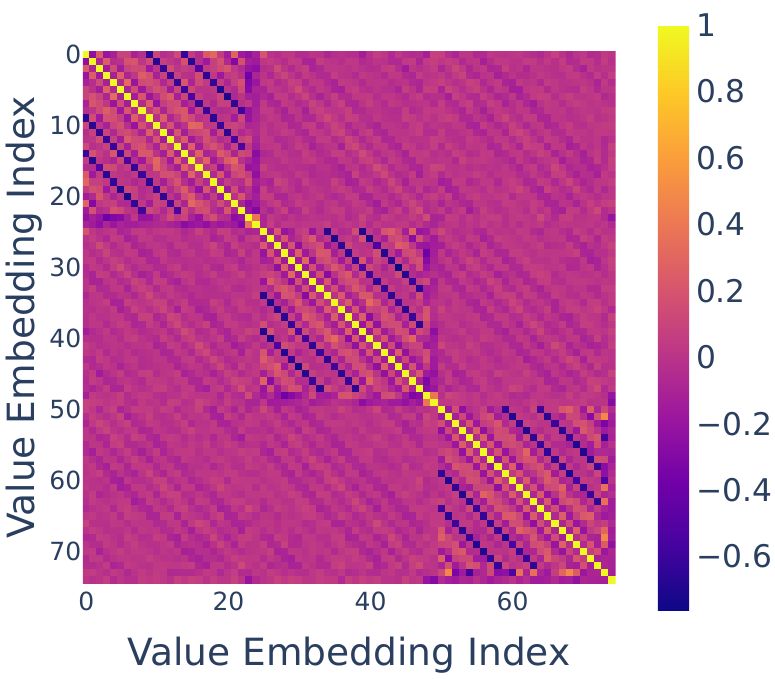}
    \caption{\small Cosine similarity of the $\mathtt{new~value}$ factored embeddings for all three variables in the residual stream after the second layer attention.}
    \label{fig:cosine_sim_heatmap}
    \vspace{-25pt}
\end{wrapfigure}
\Cref{fig:cosine_sim_heatmap} exhibits two salient structures:
\begin{itemize}
    \item \textbf{\coloremph{Inter-variable near-orthogonality}.} Cosine similarities between \texttt{value}-factored embeddings belonging to \emph{different} variables are near zero. Equivalently, entries outside the three $25\times 25$ block-diagonal submatrices are close to zero.
    \item \textbf{\coloremph{Intra-variable circulant structure}.} Within each variable, the first $23$ \texttt{value}-factored embeddings (actual values) form an approximately circulant similarity submatrix: each row is a cyclic right shift of the previous row. The last two embeddings, \texttt{empty} and \texttt{N/A}, are nearly orthogonal to those $23$ (and to each other), yielding two low-similarity rows/columns appended to the block.
\end{itemize}
Concretely, if we restrict attention to any one of the three blocks, the top-left $23\times 23$ portion displays the circulant pattern. A circulant matrix is determined by its first row and is diagonalized by the discrete Fourier basis, a fact we will use when interpreting its spectrum.



\subsubsection{Module Addition in the Frequency Domain.}
To systematically analyze how the model performs the module addition operation, we prepare a equation of the form as in \eqref{eq:pseudo_equation}, and we change the previous equations to alter the value of each variable $\VAR{0}$, $\VAR{1}$, and $\VAR{2}$. 
Specifically, we let $\VAR{0}$, $\VAR{1}$, and $\VAR{2}$ to iterate over the set $\{0, 1, 2, \ldots, 22\}$ since the model is trained on modular-23 addition.
To study how the MLP performs the module addition operation, we pick the following four positions in the model: (i) pre-activation of the second layer's MLP, (ii) post-activation of the second layer's MLP, (iii) the output of the second layer's MLP, and (iv) the model's decoder output. 

For each of these network positions, we take the corresponding vector in the residual stream at the RHS position of the probe equation. To simplify the notation, we denote such a vector as $v(x, y, z)$ and let $d$ denote its dimension, when the input variables are $\VAR{0}=x$, $\VAR{1}=y$, and $\VAR{2}=z$.
We then compute the three-dimensional 23-point Discrete Fourier Transform (DFT) applied independently to each coordinate of $v$ over $(x, y, z)$, which is defined as:
\begin{align*}
    \mathrm{DFT}_3(v)_{j,k,l} = \frac{1}{\sqrt{23^3}} \sum_{x=0}^{22} \sum_{y=0}^{22} \sum_{z=0}^{22} v(x,y,z) \, e^{-2\pi i \frac{x j + y k + z l}{23}}, \quad j,k,l = 0, 1, \ldots, 22,
\end{align*}
Here $j, k, l$ are the frequencies in the three dimensions, and we apply the DFT to $v$ coordinatewisely. Thus, for each $(j, k, l)$,  $\mathrm{DFT}_3(v)_{j,k,l} $ is a $d$-dimensional complex vector, and $\mathrm{DFT}_3(v)$ is a four-dimensional tensor with dimension $23^3\times d$.
We then compute the norm of the DFT tensor along the last dimension for each $(j, k, l)$, which represents the magnitude of the corresponding frequency component.
Since the obtained DFT tensor is conjugate symmetric, we have
\begin{align*}
    \mathrm{DFT}_3(v)_{j,k,l} = \overline{\mathrm{DFT}_3(v)_{22-j,22-k,22-l}},
\end{align*}
Therefore, we only need to focus on the first half of the tensor, which has dimension $12^3$.

\paragraph{Studying the DFT Tensor by Frequency Group.}
We further partition the tensor into 7 groups by the algebraic patterns of the frequency component $(j, k, l)$:
\begin{itemize}
    \item Group 1: $(0, 0, 0)$
    \item Group 2: $(0, 0, a)$, $(0, a, 0)$, $(a, 0, 0)$ for $a \neq 0$
    \item Group 3: $(0, a, b)$, $(a, 0, b)$, $(a, b, 0)$ for nonzero $a \neq b$
    \item Group 4: $(0, a, a)$, $(a, 0, a)$, $(a, a, 0)$ for $a \neq 0$
    \item Group 5: $(a, b, c)$ for nonzero $a \neq b$, $b \neq c$, $c \neq a$
    \item Group 6: $(a, a, b)$, $(a, b, a)$, $(b, a, a)$ for nonzero $a \neq b$
    \item Group 7: $(a, a, a)$ for $a \neq 0$
\end{itemize}
We plot the histograms of the norm of the DFT tensor in the last dimension for each group, as shown in \Cref{fig:fft_combined}. 
At the pre-activation stage of the second layer MLP, the DFT tensor shows its highest norm for the group $(0, 0, 0)$, which suggests a dominant bias term that is independent of the input variables. 
Progressing from the pre-activation (\Cref{fig:fft_l1_mlp_preact}) to the MLP output (\Cref{fig:fft_l1_mlp_output}), this bias term gradually diminishes, while the norm corresponding to the group $(a, a, a)$ steadily increases. 
This trend indicates that the MLP output contains a strong frequency component of the form 
\begin{align}
\cos\left(\frac{2\pi a x}{23}\right) \cdot \cos\left(\frac{2\pi a y}{23}\right) \cdot \cos\left(\frac{2\pi a z}{23}\right),
\label{eq:fft_comp}
\end{align}
or a similar combination involving both sine and cosine functions with the same frequency $a$. In \eqref{eq:fft_comp}, $x$, $y$, and $z$ denote the value of the three variables in the equation, and $a$ is the frequency.
The term in \eqref{eq:fft_comp} corresponds to a degree-3 term on frequency $a$, indicating that the model is capable of computing terms in the form of $\cos(2\pi a (x+ y + z)/23 + \varphi)$ for some frequencies $a$ and phase $\varphi$, and eventually decodes to the correct answer $x+y+z \mod{23}$.

\subsubsection{Conclusion}
For the second layer MLP, we have the following mechanistic interpretation:
\begin{highlightbox}
\coloremph{\textbf{Mechanistic Interpretation of the Second Layer MLP.}}
The second layer MLP performs modular addition in the frequency domain. It takes as input the sum of three transformed value embeddings ($\mathtt{new\_value}_0$, $\mathtt{new\_value}_1$, $\mathtt{new\_value}_2$) from the second layer attention and computes their modular sum through sinusoidal basis functions. The MLP progressively amplifies diagonal frequency components of the form $(a, a, a)$ with $a \in \{1, \ldots, 22\}$, which encode the sum $x + y + z \bmod 23$, while suppressing the bias term $(0, 0, 0)$, ultimately enabling the decoder to output the correct result.
\end{highlightbox}

\begin{figure}[ht]
    \centering
    \begin{subfigure}[b]{0.7\textwidth}
        \centering
        \includegraphics[width=\linewidth]{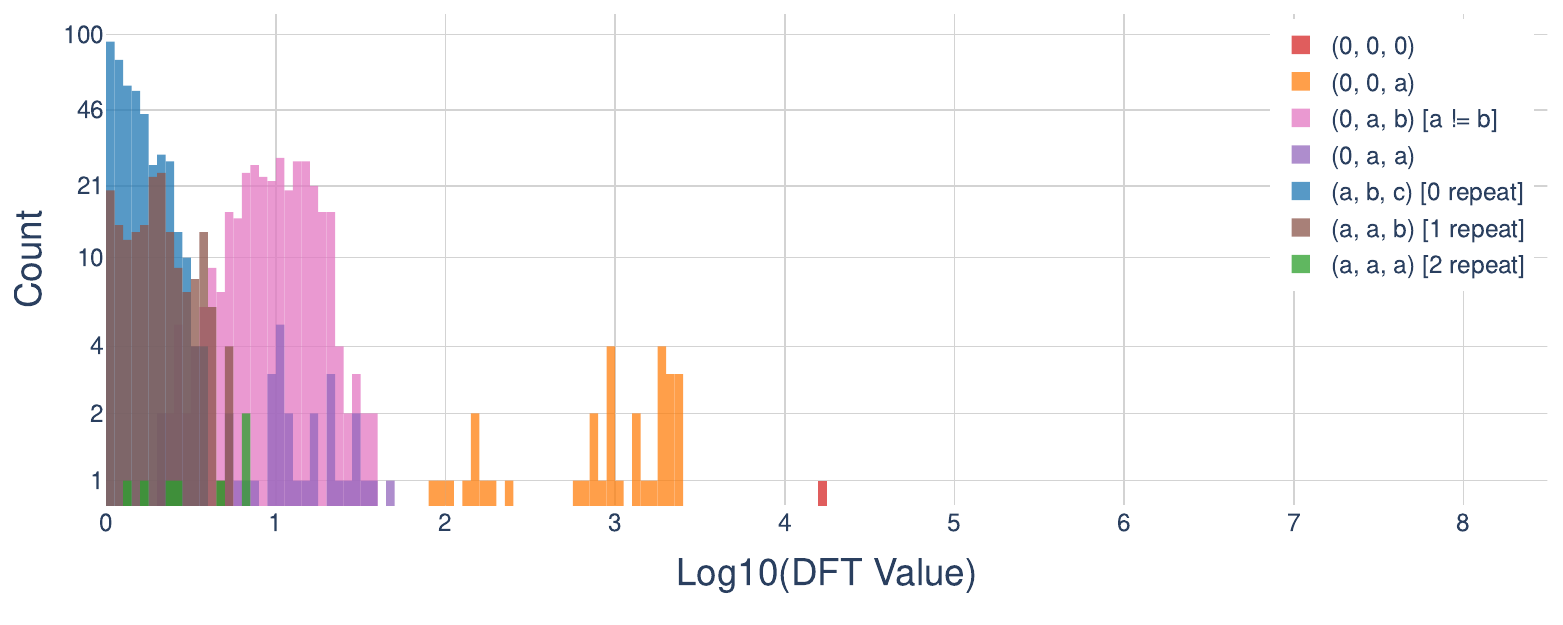}
        \caption{DFT of L1 MLP pre-activation}
        \label{fig:fft_l1_mlp_preact}
    \end{subfigure}
    \begin{subfigure}[b]{0.7\textwidth}
        \centering
        \includegraphics[width=\linewidth]{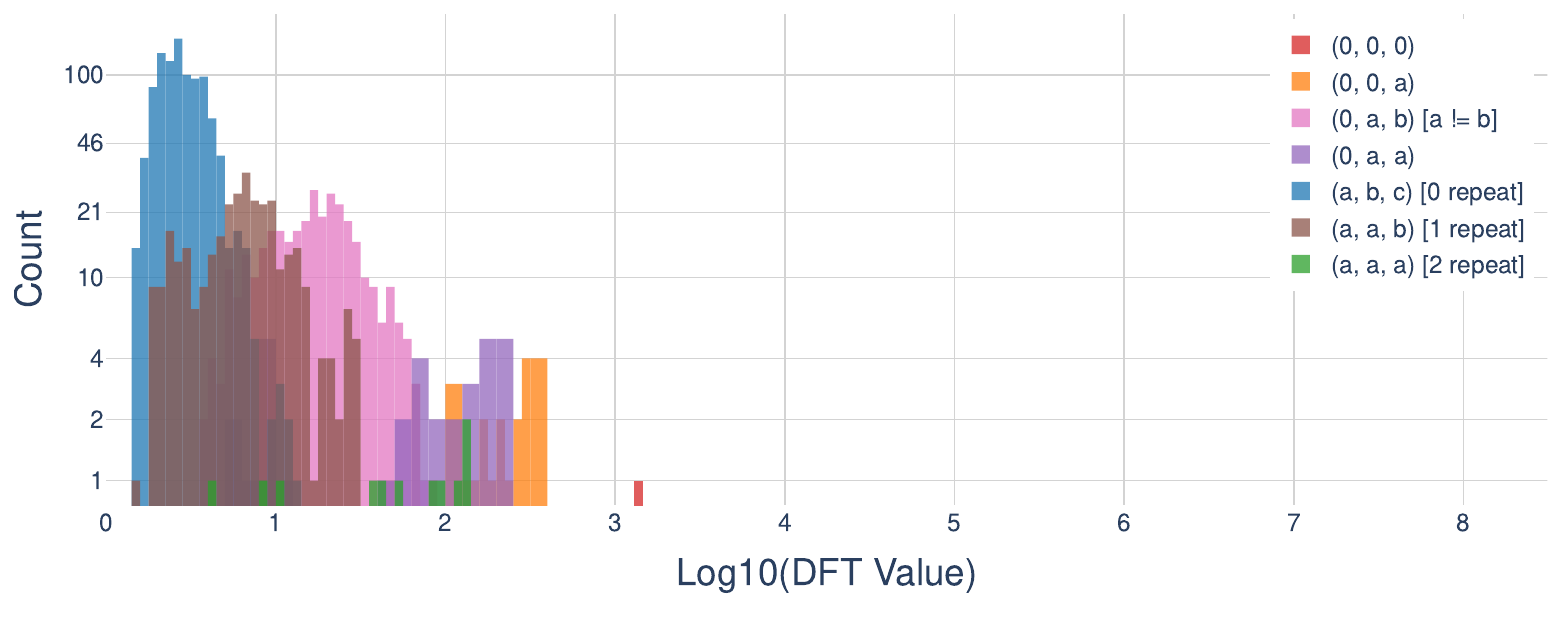}
        \caption{DFT of L1 MLP post-activation}
        \label{fig:fft_l1_mlp_postact}
    \end{subfigure}
    
    
    \begin{subfigure}[b]{0.7\textwidth}
        \centering
        \includegraphics[width=\linewidth]{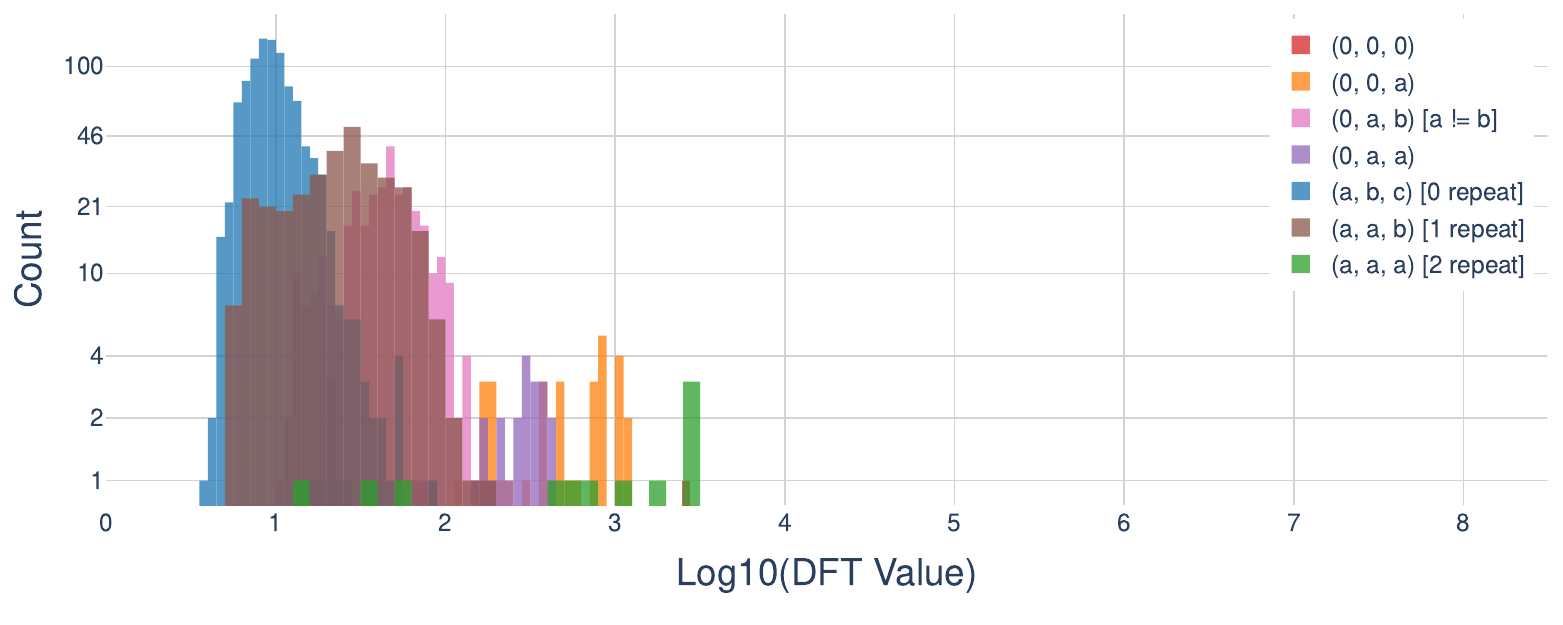}
        \caption{DFT of L1 MLP output}
        \label{fig:fft_l1_mlp_output}
    \end{subfigure}
    \hfill
    \begin{subfigure}[b]{0.7\textwidth}
        \centering
        \includegraphics[width=\linewidth]{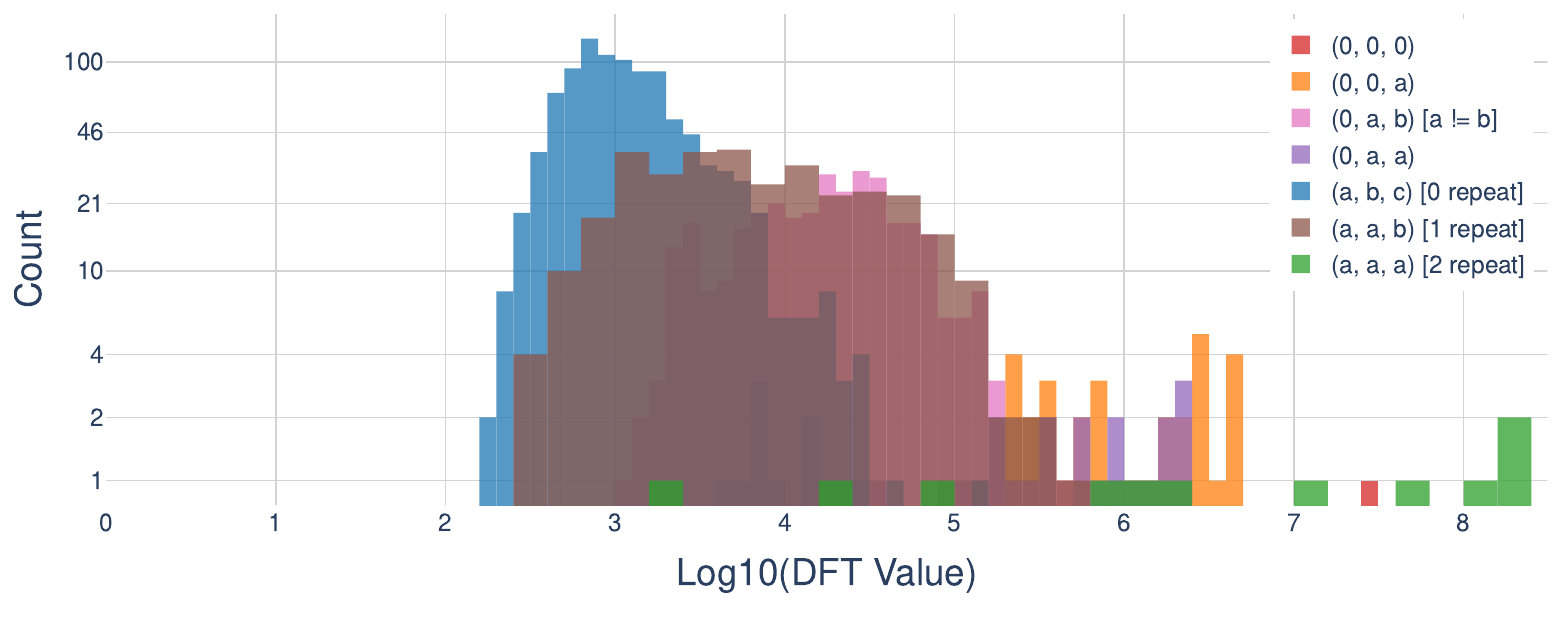}
        \caption{DFT of decoder output}
        \label{fig:fft_decoder_output}
    \end{subfigure}
    \caption{Combined DFT histograms for the second layer MLP pre-activation, MLP post-activation, MLP output, and decoder output.}
    \label{fig:fft_combined}
\end{figure}

\subsection{Error Analysis}
\aawarning{May need to update fig x-axis to make more clear.}

To better understand the probed model's performance, we analyze its prediction errors. 
As we have three functional components in the model --- the first layer attention, the second layer attention, and the last feedforward layer --- we consider three sources of errors: \textit{(\romannumeral1)} the first layer's attention mapping copies from the wrong variable position, \textit{(\romannumeral2)} the second layer's attention fails to copy the correct variable value, and \textit{(\romannumeral3)} the feedforward layer miscalculates the sum of the LHS variables. 
An account of the errors by source is shown in \Cref{tab:error-attribution}, where the major source of error is the feedforward layer calculation.
Note that when considering the three sources of errors, if the error \textit{(\romannumeral1)} occurs, we do not  count towards error \textit{(\romannumeral2)} and \textit{(\romannumeral3)}.
Similarly, when error \textit{(\romannumeral2)} occurs, we don't count towards error \textit{(\romannumeral3)}.
In the following, we details how we identify the three sources of errors.

\subsubsection{Identifying Different Sources of Errors}
When our recurrent transformer model is computing the RHS value for all the equations in the sequence, we have two key concepts: 
\begin{itemize}
    \item \textbf{Depth of equation}: The depth of an equation is the number of iterations required to compute the correct RHS value. More formally, the depth of an equation is the depth of the RHS variable in the computation graph. Take \Cref{fig:dag_example} as an example, the depth of the equation ``$20=x_7$'' is $1$, as the model only needs a single loop to compute the correct RHS value, and the depth of the equation ``$x_7 + x_{42}= x_{23}$'' is $2$, as the model needs two loops to compute the correct RHS value.
    \item \textbf{Number of iterations}: The number of iterations describes how many times the loop transformer model has iterated over the input sequence.
\end{itemize}
By definition, the minimum number of iterations needed for computing the correct RHS value of an equation of depth $d$ is at least $d$.
In fact, we observe that most of the equations can be computed with exactly the number of iterations equal to the depth.
For this reason, we only consider the equations and the number of iterations such that 
\begin{align} 
    \mathtt{depth~of~equation} \ge \mathtt{number~of~iterations}, \quad \text{or for short,}\quad \mathtt{depth} \ge \mathtt{iter}.
    \label{eq:depth_ge_iter}
\end{align}
Moreover, we do not add any probe equations in this error analysis. 
This means that we apply the knowledge learned from the previous experiments with probe equations to identify errors happening in the whole sequence.

In the following, we details how we identify the three sources of errors.

\begin{wraptable}[11]{r}{0.48\textwidth}
    \centering
    \caption{Attribution of errors by source in the testing dataset with $N=128$ and 23k sentences.}
    \label{tab:error-attribution}
    \begin{tabular}{l c}
        \toprule
        Error Source & Count \\
        \midrule
        First Layer Attention Error & 9 \\
        Second Layer Copy Error    & 1 \\
        Feedforward Calculation Error & 30 \\
        \midrule
        \textbf{Total} & \textbf{40} \\
        \bottomrule
    \end{tabular}
\end{wraptable}
\paragraph{First Layer Attention Error.} We identify first layer attention errors by analyzing how well each attention head group focuses on its assigned variable position. For each equation's RHS position, we examine the attention map (an $H\times L \times L$ tensor, where $H$ is the number of attention heads and $L$ is the sequence length) to extract the relevant attention probabilities.

    Consider a concrete example: For the head group $\calH_0$ that is responsible for attending to $\VAR{0}$, we look at the attention probabilities where the query is at the $\RHS$ position and the key is at the $\VAR{0}$ position, for all heads in $\calH_0$. We then average these probabilities within the head group.

    For each equation, we can use the above strategy to obtain a single \textbf{group-wise attention probability} for each head group at the $\RHS$ position.
    If this group-wise attention probability is less than our threshold of $0.9$, we classify it as a first layer attention error,  indicating that the head group failed to maintain sufficient focus on its designated variable position.
    In fact, the error analysis is not very sensitive to the choice of the threshold. 
    As we will see later in \Cref{fig:error-analysis} (Top Row), the computed group-wise attention probability is either very close to $1$ or very close to $0$ (for $\VAR{0}$ and $\VAR{2}$, where $\VAR{1}$ has a slightly larger deviation from $1$ on the high end). It is very easy to identify when an error occurs in the first layer attention.

\paragraph{Second Layer Copy Error.} For the second layer attention, we analyze the attention head's output rather than the attention map. This approach is necessary because the ``$\mathtt{value}$'' factored embedding from the first layer may be distributed across multiple positions, including special tokens (like delimiters or operators), rather than being confined to the original variable position.
Fortunately, we already have the extracted ``$\mathtt{new~value} (i)$'' factored embeddings for each $\VAR{~i}$ in the previous experiment. We thus treat these ``$\mathtt{new~value} (i)$'' factored embeddings as the ground truth value embeddings for $\VAR{~i}$ in the second layer attention output. 

    For each equation containing $\VAR{~i}$, we compute the cosine similarity between the ground truth value embedding for $\VAR{~i}$ and the designated head group's output at the $\RHS$ position in the second layer attention.
    We call this cosine similarity the ``group-wise cosine similarity''.
    If the cosine similarity is less than our pre-determined threshold of $0.9$, we consider it a second layer copy error for that head group, indicating the model fails to copy the correct variable value to the RHS position.

    Similar to the first layer attention analysis, the choice of threshold is not critical.
    As shown in \Cref{fig:error-analysis} (Middle Row), the cosine similarity between the second layer attention outputs and the target value embeddings exhibits a clear pattern: either very close to $1$ for correct copies, or significantly lower for incorrect copies.
    This stark separation makes it straightforward to identify second layer copy errors.

\paragraph{Feedforward Calculation Error.} The feedforward calculation error is defined in the following way: If an equation passes the first two error checks, meaning that the first layer attention successfully attends to the correct variable position, and the second layer attention successfully copies the correct variable value to the RHS position, but the model still makes a mistake when applying the factored decoder after the second layer MLP, we consider it a feedforward calculation error.

An account of the errors by source is shown in \Cref{tab:error-attribution}, where the major source of error is the feedforward layer calculation.
Overall, the model demonstrates a remarkable accuracy, where the total number of errors is only $40$ out of 23k examples.
A more detailed analysis of the errors is shown in \Cref{fig:error-analysis}.

\subsubsection{Additional Error Analysis}
Here, we provide additional evidence for the above discussion.
In \Cref{fig:error-analysis}, instead of just counting the number of times a specific error occurs, we histogram all the statistics used by the above error analysis procedure.
\Cref{fig:error-analysis} (Top Row) is a histoplot of the the group-wise attention probability in the first layer, organized by three head groups $\calH_0$, $\calH_1$, and $\calH_2$, where each $\calH_i$ is responsible for copying the value of $\VAR{~i}$.
See the ``First Layer Attention Error'' paragraph above for more details.
We see that the attention scores generally concentrate their probability mass around $1$ on the correct variable; however, the heads responsible for copying $\VAR{2}$ are somewhat less concentrated, resulting in more errors.
Moreover, for some examples where the final prediction is incorrect, we observe a clear error pattern in the histogram: the attention head group completely fails to attend to the correct variable position, with the group-wise attention probability dropping to nearly $0$. This stark contrast between successful and failed attention patterns makes it easy to identify first layer attention errors.

In addition, \Cref{fig:error-analysis} (Middle Row) histogram the cosine similarity between the second layer attention outputs and the target value embedding, again for all three head groups.
For most examples, the cosine similarity is close to $1$, showing that the second layer retrieves the value embeddings.
However, for some examples where the final prediction is incorrect, we also observe a clear error pattern in the histogram: the cosine similarity drops to nearly $0$.
This stark contrast between successful and failed second layer copy patterns makes it easy to identify second layer copy errors as well.

\begin{figure}[h]
    \centering
    \newcommand{\imageext}{png} 
    \begin{tabular}{ccc}
        \includegraphics[width=0.3\linewidth]{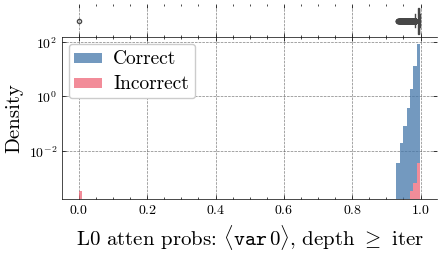} &
        \includegraphics[width=0.3\linewidth]{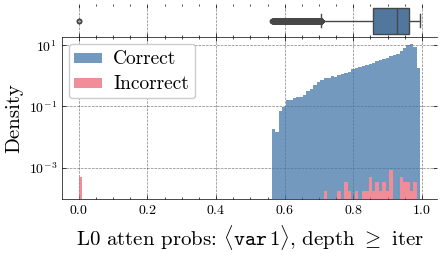} &
        \includegraphics[width=0.3\linewidth]{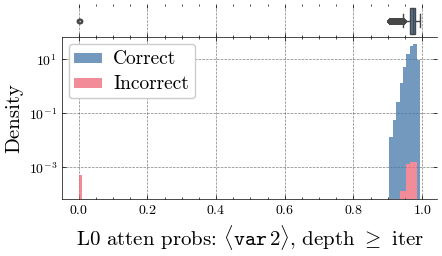} \\
        \includegraphics[width=0.3\linewidth]{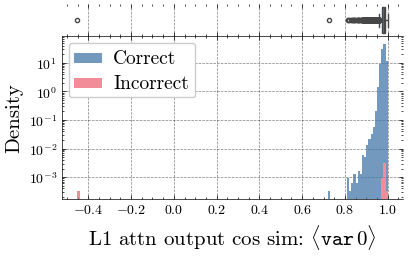} &
        \includegraphics[width=0.3\linewidth]{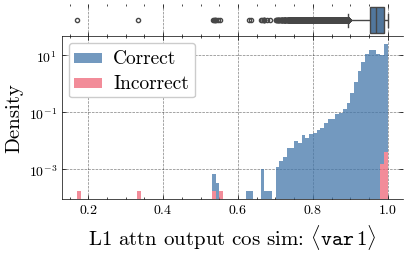} &
        \includegraphics[width=0.3\linewidth]{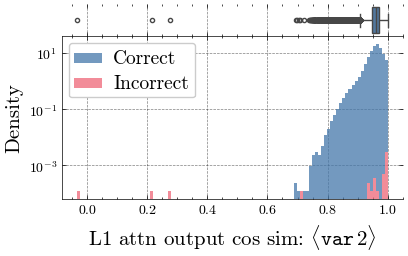} \\
        \includegraphics[width=0.3\linewidth]{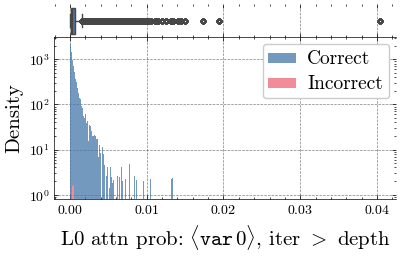} &
        \includegraphics[width=0.3\linewidth]{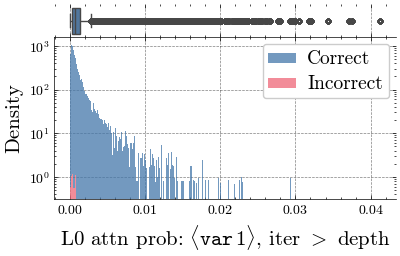} &
        \includegraphics[width=0.3\linewidth]{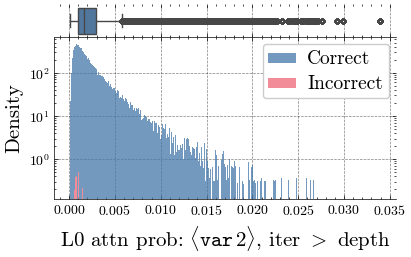}
    \end{tabular}
    \caption{Error analysis. \textbf{Top Row.} Histograms for the group-wise attention probability in the first layer for all three head groups attending to $\VAR{0}$, $\VAR{1}$, and $\VAR{2}$, respectively.
    Here, the target equations considered all satisfy $\mathtt{depth} \ge \mathtt{iter}$ as defined in \eqref{eq:depth_ge_iter}.
    We use different colors to separate the equations based on whether the decoded RHS value is correct or not after the second layer MLP.
    \textbf{Middle Row.} Histograms of the group-wise cosine similarity for the second layer attention head groups' outputs with the target values' embedding. Only equations with $\mathtt{depth} \ge \mathtt{iter}$ are included. 
    \textbf{Bottom Row.} Histograms of the group-wise attention probability in the first layer for all three head groups.
    Here, the target equations considered all satisfy $\mathtt{depth} < \mathtt{iter}$, meaning that the number of iterations is beyond the depth of the equations.
    }
    \label{fig:error-analysis}
\end{figure}

\textbf{\textit{Does the Model Perform Self-Correction?}}
The first two rows in \Cref{fig:error-analysis} are reported only for equations with $\mathtt{depth} \ge \mathtt{iter}$.
This is because the number of iterations required for computing the correct RHS value of equations is at most its depth.
However, if we let the number of iterations go beyond the depth of the equations, as shown in \Cref{fig:error-analysis} (Bottom Row), the first layer attention heads are not able to concentrate their probability mass on the correct variable. 
This finding indicates that there is no further computation performed by the model at an equation position after the number of iterations reaches the depth of the equations, hence the model does not perform self-correction. 
One possible explanation for this to happen is the use of weight-decay in the training process. 
As the value for the $\RHS$ variable is already computed after the number of iterations reaches the depth of the equations, the model can directly pass on the computed value to the next iteration via the residual stream without any further computation.

How to let the model perform self-correction? We observe that the model does not perform self-correction because we only train the model on ``perfect'' data, where the model has no need to perform any further computation beyond the depth of the equations.
In fact, we can let the model perform self-correction by training the model on ``imperfect'' data, where the model has to perform some further computation beyond the depth of the equations.
This motivates our proposal of \methodcolorhighlight{discretesupervisionrecorrectioncolor}{Discrete Latent Space Supervision $\circlearrowleft$} method, which trains the model with corrupted data to teach the model to recover from errors. Consequently, increasing the number of iterations beyond the depth of the input can be useful because it allows the model to correct any errors in previous iterations.
}{}

\ifthenelse{\equal{\compileversion}{bodyonly}}{
    \makecustomtitle
    {\mytitle}
    {
    Systematic, compositional generalization beyond the training distribution remains a core challenge in machine learning---and a critical bottleneck for the emergent reasoning abilities of modern language models.
    This work investigates out-of-distribution (OOD) generalization in Transformer networks using a GSM8K-style modular arithmetic on computational graphs task as a testbed. 
    We introduce and explore a set of four architectural mechanisms aimed at enhancing OOD generalization: \coloremph{\textit{(i)} input-adaptive recurrence}; \coloremph{\textit{(ii)} algorithmic supervision}; \coloremph{\textit{(iii)} anchored latent representations via a discrete bottleneck}; and \coloremph{\textit{(iv)} an explicit error-correction mechanism}.
    Collectively, these mechanisms yield an architectural approach for native and scalable latent space reasoning in Transformer networks with robust algorithmic generalization capabilities.
    We complement these empirical results with a detailed mechanistic interpretability analysis that reveals how these mechanisms give rise to robust OOD generalization abilities.
}
    {\today}
    {\url{https://github.com/Awni00/algorithmic-generalization-transformer-architectures}}

    \vspace{1em}
    \vskip 0.5em  
\begin{figure}[h!]
    \centering
    \begin{subfigure}[t]{0.24\linewidth}
        \centering
        \includegraphics[width=0.90\linewidth]{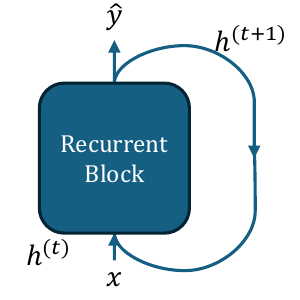}
        \captionsetup{width=.95\linewidth,font=scriptsize}
        \caption{\scriptsize Recurrence \& Adaptive Computation}
    \end{subfigure}
    \begin{subfigure}[t]{0.24\linewidth}
        \centering
        \includegraphics[width=0.90\linewidth]{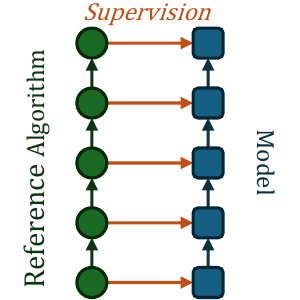}
        \captionsetup{width=.95\linewidth,font=scriptsize}
        \caption{\scriptsize Algorithmic Supervision}
    \end{subfigure}
    \begin{subfigure}[t]{0.24\linewidth}
        \centering
        \includegraphics[width=0.90\linewidth]{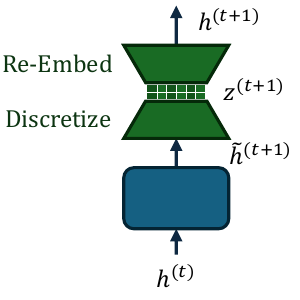}
        \captionsetup{width=.95\linewidth,font=scriptsize}
        \caption{\scriptsize Anchored Discrete Latent Space}
    \end{subfigure}
    \begin{subfigure}[t]{0.24\linewidth}
        \centering
        \includegraphics[width=0.90\linewidth]{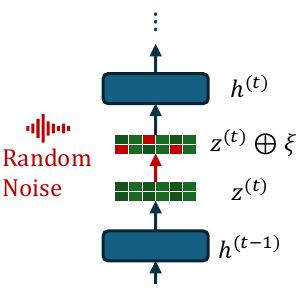}
        \captionsetup{width=.95\linewidth,font=scriptsize}
        \caption{\scriptsize Error Correction}
    \end{subfigure}
    \caption{\textbf{Four key mechanisms enabling robust out-of-distribution generalization in transformer architectures.}
    \textbf{(a)} Recurrence and input-adaptive computation allows models to dynamically allocate computational resources based on problem complexity.
    \textbf{(b)} Algorithmic supervision guides the learning process through structured intermediate representations.
    \textbf{(c)} Anchored discrete latent spaces provide stable reference points for compositional reasoning.
    \textbf{(d)} Error correction mechanisms enable iterative refinement of predictions through feedback loops.
    Together, these mechanisms enable transformers to develop recursive reasoning patterns that generalize beyond their training distribution.}\label{fig:mechanisms}
\end{figure}
    \clearpage

    \vspace{1em}
    {\color{YaleBlue}\tableofcontents}
    \clearpage

    \printbibliography
}{}

\makeatletter
\ifthenelse{\equal{\compileversion}{appendixonly}}{%
    \begin{center}
        {\LARGE\bfseries [Appendix] \@title\par}
    \end{center}
    \vspace{0.5in}

    \printbibliography

}{}
\makeatother

\end{document}